\documentclass[journal]{IEEEtran}
\ifCLASSINFOpdf
   \usepackage[pdftex]{graphicx}
\else
\fi
\usepackage{amsmath}
\usepackage{amssymb}
\usepackage{enumerate} 
\usepackage{cite}
\usepackage{color}
\usepackage{soul}
\usepackage{tabularray}
\usepackage{stfloats}
\usepackage{gensymb}
\usepackage{multirow}
\usepackage{bm}
\usepackage{graphicx}
\usepackage{colortbl}
\usepackage{xcolor}
\usepackage{fontawesome5}
\usepackage{xspace}
\usepackage{suffix}
\usepackage{booktabs}
\usepackage{mathtools}
\usepackage[colorlinks,linkcolor=blue]{hyperref}

\usepackage{algorithm}
\usepackage[noend]{algpseudocode}

\usepackage{array}
\soulregister{\cite}7
\soulregister{\ref}7
\UseRawInputEncoding
\usepackage{tikz}
\usepackage[edges]{forest} 
\floatname{algorithm}{Algorithm}

\usepackage{booktabs}    
\usepackage{geometry}    

\definecolor{Gray}{gray}{0.90}
\definecolor{mygray}{rgb}{.80,.80,.80}
\definecolor{myblue}{rgb}{0.94, 0.95, 1.0}
\definecolor{mylightblue}{rgb}{0.96, 0.97, 1.0}
\definecolor{forestgreen}{rgb}{0.0, 0.5, 0.0}
\definecolor{ashgrey}{rgb}{0.7, 0.75, 0.71}
\definecolor{darkorange}{rgb}{1.0, 0.55, 0.0}
\definecolor{carminepink}{rgb}{0.92, 0.3, 0.26}
\definecolor{lighttealblue}{RGB}{41, 157, 143}
\definecolor{lightplum}{RGB}{233, 196, 106}
\definecolor{harvestgold}{RGB}{216, 118, 89}
\definecolor{lightcoral}{rgb}{0.94, 0.5, 0.5}

\long\def\manuscript#1\par{\makebox[0pt][r]{\hspace{10pt}}\color[rgb]{0.27,0.45,0.77}{}\\[5pt]{{"#1"}}\\[12pt]\color{black}} 
\WithSuffix\long\def\manuscript*#1\par{#1\par} 

\begin{document}

%


\title{Data-Driven Vision-Language Models for Remote Sensing: A Survey}


%
%
%

\author{Yue Zhou,~\IEEEmembership{Member,~IEEE,}
        Shujun Zhao,
        Xue~Yang*,~\IEEEmembership{Member,~IEEE,}
        Ruigang Li,   
        Tianwen Zhang,
        Mengcheng Lan,
        Chaofeng Chen,~\IEEEmembership{Member,~IEEE,}
        Lingfei Ma,~\IEEEmembership{Member,~IEEE,}\\
        Hongjie He,~\IEEEmembership{Member,~IEEE,}
        Jonathan Li*~\IEEEmembership{Fellow,~IEEE}
        
\thanks{Corresponding author: Xue Yang and Jonathan Li. Xue Yang is with the School of Automation and Intelligent Sensing, Shanghai Jiao Tong University, Shanghai, China (Email: yangxue-2019-sjtu@sjtu.edu.cn). Jonathan Li, Yue Zhou, Shujun Zhao, Lingfei Ma are with with the School of GeoAI and Hindon STAI Institute, Key Laboratory of Geographic Information Science (Ministry of Education), East China Normal University, Shanghai, China (Email: jli@geoai.ecnu.edu.cn). Ruigang Li is with Southeast University, China. Tianwen Zhang is with Southwest Jiaotong University, China. Mengcheng Lan is with Nanyang Technological University, Singapore. Chaofeng Chen is with Wuhan University, China. Hongjie He is with the University of Waterloo, Cananda.}

\thanks{This work was supported by the Shanghai Science and Technology Program (25ZR1402268), and the National Natural Science Foundation of China (62506229, 62571315).}

}


%
%

\markboth{IEEE Geoscience and Remote Sensing Magazine}{Shell \MakeLowercase{\textit{et al.}}: Bare Demo of IEEEtran.cls for IEEE Journals}
%



\maketitle


\begin{abstract}

During the deep learning era, innovations in remote sensing (RS) vision models primarily focused on optimizing network architectures for specific tasks and conducting end-to-end training. However, with the advent of the Vision-Language Models (VLMs) era, constrained by substantial computational resource demands, the mainstream research paradigm has shifted towards adapting general-purpose VLMs using RS vertical domain data for fine-tuning, rather than modifying their core architectures. This transition marks a change in the innovation pathway from model-centric to data-centric. Despite the increasing importance of data, the academic community still lacks a systematic, data-centric survey to guide researchers in constructing high-quality RS vision-language datasets for various application scenarios and in conducting effective model fine-tuning and performance evaluation based on these datasets. To address this gap, this paper presents a systematic review of the development status of RS datasets tailored for VLMs. It further provides a comprehensive practical guide for RS researchers in the large model era, covering the entire pipeline from data construction to fine-tuning and evaluation. This paper aims to encourage the academic community to build more diverse and higher-quality vision-language datasets for specialized RS sub-fields. Ultimately, the goal is to converge these efforts into a comprehensive RS vision-language corpus that offers extensive task coverage and a wide geographical distribution, thereby laying a solid data foundation for advancing towards general artificial intelligence in RS.

\end{abstract}

\begin{IEEEkeywords}
Vision-Language Model, Remote Sensing, Dataset, Data Pipeline, Benchmark.
\end{IEEEkeywords}


%
\IEEEpeerreviewmaketitle

\section{Introduction}


\begin{figure*}[!t]
	\begin{center}             
        \includegraphics[width=1\linewidth]{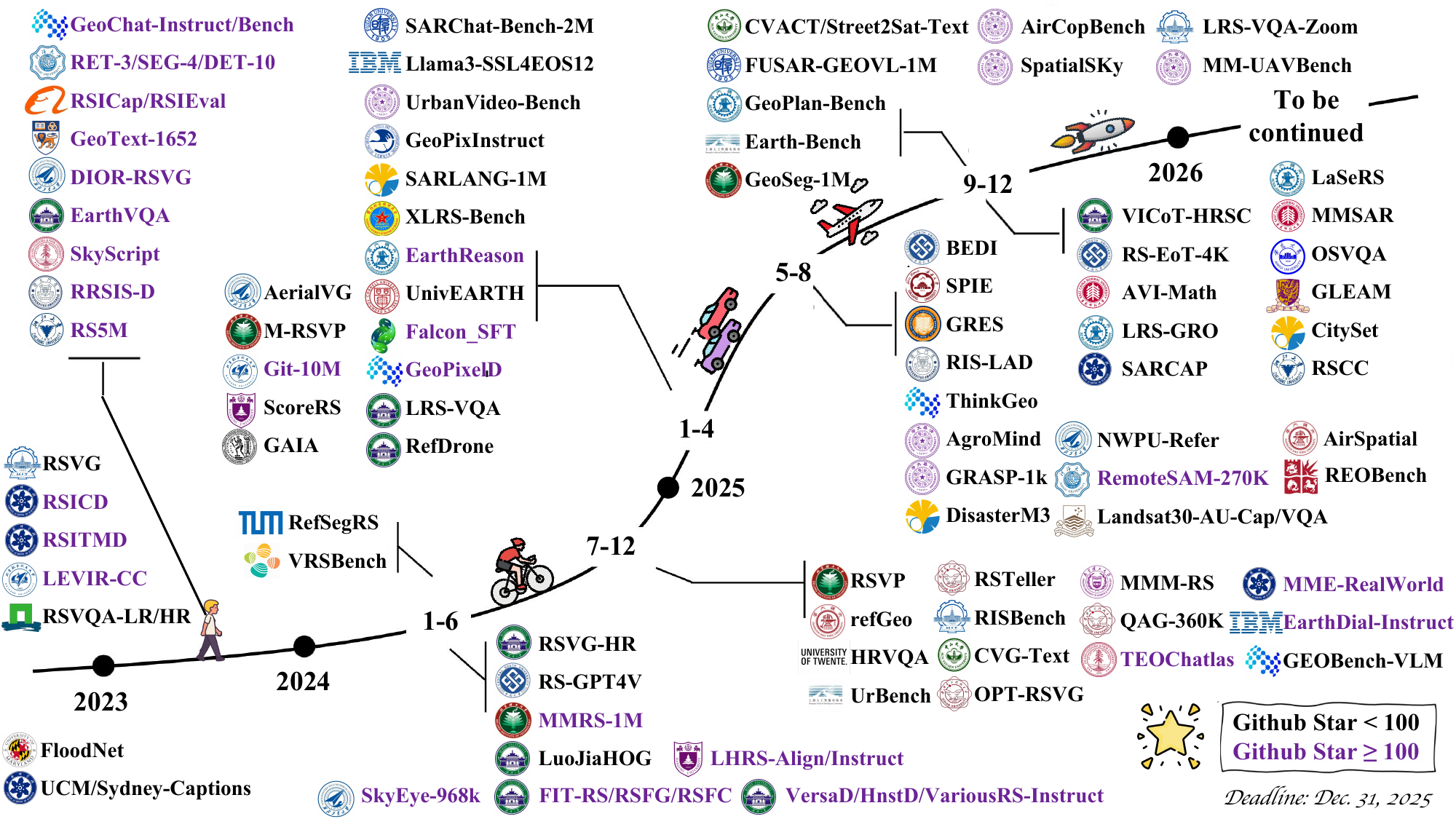}
    \end{center}
    \caption{Representative Vision-Language Datasets in RS.}
    \label{fig:main-logo}        
\end{figure*}




The advent of large-scale pretraining techniques \cite{radford2021learning,hafner2021clip,sun2024alpha} has ushered in artificial intelligence's era of foundation models \cite{moor2023foundation,qin2024tool,awais2025foundation}. Within this context, Vision-Language Models (VLMs) \cite{li2024vision,zhang2024vision,zhang2025large} have demonstrated remarkable capabilities in cross-modal understanding and generation, establishing themselves as foundational architectures across numerous research domains. However, the enormous parameter counts and substantial computational requirements of these models render direct architectural modifications for specific vertical domains \cite{zeng2023x,zhao2025med,hao2025mimoembodied}, such as remote sensing (RS) \cite{weng2025vision,xue2025regression}, prohibitively expensive and technically demanding \cite{achiam2023gpt,chen2023minigpt}. Consequently, a significant paradigm shift is underway in the research community: from a model-centric paradigm focused on designing and optimizing model architectures toward a data-centric paradigm centered on domain-specific data curation and efficient adaptive fine-tuning of general-purpose foundation models \cite{bai2023qwen,chen2023internvl}. This transformation implies that high-quality domain-specific data, rather than model architecture itself, has become the critical enabling factor for unlocking the potential of foundation models and addressing specialized tasks \cite{lu2024visual,chen2026overview}. In particular, fine-tuning general VLMs with large-scale image-text RS data has become the primary method for achieving advanced functionalities \cite{schuhmann2022laion,luo2023lexlip,yu2024capsfusion}.

Broadly, VLMs for RS cover three major technical branches: 1) contrastive learning-based cross-modal representation models (e.g., RemoteCLIP \cite{liu2024remoteclip}); 2) text-guided RS image generation models (e.g., Text2Earth \cite{liu2025text2earth}); and 3) Multimodal Large Language Models (MLLMs). The core research boundary of this survey is the third category, MLLMs, as this class of models has driven a paradigm shift in RS data construction from static annotations to free-form instruction dialogues, which is fully aligned with the data-centric perspective of this work. Since 2023, a paradigm shift driven by VLMs has gained significant momentum \cite{wang2025dynamic,zhang2025critic,prabhu2025trust}. Academic institutions and research organizations worldwide are actively engaging in this interdisciplinary field, leading to the creation of numerous large-scale, multi-task, and multimodal vision-language datasets \cite{wu2024visionllm, truong2025insect, xu2024lvlm}. The field has evolved from foundational works, such as RSVQA \cite{lobry2020rsvqa} in 2023, to more versatile datasets, like GeoChat-Instruct \cite{kuckreja2024geochat} in 2024. By 2025, the development of relevant datasets experienced explosive growth \cite{wei2025hq, chen2024vitamin, dai2023instructblip}, with expectations to enter an intensive launch phase from 2026 onward. 

As shown in Figure \ref{fig:main-logo}, the accelerating pace of innovation is vividly illustrated by a timeline where icons transition from sparse to densely packed, metaphorically shifting from pedestrians and bicycles to cars, aircraft, and rockets, indicating exponential progress. Concurrently, community engagement has become a core driver \cite{pol2024hugging}. The increasing number of GitHub stars for relevant projects reflects a vibrant open-source ecosystem characterized by continuous release, validation, and iteration cycles. A fundamental shift in task paradigms is also evident: dataset nomenclature has evolved from conventional detection \cite{xia2018dota,li2020object,sun2022fair1m,yang2022detecting, zhou2022mmrotate, yang2021r3det} and segmentation \cite{su2017local,volpi2018deep,wang2023samrs,he2023multimodal,li2024review} to encompass advanced tasks requiring complex cognition, such as instruction tuning \cite{liu2023visual,zhang2023instruction,shi2024instruction,liu2024improved,deng2025changechat}, visual reasoning \cite{gupta2023visual,malkinski2023review,chen2023large,chen2024visual}, video understanding \cite{huang2024vtimellm,nie2024slowfocus,shu2025audio,yuan2025videorefer,wang2025videotree}, and geospatial tasks \cite{hou2025geocogent,zhang2024geospatial,yu2025monkuu,singh2024geollm,zhu2025ger}. These trends collectively signal an emerging consensus: the integration of visual and linguistic reasoning is transforming satellite perception from mere seeing to genuine insight, heralding a new era of general-purpose RS intelligence. This data-driven research surge not only underscores the maturation of multimodal intelligence in RS but also provides a dynamic, rapidly evolving platform for global scholars to collectively advance the field toward higher-level universal perception and comprehension \cite{scheider2023geoai,janowicz2020geoai,mai2022review}.

Existing surveys on RS VLMs \cite{li2024vision,zhou2024towards, zhang2024geoscience, xiao2024foundation, weng2025vision} have predominantly focused on methodological analyses at the model level. However, it is essential to acknowledge that, compared to the deep learning era \cite{lecun2015deep,rusk2016deep,zhang2016deep,li2018deep,li2022deep}, the design space for architectural innovation in the foundation model era is relatively constrained, with most contributions adopting modular augmentation and integration strategies \cite{lai2023lisa,zhang2023next,wang2024visionllm}. The fundamental reason lies in the fact that substantial architectural modifications typically necessitate training from scratch on large-scale data to recover the generalization capabilities acquired from massive general-domain corpora \cite{uchida2024using,yu2025opencsg}. The scale of datasets constructed by researchers for specific scenarios is generally insufficient to support such from-scratch training of foundation models. Even when models trained from scratch achieve satisfactory fitting performance on scenario-specific datasets, they often forfeit the powerful zero-shot generalization capabilities inherent to general-purpose models, arguably one of the most valuable properties of foundation models \cite{mall2023remote,xu2024benchmarking}. Although theoretically feasible to incorporate general-domain data for from-scratch training of novel architectures, such research imposes extraordinary computational resource requirements that remain beyond the reach of most academic laboratories \cite{chen2023minigpt,wang2024qwen2}. Therefore, we contend that in the foundation model era, the core competitive advantage of small-to-medium research teams resides in data rather than models. Furthermore, while several surveys on RS datasets exist, their treatment of RS vision-language datasets lacks sufficient depth \cite{xiong2022earthnets,zhou2024towards}. Compounding this issue, the explosive growth of RS vision-language datasets in recent years has rendered existing surveys incapable of providing comprehensive coverage of the current landscape \cite{li2024vision,xiao2024foundation}.

To address these gaps, this paper presents the first systematic review and taxonomy of RS datasets designed for VLMs fine-tuning, aiming to help researchers accurately grasp the developmental trajectory and frontier trends in this field. Beyond serving as an evolutionary chronicle of RS vision-language datasets, this work aspires to provide a practically valuable reference manual that can genuinely assist researchers in overcoming research bottlenecks. Specifically, researchers urgently need clarity on how to systematically construct, annotate, evaluate, and leverage high-quality vision-language datasets for diverse RS application scenarios \cite{wang2025disasterm3,wu2025fit,wang2026cityset}, as well as how to select effective fine-tuning strategies \cite{lin2024data,zheng2025learning,anisuzzaman2024fine,tian2024hydralora,bian2025lora} and conduct rigorous performance evaluation \cite{prabhu2025trust, NEURIPS2024_fe2fc7dc,zhang2024vision,shool2025systematic} based on such high-quality data. The absence of such guiding knowledge has, to a certain extent, led to redundant data construction efforts, ad hoc fine-tuning practices, and fragmented evaluation standards, thereby hindering effective knowledge accumulation and efficient technological iteration within the field.

To address the aforementioned challenges, this paper proposes a novel Data Curation research framework, GeoChef, which draws a systematic analogy between the development process of RS VLMs and culinary procedures. Within this framework, we reconceptualize the role of the researcher as a Chef who must not only possess keen observational acuity for iterative model optimization but also engage deeply in every stage of data selection and processing. The framework encompasses four key phases. The first phase, Ingredient Atlas, systematically establishes a taxonomic structure for RS vision-language datasets and analyzes the characteristics and evolutionary patterns of different data types. Just as a chef must thoroughly understand the properties of each ingredient, researchers must familiarize themselves with the task attributes and applicable scenarios of various RS data modalities. The second phase, Ingredient Processing, provides a comprehensive methodology for constructing the dataset. Analogous to ingredient preparation in cooking, raw RS data must undergo multiple processing stages to ultimately yield refined, well-annotated, high-quality datasets that preserve the professional characteristics of RS data while conforming to the format specifications required for MLLM training \cite{hu2023mdas,hong2023cross}. The third phase, Cooking Manual, elaborates on the selection and implementation of dataset fine-tuning strategies, enabling researchers to choose optimal fine-tuning approaches based on specific task requirements and resource constraints. The fourth phase, Gourmet Tasting, establishes a comprehensive evaluation framework that introduces commonly used benchmarks and multidimensional evaluation metrics for different tasks \cite{hossin2015review,dey2020robust,naidu2023review}. This evaluation approach resembles the professional assessment of culinary critics, attending to both overall performance and fine-grained details. Throughout the entire research cycle, the Chef (researcher) consistently assumes the central role of observation and iterative optimization, progressively mastering the complete methodology to customize high-performance VLMs for specific RS application scenarios through continuous experimentation, evaluation, and refinement. Through this framework, this paper provides RS researchers with end-to-end guidance from data preparation to model evaluation, facilitating the transformation of researchers from passive data consumers to proactive GeoChef, ultimately enabling qualitative leaps in RS MLLM performance. The main contributions of this paper are threefold:

\begin{itemize}

\item More than 120 publicly available vision-language datasets in the field of Earth observation are systematically surveyed, with detailed categorization and in-depth analysis conducted according to task types.

\item A dynamic, community-driven data warehouse is constructed, support the addition of independent datasets by researchers and enabling multidimensional retrieval, which significantly improves the efficiency of discovering and accessing RS vision-language datasets.

\item A step-by-step practical guide is provided for researchers in RS, covering the entire process from data preparation to model evaluation, thereby promoting the development of RS vision-language models toward greater specialization and practicality.

\end{itemize}

The remainder of this paper is organized as follows. Section \ref{sec:taxonomy} introduces the taxonomy and a tree diagram of remote sensing vision-language datasets. Section \ref{sec:evolution} elaborates on the details of task-oriented datasets and their evolution trends. Section \ref{sec:warehouse} proposes a community-driven repository to facilitate efficient data access for researchers. Section \ref{sec:construction} reviews three mainstream methodologies for constructing datasets. Section \ref{sec:fine-tuning} discusses common fine-tuning strategies for vision-language models. Section \ref{sec:evaluation} outlines benchmarks and evaluation metrics for assessing the model. Section \ref{sec:future} presents future research directions. Section \ref{sec:conclusion} concludes the paper. The overall framework follows a logical progression from fundamental concepts and methodologies to evaluation and outlook.

\section{The Taxonomy of RS Vision-Language Datasets}
\label{sec:taxonomy}

\begin{figure}[!t]
	\begin{center}             
        \includegraphics[width=1\linewidth]{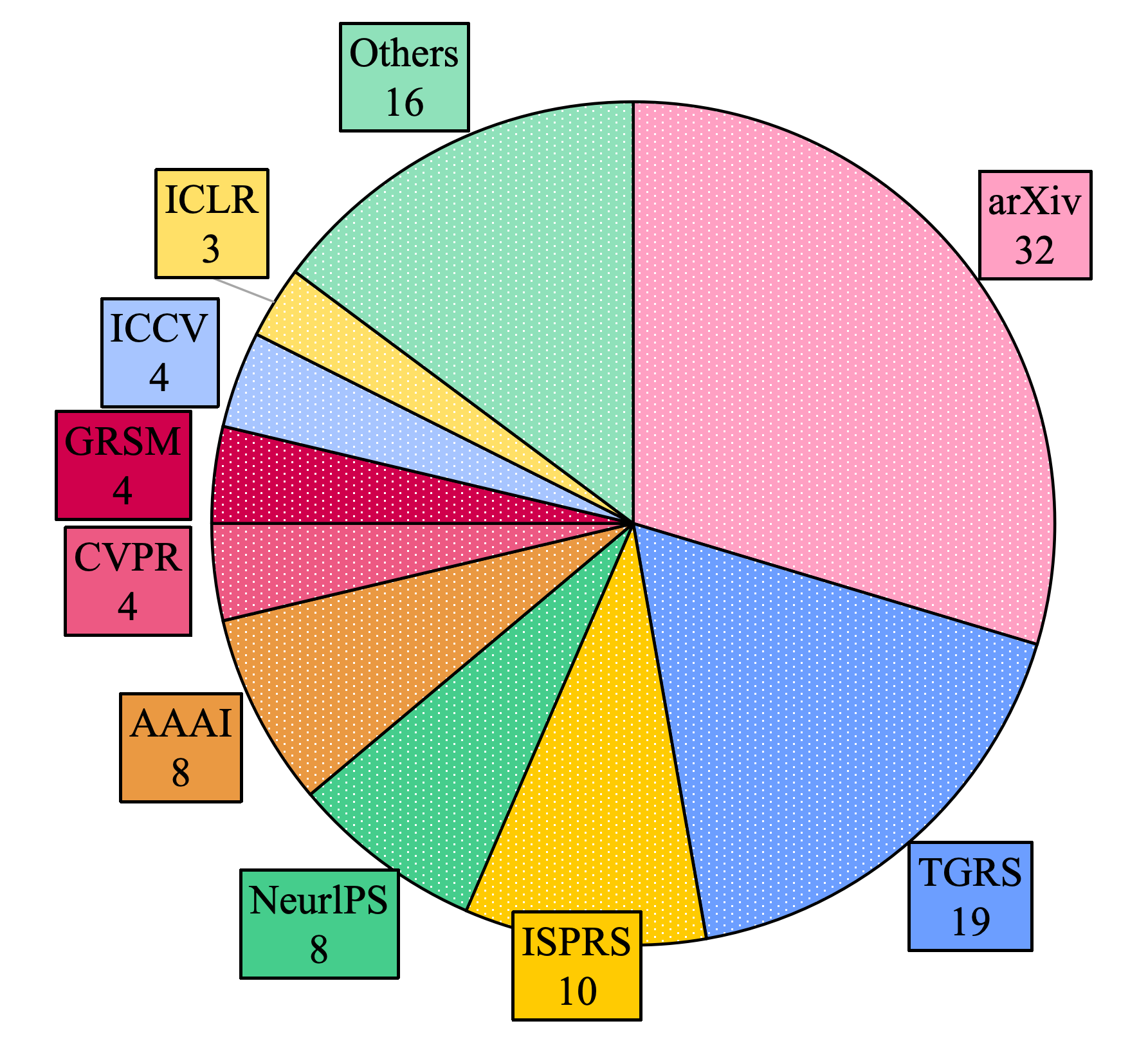}
    \end{center}
    \caption{Publication of RS Vision-Language Datasets}
    \label{fig:Venue}        
\end{figure}

This survey provides a systematic analysis of over 100 collected vision-language datasets. Figure \ref{fig:main-logo} illustrate the growth trend of RS vision-language datasets over the past five years. The emergence of MLLMs has significantly accelerated the expansion of RS vision-language data. Notably, the number of RS vision-language datasets doubled from 2023 to 2024. Furthermore, the number of new datasets introduced solely in 2025 has surpassed the cumulative total of all such datasets created before that year. The growth rate observed in 2025 exceeds an exponential pattern, indicating that the field of RS vision-language datasets is currently experiencing a phase of rapid expansion. Additionally, Figure \ref{fig:Venue} illustrates the distribution of publications related to existing RS vision-language datasets. Aside from works posted on the arXiv preprint server, TGRS is the preferred journal for such papers, accounting for 19 publications. It is followed by journals and conferences such as ISPRS and NeurIPS. Moreover, leading venues, including AAAI, CVPR, GRSM, ICCV, ICLR, and ECCV, have also published research on RS vision-language datasets. This publication distribution suggests that peer-reviewed work in this field generally maintains high quality and attracts broad attention from reputable forums.

\begin{figure*}[!t]
	\begin{center}             
        \includegraphics[width=1\linewidth]{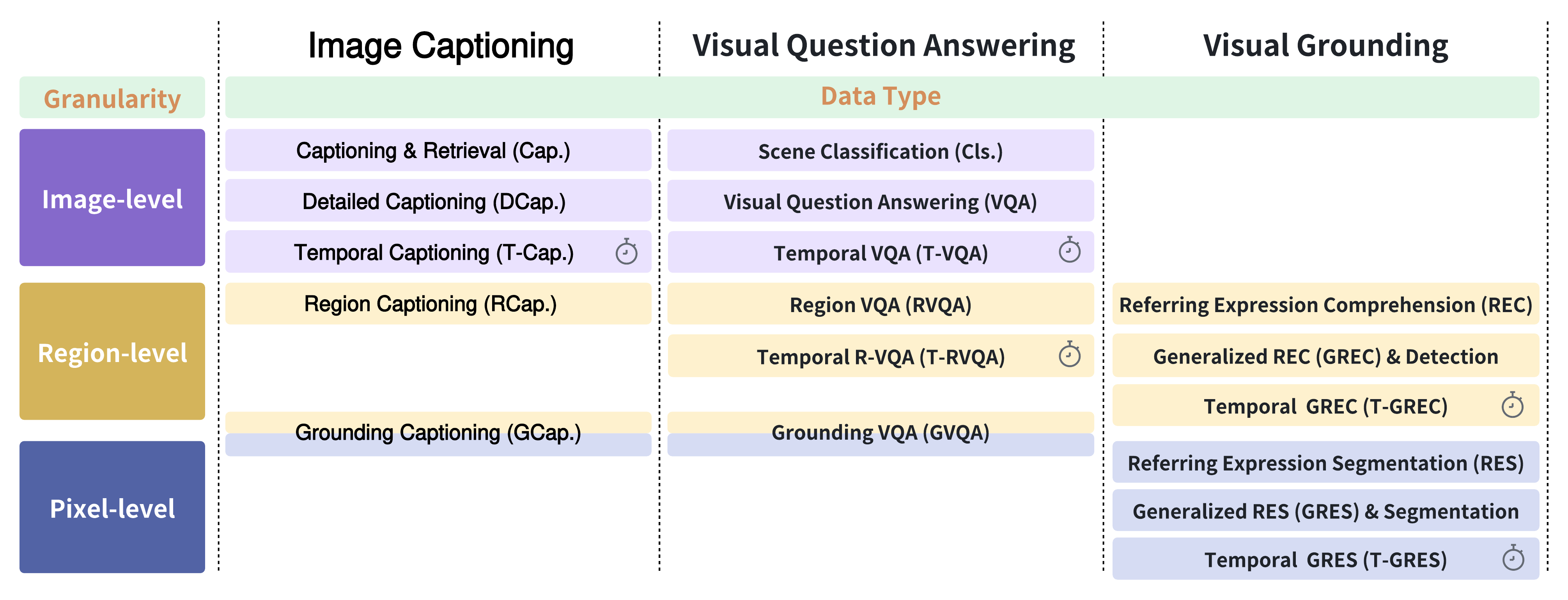}
    \end{center}
    \caption{Taxonomy of RS Vision-Language Data.}
    \label{fig:data_type}        
\end{figure*}

\subsection{Taxonomy}
\label{sec:tax}

In existing surveys on RS VLMs, a common practice is to categorize tasks based on model capabilities. However, we observe that such a capability-oriented taxonomy is inherently non-exhaustive and lacks completeness. Whenever new research emerges, it necessitates continuous supplementation of new capability types, resulting in a classification system that remains in a state of passive expansion. To address this limitation, this paper shifts perspective to the forms of data input and output, proposing a data-centric taxonomy for RS vision-language tasks. 

As shown in Figure \ref{fig:data_type}, this framework categorizes tasks into three major types, distinguished by their conventional definitions in the RS community and distinct evaluation metrics:
\begin{itemize}
    \item \textbf{Image Captioning} is primarily evaluated using \textbf{text similarity metrics}, since there often exist multiple valid and natural descriptions for an image, making a simple binary ``correct/incorrect'' judgment infeasible.

    \item \textbf{VQA} is uniquely evaluated by \textbf{objective binary correctness}, judging whether the model's response to multiple-choice, true/false, or open factual questions is definitively right or wrong.

    \item \textbf{Visual grounding}  inherits the evaluation paradigm of classic object detection and segmentation tasks, using spatial overlap metrics to quantify alignment between predictions and ground-truth annotations.
\end{itemize}

Furthermore, each task category can be subdivided according to the granularity of the image region being processed into image-level, region-level, and pixel-level subcategories. Additionally, by extending the input from a single image to multiple images, corresponding temporal tasks can be constructed to accommodate the needs of dynamic scene analysis and change detection. Based on these three dimensions -- task type, regional granularity, and temporal dynamics -- we systematically organize existing RS vision-language tasks into three major categories and 16 subcategories.

\subsection{Tree Diagram of Existing RS Vision-Language Datasets}

To facilitate researchers in quickly locating datasets of each subcategory, we have created a tree diagram as illustrated in Figure \ref{fig:categorization_of_survey}. This diagram, comprising three hierarchical levels, vividly showcases the developmental panorama of the current RS visual language domain, and its structure inherently reveals several profound patterns. The most prominent characteristic is the blurring and restructuring of task boundaries. A single dataset frequently appears under multiple seemingly distinct task branches. For instance, EarthDial-Instruct \cite{soni2024earthdial} is simultaneously present under description, classification, and localization. This is not a classification error, but rather reveals a core trend. The research paradigm in the field is shifting from serving single tasks towards constructing multi-task general-purpose datasets capable of handling description, question answering, localization, and classification. The aim is to train or evaluate more general RS foundation models. 

\tikzstyle{my-box}=[
    rectangle,
    rounded corners,
    text opacity=1,
    minimum height=1.5em,
    minimum width=5em,
    inner sep=2pt,
    align=center,
    fill opacity=.5,
]
\tikzstyle{cause_leaf}=[my-box, minimum height=1.5em,
    fill=lighttealblue!20, text=black, align=left,font=\tiny,
    inner xsep=2pt,
    inner ysep=4pt,
]
\tikzstyle{detect_leaf}=[my-box, minimum height=1.5em,
    fill=lightplum!20, text=black, align=left,font=\tiny,
    inner xsep=2pt,
    inner ysep=4pt,
]
\tikzstyle{mitigate_leaf}=[my-box, minimum height=1.5em,
    fill=harvestgold!20, text=black, align=left,font=\tiny,
    inner xsep=2pt,
    inner ysep=4pt,
]
\begin{figure*}[!t]
    \vspace{-0.3cm}
    \centering
    \resizebox{\textwidth}{!}{
        \begin{forest}
            forked edges,
            for tree={
                grow=east,
                reversed=true,
                anchor=base west,
                parent anchor=east,
                child anchor=west,
                base=left,
                font=\small,
                rectangle,
                rounded corners,
                align=left,
                minimum width=4em,
                edge+={darkgray, line width=1pt},
                s sep=3pt,
                inner xsep=2pt,
                inner ysep=3pt,
                ver/.style={rotate=90, child anchor=north, parent anchor=south, anchor=center},
            },
            where level=1{text width=4.0em,font=\scriptsize,}{},
            where level=2{text width=5.0em,font=\scriptsize,}{},
            where level=3{text width=20.0em,font=\scriptsize,}{},
            where level=4{text width=20.0em,font=\scriptsize,}{},
            [
                Remote Sensing Vision Language Data, ver, color=carminepink!100, fill=carminepink!15,
                text=black
                [
                    Cap.  (\S \ref{sec:cap}), color=lighttealblue!100, fill=lighttealblue!100, text=black
                    [
                        Cap. (\S \ref{sec:cap_sub}), color=lighttealblue!100, fill=lighttealblue!60, text=black
                        [
                            {CVACT-Text~\cite{ye2025street2sat}, CVG-Text~\cite{ye2024cross}, DisasterM3~\cite{wang2025disasterm3}, Falcon\_SFT~\cite{yao2025falcon}, FIT~\cite{wu2025fit}, FUSAR-GEOVL-1M~\cite{yang2025fusarklip}, \\GAIA~\cite{zavras2025gaia}, GeoLangBind-2M~\cite{xiong2025geolangbind}, GLEAM~\cite{lu2025gleam}, Git-10M~\cite{liu2025text2earth}, LAION-EO~\cite{czerkawski2023laion}, LHRS-Align~\cite{muhtar2024lhrs}, \\LHRS-Align-Recap~\cite{li2025lhrsbotnova}, ScoreRS~\cite{muhtar2025scorers}, SPIE~\cite{si2025spex}, SSL4EOS12~\cite{marimo2025beyond}, Street2Sat-Text~\cite{ye2025street2sat}, \\Landsat30-AU-Cap~\cite{ma2025landsat30au}, LRS2M~\cite{chen2025lrsclip}, MMRS-1M~\cite{zhang2024earthgpt}, MMM-RS~\cite{luo2024mmm}, MOCO~\cite{pan2024moco}, MMSAR~\cite{wang2025sarclip}, \\M-RSVP~\cite{zhang2025earthgpt}, NAIP-OSM~\cite{mall2023remote}, NWPU-Captions~\cite{cheng2022nwpu-cap}, RemoteCLIP~\cite{liu2024remoteclip}, REOBench~\cite{li2025reobench}, RS5M~\cite{zhang2024rs5m}, \\RSICD~\cite{lu2018rsicd}, RSITMD~\cite{yuan2022rsitmd}, RS-VL3M~\cite{hu2025ringmoagent}, SARCap~\cite{jiang2025sarcap}, SARChat-Bench-2M~\cite{ma2025sarchat}, SARLANG-1M~\cite{wei2025sarlang}, \\SAR-TEXT~\cite{he2025sartext}, SARVLM-1M~\cite{ma2025sarvlm}, SkyScript~\cite{wang2024skyscript}, Sydney-Captions~\cite{qu2016cap}, UCM-Captions~\cite{qu2016cap}, \\UAV-Captions~\cite{hoxha2022uavcap}}, cause_leaf, text width=25em
                        ]
                    ]
                    [
                        DCap. (\S \ref{sec:d_cap}), color=lighttealblue!100, fill=lighttealblue!60, text=black
                        [
                            {CVACT-Text~\cite{ye2025street2sat}, DDFAV~\cite{li2025ddfav}, Falcon\_SFT~\cite{yao2025falcon}, FIT~\cite{wu2025fit}, FIT-RS~\cite{luo2024skysensegpt}, FIT-RSFG~\cite{luo2024skysensegpt}, GEOBench-VLM~\cite{danish2024geobench}, \\GeoChat-Instruct~\cite{kuckreja2024geochat}, GeoPixInstruct~\cite{ou2025geopix}, GLEAM~\cite{lu2025gleam}, Landsat30-AU-Cap~\cite{ma2025landsat30au}, LHRS-Align~\cite{muhtar2024lhrs}, \\LHRS-Align-Recap~\cite{li2025lhrsbotnova}, LRS2M~\cite{chen2025lrsclip}, LuoJiaHOG~\cite{zhao2025luojiahog}, M-RSVP~\cite{zhang2025earthgpt}, RS-GPT4V~\cite{xu2024rs}, RSCC~\cite{chen2025rscc}, \\RSICap~\cite{hu2023rsgpt}, RSTeller~\cite{ge2024rsteller}, RSVP~\cite{zhang2025earthmaker}, SAR-TEXT~\cite{he2025sartext}, Street2Sat-Text~\cite{ye2025street2sat}, VersaD~\cite{pang2024vhm}, VRSBench~\cite{li2024vrsbench}, \\XLRS-Bench~\cite{wang2025xlrs}}, cause_leaf, text width=25em
                        ]
                    ] 
                    [
                        RCap. (\S \ref{sec:r_cap}), color=lighttealblue!100, fill=lighttealblue!60, text=black
                        [
                            {CapERA~\cite{bashmal2023capera}, ChangeChat-87k~\cite{deng2025changechat}, EarthDial-Instruct~\cite{soni2024earthdial}, Falcon\_SFT~\cite{yao2025falcon}, FIT-RS~\cite{luo2024skysensegpt}, FIT-RSFG~\cite{luo2024skysensegpt}, \\GeoLLaVA~\cite{elgendy2024geollava}, LEVIR-CC~\cite{liu2022levir-cc}, MMRS-1M~\cite{zhang2024earthgpt}, M-RSVP~\cite{zhang2025earthgpt}, RSVP~\cite{zhang2025earthmaker}, SARChat-Bench-2M~\cite{ma2025sarchat}}, cause_leaf, text width=25em
                        ]
                    ] 
                    [
                        GCap. (\S \ref{sec:g_cap}), color=lighttealblue!100, fill=lighttealblue!60, text=black
                        [
                            {EarthDial-Instruct~\cite{soni2024earthdial}, GeoChat-Bench~\cite{kuckreja2024geochat}, GeoChat-Instruct~\cite{kuckreja2024geochat}, GeoPixelD~\cite{shabbir2025geopixel}, RSVP~\cite{zhang2025earthmaker}, SPIE~\cite{si2025spex}}, cause_leaf, text width=25em
                        ]
                    ] 
                    [
                        T-Cap. (\S \ref{sec:t_cap}), color=lighttealblue!100, fill=lighttealblue!60, text=black
                        [
                            {EarthDial-Instruct~\cite{soni2024earthdial}, RSCC~\cite{chen2025rscc}}, cause_leaf, text width=25em
                        ]
                    ] 
                ]
                [
                    VQA(\S \ref{sec:vqa}), color=lightplum!100, fill=lightplum!100, text=black
                    [
                        VQA (\S \ref{sec:vqa_sub}), color=lightplum!100, fill=lightplum!60, text=black
                        [
                            {AVI-Math~\cite{zhou2025avimath}, AgroMind~\cite{li2025agromind}, A2Seek~\cite{mo2025a2seek}, BEDI~\cite{guo2025bedi}, COREval~\cite{an2024coreval}, CRSVQA~\cite{zhang2023crsvqa}, CitySet~\cite{wang2026cityset}, \\DDFAV~\cite{li2025ddfav}, DisasterM3~\cite{wang2025disasterm3}, Earth-Bench~\cite{feng2025earthagent}, EarthVQA~\cite{wang2024earthvqa}, FIT-RSFG~\cite{luo2024skysensegpt}, FIT-RSRC~\cite{luo2024skysensegpt}, \\RS-VL3M~\cite{hu2025ringmoagent}, RS-EoT-4K~\cite{shao2025socrates}, ScoreRS~\cite{muhtar2025scorers}, FloodNet~\cite{maryam2021floodnet}, FUSAR-GEOVL-1M~\cite{yang2025fusarklip}, GeoPlan-Bench~\cite{li2025geoplan}, \\GeoPixInstruct~\cite{ou2025geopix}, Geo3DVQA~\cite{tsujimoto2025geo3dvqa}, HnstD~\cite{pang2024vhm}, HRVQA~\cite{li2024hrvqa}, KnowFlow-Bench~\cite{chen2025canglingknowflow}, \\Landsat30-AU-VQA~\cite{ma2025landsat30au}, LaSeRS~\cite{xin2025segearthr2}, LHRS-Instruct-Plus~\cite{li2025lhrsbotnova}, LRS-GRO~\cite{liu2025zoomearth}, LRS-VQA~\cite{luo2025large}, \\MME-RealWorld~\cite{zhang2024mme}, MM-UAVBench~\cite{dai2025mmuavbench}, OSVQA~\cite{zhao2025osvqa}, RemoteCount~\cite{liu2024remoteclip}, REOBench~\cite{li2025reobench}, RSIEval~\cite{hu2023rsgpt}, \\RSIVQA~\cite{zheng2021mutual}, RSVQA-HR~\cite{lobry2020rsvqa}, RSVQA-LR~\cite{lobry2020rsvqa}, RSVQAxBEN~\cite{lobry2021rsvqaxben}, SARLANG-1M~\cite{wei2025sarlang}, ThinkGeo~\cite{shabbir2025thinkgeo}, \\UnivEARTH~\cite{kao2025univearth}, UrbanVideo-Bench~\cite{zhao2025urbanvideobench}, VLEO-Bench~\cite{zhang2024good}, VICoT-HRSC~\cite{wang2025vicotagent}, VQA-TextRS~\cite{al2022open}, \\VRSBench~\cite{li2024vrsbench}, XLRS-Bench~\cite{wang2025xlrs}}, detect_leaf, text width=25em
                        ]
                    ]
                    [
                        Cls. (\S \ref{sec:class}), color=lightplum!100, fill=lightplum!60, text=black
                        [
                            {EarthDial-Instruct~\cite{soni2024earthdial}, Falcon\_SFT~\cite{yao2025falcon}, FIT-RS~\cite{luo2024skysensegpt}, FUSAR-GEOVL-1M~\cite{yang2025fusarklip}, MMRS-1M~\cite{zhang2024earthgpt}, \\M-RSVP~\cite{zhang2025earthgpt}, RSVP~\cite{zhang2025earthmaker}, SARCap~\cite{jiang2025sarcap}, SARChat-Bench-2M~\cite{ma2025sarchat}, SARLANG-1M~\cite{wei2025sarlang}, VLEO-Bench~\cite{zhang2024good}, \\VariousRS-Instruct~\cite{pang2024vhm}}, detect_leaf, text width=25em
                        ]
                    ]
                    [
                        RVQA (\S \ref{sec:r_vqa}), color=lightplum!100, fill=lightplum!60, text=black
                        [
                            {RSVP~\cite{zhang2025earthmaker}, TEOChatlas~\cite{irvin2024teochat}, UrBench~\cite{zhou2024urbench}, XLRS-Bench~\cite{wang2025xlrs}}, detect_leaf, text width=25em
                        ]
                    ]
                    [
                        GVQA (\S \ref{sec:g_vqa}), color=lightplum!100, fill=lightplum!60, text=black
                        [
                            {LaSeRS~\cite{xin2025segearthr2}, LRS-GRO~\cite{liu2025zoomearth}, LRS-VQA~\cite{luo2025large}, QAG-360K~\cite{li2024qvg}}, detect_leaf, text width=25em
                        ]
                    ]
                    [
                        T-VQA (\S \ref{sec:t_vqa}), color=lightplum!100, fill=lightplum!60, text=black
                        [
                            {A2Seek~\cite{mo2025a2seek}, CDVQA~\cite{yuan2022cdvqa}, ChangeChat-87k~\cite{deng2025changechat}, EarthDial-Instruct~\cite{soni2024earthdial}, GEOBench-VLM~\cite{danish2024geobench}, \\QVG-360K~\cite{li2024qvg}, TEOChatlas~\cite{irvin2024teochat}, UrbanVideo-Bench~\cite{zhao2025urbanvideobench}, VLEO-Bench~\cite{zhang2024good}}, detect_leaf, text width=25em
                        ]
                    ]
                ]
                [
                    VG (\S \ref{sec:vg}), color=harvestgold!100, fill=harvestgold!100, text=black
                    [
                        REC (\S \ref{sec:rec}), color=harvestgold!100, fill=harvestgold!60, text=black
                        [
                            {AVVG~\cite{zhou2024geoground}, A2Seek~\cite{mo2025a2seek}, COREval~\cite{an2024coreval}, DIOR-RSVG~\cite{zhan2023dior-rsvg}, RS-EoT-4K~\cite{shao2025socrates}, EarthDial-Instruct~\cite{soni2024earthdial}, \\GeoPixInstruct~\cite{ou2025geopix}, GeoText-1652~\cite{chu2024towards}, LaSeRS~\cite{xin2025segearthr2}, MMRS-1M~\cite{zhang2024earthgpt}, OPT-RSVG~\cite{li2024opt-rsvg}, RSVG~\cite{sun2022rsvg}, \\RSVG-HD~\cite{lan2024lqvg}, SARLANG-1M~\cite{wei2025sarlang}, ScoreRS~\cite{muhtar2025scorers}, ThinkGeo~\cite{shabbir2025thinkgeo}, UAVVG~\cite{liu2025aerialvg}, UrBench~\cite{zhou2024urbench}, \\VLEO-Bench~\cite{zhang2024good}, VRSBench~\cite{li2024vrsbench}, XLRS-Bench~\cite{wang2025xlrs}, refGeo~\cite{zhou2024geoground}}, mitigate_leaf, text width=25em
                        ]
                    ]
                    [
                        GREC (\S \ref{sec:grec}), color=harvestgold!100, fill=harvestgold!60, text=black
                        [
                            {Falcon\_SFT~\cite{yao2025falcon}, FIT-RS~\cite{luo2024skysensegpt}, FUSAR-GEOVL-1M~\cite{yang2025fusarklip}, GeoChat-Bench~\cite{kuckreja2024geochat}, GeoChat-Instruct~\cite{kuckreja2024geochat}, \\GEOBench-VLM~\cite{danish2024geobench}, GeoPixInstruct~\cite{ou2025geopix}, LHRS-Instruct-Plus~\cite{li2025lhrsbotnova}, MMRS-1M~\cite{zhang2024earthgpt}, RemoteSam-270K~\cite{yao2025remotesam}, \\REOBench~\cite{li2025reobench}, refDrone~\cite{sun2025refdrone}, RS-VL3M~\cite{hu2025ringmoagent}, SARChat-Bench-2M~\cite{ma2025sarchat}}, mitigate_leaf, text width=25em
                        ]
                    ]
                    [
                        T-GREC (\S \ref{sec:grec}), color=harvestgold!100, fill=harvestgold!60, text=black
                        [
                            {TEOChatlas~\cite{irvin2024teochat}}, mitigate_leaf, text width=25em
                        ]
                    ]
                    [
                        RES (\S \ref{sec:res}), color=harvestgold!100, fill=harvestgold!60, text=black
                        [
                            {AVVG~\cite{zhou2024geoground}, GeoSeg-1M~\cite{ni2025unigeoseg}, GRASP-1K~\cite{jiang2025grasp}, GRES~\cite{quenum2025lisat}, LaSeRS~\cite{xin2025segearthr2}, M-RSVP~\cite{zhang2025earthgpt}, refGeo~\cite{zhou2024geoground}, \\RISBench~\cite{dong2024risbench}, RIS-LAD~\cite{ye2025rislad}, RRSIS-D~\cite{liu2024rotated}}, mitigate_leaf, text width=25em
                        ]
                    ]
                    [
                        GRES (\S \ref{sec:gres}), color=harvestgold!100, fill=harvestgold!60, text=black
                        [
                            {EarthReason~\cite{li2025segearth}, NWPU-Refer~\cite{yang2025nwpurefer}, Falcon\_SFT~\cite{yao2025falcon}, GEOBench-VLM~\cite{danish2024geobench}, GeoPixelD~\cite{shabbir2025geopixel}, \\GeoPixInstruct~\cite{ou2025geopix}, RefSegRS~\cite{yuan2024rrsis}, RemoteSam-270K~\cite{yao2025remotesam}, REOBench~\cite{li2025reobench}}, mitigate_leaf, text width=25em
                        ]
                    ]
                    [
                        T-GRES (\S \ref{sec:tgres}), color=harvestgold!100, fill=harvestgold!60, text=black
                        [
                            {ChangeChat-87k~\cite{deng2025changechat}, Falcon\_SFT~\cite{yao2025falcon}}, mitigate_leaf, text width=25em
                        ]
                    ]
                ]
            ]
        \end{forest}
    }
    \caption{The main categorization of this survey.}
    \label{fig:categorization_of_survey}
\end{figure*}
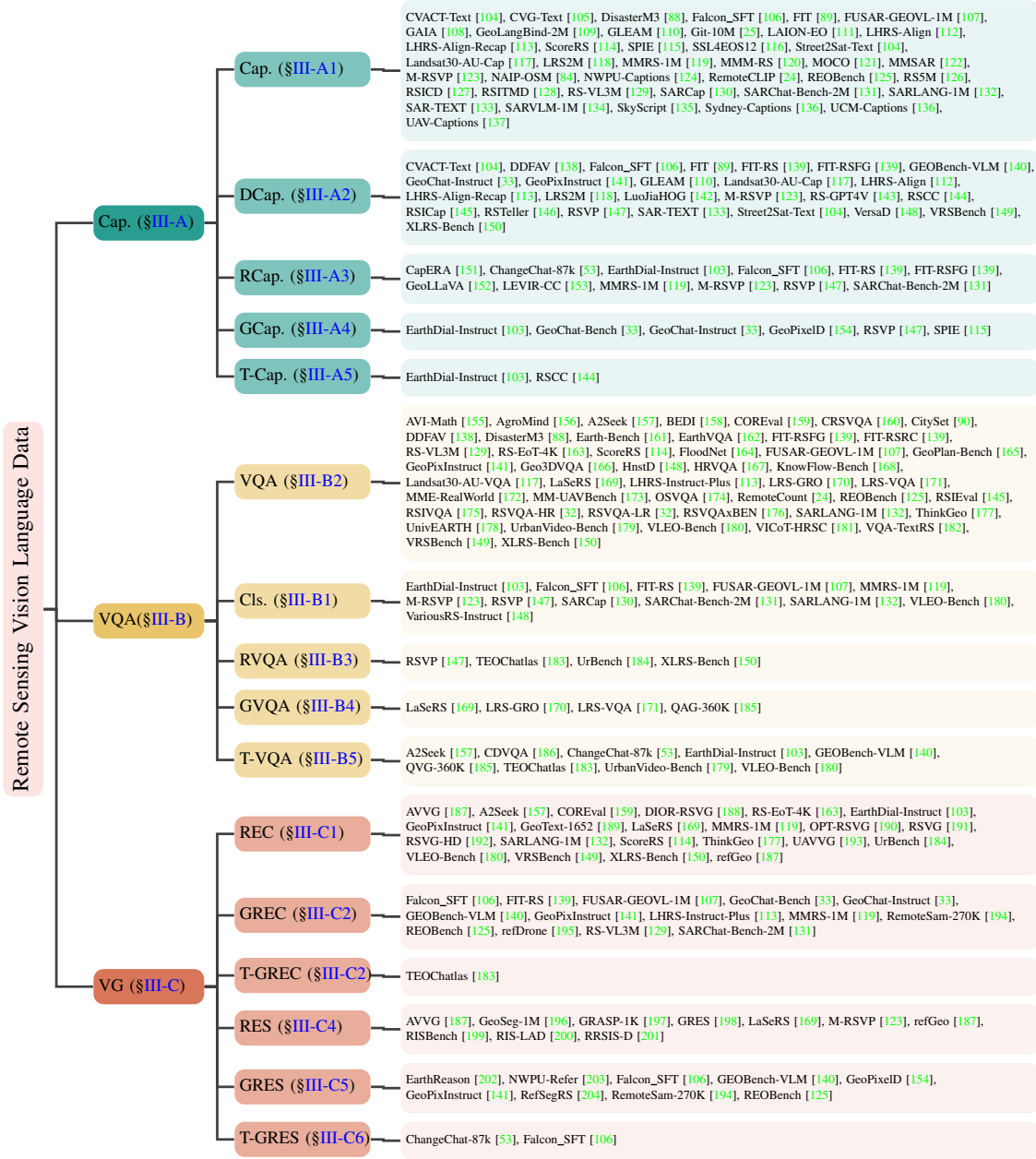

Within this trend, VQA has indisputably become the hub and main battleground of the field. In the diagram, the VQA branch not only has the finest subcategories and contains the most datasets, but has also spurred the creation of numerous specialized evaluation sets often named with terms like Bench or Eval. This signifies the field's entry into a mature stage of systematically assessing model capabilities, where the research focus shifts from whether a function can be implemented to how capable and comprehensive the model is. Simultaneously, the organizational principle of the tree diagram is distinctly marked by the domain-specific characteristics of RS. It is not simply divided by image-text modalities but is rooted in tasks, and particularly emphasizes two dimensions less standard in other visual fields: Geo-referencing and Temporal sequencing. This intuitively suggests that understanding spatial location and change perception are the core challenges and value propositions of RS visual language research. Finally, the names of the datasets themselves form a concise history of the field's development. High-frequency terms such as RS, SAR, Geo, and Earth indicate data sources. Terms like Cap, VQA, and VG define core tasks, while Instruct, Bench, and size indicators such as 1M or 10M reveal the current hotspots of instruction tuning, performance benchmarking, and large-scale pre-training, respectively. The entire map clearly delineates a developmental trajectory from specialized to general-purpose, from single-task to multi-task, and from training-driven to evaluation-driven. 


\begin{figure*}[!t]
	\begin{center}             
        \includegraphics[width=1\linewidth]{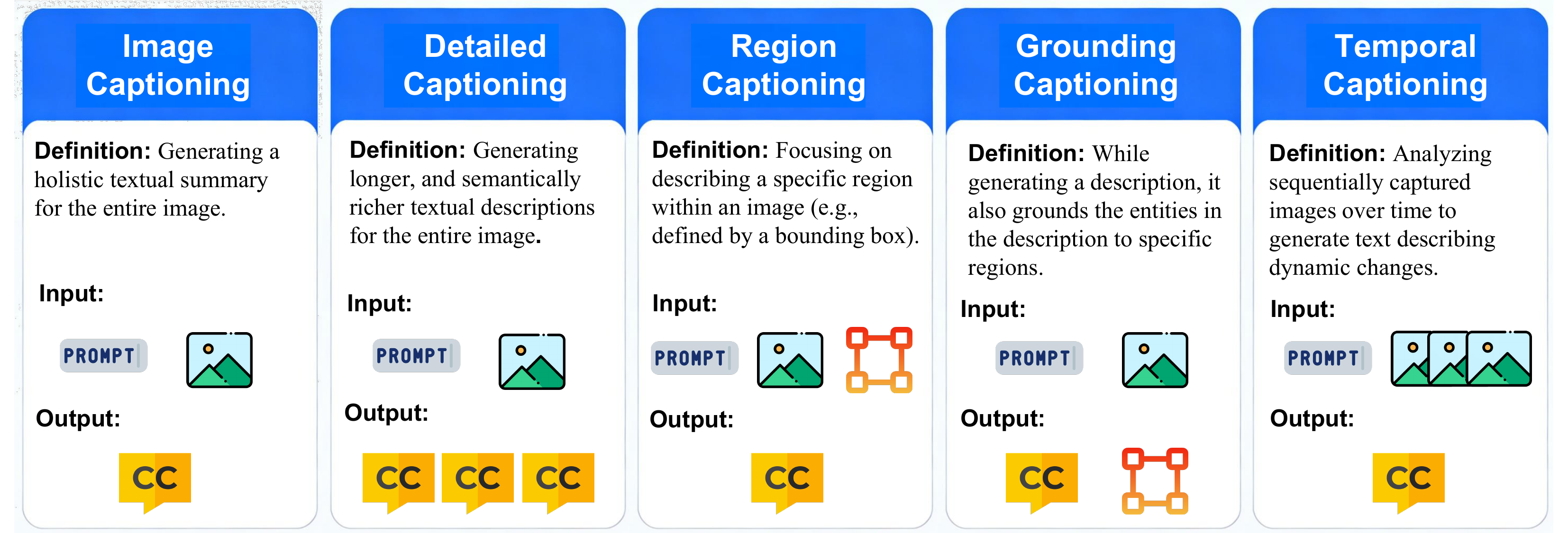}
    \end{center}
    \caption{Rapid Distinction of RS Captioning Tasks. The difficulty level increases with the rightward direction.}
    \label{fig:evo_cap}        
    \vspace{-0.3cm}
\end{figure*}

\section{The Evolution of RS Vision-Language Datasets}
\label{sec:evolution}

This section provides a detailed sequential introduction to five categories of datasets, Image Captioning, VQA, VG, Comprehensive Data, and Benchmark, along with an analysis of their developmental trends. In each subsection, we present the detailed information of the corresponding existing RS vision-language datasets in a table.

\subsection{Evolution of RS Captioning Datasets}
\label{sec:cap}

Image captioning, one of the earliest vision-language multimodal tasks in RS, aims to describe the content of RS images using natural language. Existing RS image captioning datasets can be broadly categorized into five classes, as outlined in the taxonomy of Figure \ref{fig:evo_cap}. Each of these classes will be elaborated on in this section. The development of these datasets, as chronicled in Table \ref{tab:caption_data}, shows a  trajectory from small-scale, single-modality collections to large-scale, diverse, and complex data corpora.

\begin{table*}[!t]
\centering
\setlength{\tabcolsep}{2pt}
\caption{Mainstream RS Image Captioning Datasets. Abbreviations: Multitemporal (MT), Infrared (IR), Near-Infrared (NIR), Multispectral (MS), Digital Elevation Model (DEM), Synthetic Aperture Radar (SAR), Manually Data Pipeline (Manual), Automatically Data Pipeline (Auto), Semi-automatically Data Pipeline (Semi-Auto), Ground Sample Distance (GSD), Novel Annotation Ratio (NAR)}.
\label{tab:caption_data}
\resizebox{\linewidth}{!}{
\begin{tabular}{lcc|ccccc}
\toprule
\textbf{Name} & \textbf{Year} & \textbf{Venue} & \textbf{\#Samples} & \textbf{Modality} & \textbf{GSD (m)} & \textbf{Method}  & \textbf{NAR} \\ 
\midrule
UCM-Captions \cite{qu2016cap} & 2016 & CITS & 2,100 images, 10,500 captions & RGB & 0.3048 & Auto & 1.0 \\
Sydney-Captions \cite{qu2016cap}& 2016 & CITS & 613 images, 3,065 captions & RGB & 0.7 & Auto  & 1.0\\

RSICD \cite{lu2018rsicd} & 2017 & TGRS & 10,921 images, 54,605 captions & RGB & Diverse & Manual  & 1.0\\

RSITMD \cite{yuan2022rsitmd} & 2021 & TGRS & 4,743 images, 23,715 captions & RGB & Diverse & Manual  & 1.0\\

UAV-Captions \cite{genc2022uavcap} & 2021 & TGRS & 2,628 images, 7,884 captions & RGB & 0.02 & Manual  & 1.0\\

NWPU-Captions \cite{cheng2022nwpu-cap} & 2022 & TGRS & 31,500 images, 157,500 captions & RGB & 0.2$\sim$30 & Manual  & 1.0\\

LEVIR-CC \cite{liu2022levir-cc} & 2022 & TGRS & 20,154 images, 50,358 captions & MT & 0.5 & Manual  & 1.0\\

CapERA \cite{bashmal2023capera} & 2022 & RS & 2,864 videos, 1,420 captions & Video & Diverse & Semi-Auto  & 1.0\\

RSICap \cite{hu2023rsgpt} & 2023 & ISPRS & 2,585 captions & RGB & Diverse & Manual  & 1.0\\

LAION-EO \cite{czerkawski2023laion} & 2023 & ICCV & 24,933 captions & Video & Diverse & Auto  & 0.0\\

SkyScript \cite{wang2024skyscript} & 2023 & AAAI & 2.6M captions & RGB & 0.1$\sim$30 & Semi-Auto & 1.0\\

RemoteCLIP \cite{liu2024remoteclip} & 2024 & TGRS & 828,725 captions & RGB & Diverse & Auto & 1.0\\

RS5M \cite{zhang2024rs5m} & 2024 & TGRS & 5M captions & RGB & Diverse & Semi-Auto & 0.4\\

MMM-RS \cite{luo2024mmm} & 2024 & NeurIPS & 2,103,273 captions & RGB,SAR,NIR & Diverse & Semi-Auto & 1.0\\

LuoJiaHOG \cite{zhao2025luojiahog} & 2024 & ISPRS & 94,856 images & RGB & Diverse & Semi-Auto & 1.0\\

MOCO \cite{pan2024moco} & 2024 & Drone & 7,449 images, 37,245 captions & RGB & RGB & Manual & 1.0\\
LRS2M \cite{chen2025dgtrsd} & 2025 & arXiv & 2,023,688 captions & Diverse & Diverse & Auto & 0.5\\

RSTeller \cite{ge2024rsteller} & 2025 & ISPRS & 1,309,926 images, 2,619,852 captions & RGB & 0.6 & Auto & 1.0 \\

GeoLangBind-2M \cite{xiong2025geolangbind} & 2025 & arXiv & 2,050,983 captions & RGB,SAR,MS,DEM,IR & Diverse & Auto & 0.2\\

Git-10M \cite{liu2025text2earth} & 2025 & GRSM & 10.5M captions & RGB & 0.5$\sim$128 & Semi-Auto & 1.0\\

GAIA \cite{zavras2025gaia} & 2025 & arXiv & 41,030 images, 205,150 captions & MS,NIR,SAR,RGB & Diverse & Auto & 1.0\\

CVG-Text \cite{ye2024cross} & 2025 & ICCV & 30,000 coordinate points & RGB & 0.12 & Semi-Auto & 1.0\\

Llama3-SSL4EOS12 \cite{marimo2025ssl4eos} & 2025 & PKDD & 975,000 images, 1M captions & MS & 10 & Semi-Auto & 1.0\\

SPIE \cite{si2025spex} & 2025 & arXiv & 403,432 images & MS & Diverse & Auto & 1.0\\

GLEAM \cite{lu2025gleam} & 2025 & arXiv & 504 images, 202,832 captions & RGB & Diverse & Semi-Auto & 1.0\\

Street2Sat-TEXT \cite{ye2025street2sat} & 2025 & GRSM & 36,000 coordinate points & RGB & Diverse & Semi-Auto & 1.0\\

CVAVT-TEXT \cite{ye2025street2sat} & 2025 & GRSM & 44,000 image pairs & RGB & Diverse & Auto & 1.0\\

SAR-TEXT \cite{he2025sartext} & 2025 & arXiv & 136,584 images & SAR & Diverse & Auto & 1.0\\

RSCC \cite{chen2025rscc} & 2025 & NeurIPS & 124,630 images, 62,315 captions  & MT & Diverse & Semi-Auto & 0.3\\

Landsat30-AU-CAP \cite{ma2025landsat30au} & 2025 & AAAI & 196,262 captions  & RGB & 30 & Semi-Auto & 1.0\\

SARCAP \cite{jiang2025sarcap} & 2025 & ISPRS & 400,000 captions & SAR & 0.1$\sim$20 & Auto & 1.0\\

MMSAR \cite{wang2025sarclip} & 2025 & TGRS & 105,685 images, 304K captions & SAR & Diverse & Auto & 1.0\\

SARVLM-1M  \cite{ma2025sarvlm} & 2025 & arXiv  & 365,984 images, 1,727,408 captions & SAR & Diverse & Auto & 0.0\\
\bottomrule
\end{tabular}

}
\end{table*}

\subsubsection{Captioning \& Retrieval}
\label{sec:cap_sub}

In original image captioning tasks, the model is required to generate corresponding textual descriptions from input images. For cross-modal retrieval tasks, the model must identify the most relevant text descriptions given images (image-to-text retrieval) or retrieve the most matching images based on text queries (text-to-image retrieval). Early RS datasets did not distinguish between these two tasks, with each sample typically comprising a single RS image paired with a concise textual caption. Foundational datasets like UCM-Captions and Sydney-Captions~\cite{qu2016cap}, created via automatic pipelines, and the manually annotated RSICD~\cite{lu2018rsicd} exemplify this category. They established the basic paradigm of single-sentence, image-level description for RGB imagery, with scales ranging from a few thousand to tens of thousands of image-caption pairs.

\subsubsection{Detailed Captioning}
\label{sec:d_cap}

With the increasing adoption of LLMs, the captions generated in new datasets have exhibited significant improvements in both textual length and semantic richness (incorporating multiscale spatial relationships and domain-specific terminology). This evolution reflects the growing complexity of vision language alignment requirements in RS applications. Recent large-scale datasets are pivotal for pre-training robust vision-language models. For instance, RS5M~\cite{zhang2024rs5m} provides 5 million captions, and RSTeller~\cite{ge2024rsteller} offers over 2.6 million image-caption pairs. Furthermore, datasets like GeoLangBind-2M~\cite{xiong2025geolangbind} and MMM-RS~\cite{luo2024mmm} have expanded beyond RGB to include diverse modalities (e.g., SAR, MS, DEM, IR, NIR), capturing richer geospatial semantics. This trend towards massive scale and multi-modality, as seen in SkyScript (2.6M captions)~\cite{wang2024skyscript} and RemoteCLIP~\cite{liu2024remoteclip}, is crucial for developing foundational models with deep domain understanding.

\subsubsection{Region Captioning}
\label{sec:r_cap}

Region captioning distinguishes itself from conventional image captioning by focusing on generating textual descriptions for specific image regions, typically delineated through bounding boxes. This localized approach enables the precise characterization of individual objects or regions of interest, which is particularly valuable for complex visual analysis tasks in RS applications. Datasets supporting this task, e.g, Falcon\_SFT \cite{yao2025falcon} and GeoLLaVA \cite{elgendy2024geollava}, require fine-grained annotations that link phrases or sentences to specific areas within an image. This represents a move towards more structured and object-aware textual descriptions compared to global image captioning.

\subsubsection{Grounding Captioning}
\label{sec:g_cap}

Grounding Captioning represents an advanced form of detailed image captioning that not only generates comprehensive descriptions of visual content, but also provides precise spatial locations (bounding boxes or masks) for each described object in the image. This technique imposes significant challenges, as it requires the model to establish an accurate point-to-point correspondence between textual descriptions and their corresponding visual elements in the image space. This task can be performed at either the region level or the pixel level, depending on whether the target objects are represented by bounding boxes or masks. It is a natural extension of region captioning with an explicit localization requirement. The creation of such datasets is an active research area, e.g, EarthDial-Instruct \cite{soni2024earthdial} and GeoPixelD \cite{shabbir2025geopixel}, and often builds upon the foundational data from large-scale captioning collections.

\subsubsection{Temporal Captioning}
\label{sec:t_cap}

Temporal captioning tasks process sequential image inputs (either multiple still frames or video sequences) to generate coherent descriptions of temporal dynamics. In RS applications, the most prominent implementation is change captioning derived from change detection datasets, which requires models to summarize object-level transformations (e.g., construction progress, land cover changes, or disaster impacts) across temporal observations. This task presents unique challenges in spatiotemporal reasoning. Datasets like LEVIR-CC~\cite{liu2022levir-cc}, a multi-temporal change captioning dataset with over 50k captions, are specifically designed for this purpose. Other resources like CapERA~\cite{bashmal2023capera}, a video captioning dataset, also support modeling temporal sequences. The evolution in this category focuses on capturing what has changed in addition to what is there.

In summary, the development of RS image captioning datasets has evolved from early, small-scale collections with brief, image-level captions for basic tasks towards large-scale, semantically rich corpora that support complex multimodal understanding. This progression is marked by a significant expansion in data volume and caption complexity, driven by advances in foundation models. The focus has shifted to more sophisticated objectives, including fine-grained region description, precise visual grounding, and temporal dynamics analysis. Concurrently, datasets have increasingly incorporated diverse data modalities beyond optical imagery, and annotation methodologies have transitioned from manual efforts to automated, model-assisted processes, collectively enabling deeper and more structured geospatial language-vision alignment.

\begin{figure*}[!t]
	\begin{center}             
        \includegraphics[width=1\linewidth]{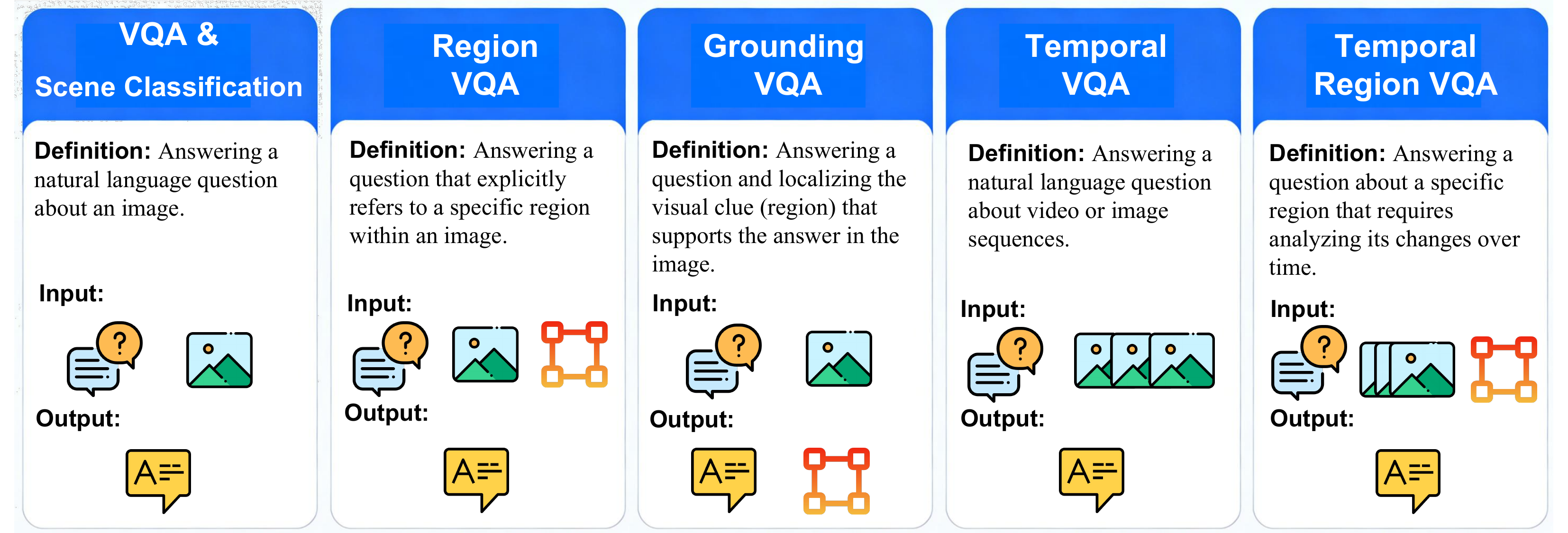}
    \end{center}
    \caption{Rapid Distinction of RS VQA Tasks. The difficulty level increases with the rightward direction.}
    \label{fig:evo_vqa}        
    \vspace{-0.3cm}
\end{figure*}

\begin{table*}[!t]
\centering
\setlength{\tabcolsep}{1pt}
\caption{Mainstream RS VQA Datasets. Abbreviations: Multitemporal (MT), Infrared (IR), Near-Infrared (NIR), Multispectral (MS), Digital Elevation Model (DEM), Synthetic Aperture Radar (SAR), Hyperspectral (HS), Free Format (FRE), Multiple Choice (CHO), True or False (T/F), Manually Data Pipeline (Manual), Automatically Data Pipeline (Auto), Semi-automatically Data Pipeline (Semi-Auto), Ground Sample Distance (GSD), Novel Annotation Ratio (NAR)}.
\label{tab:vqa_data}
\resizebox{\linewidth}{!}{
\begin{tabular}{lcc|cccccc}
\toprule
\textbf{Name} & \textbf{Year} & \textbf{Venue} & \textbf{\#Samples} & \textbf{Modality} & \textbf{GSD (m)} & \textbf{Method} & \textbf{Type} & \textbf{NAR} \\ \midrule
RSVQA-LR \cite{lobry2020rsvqa}& 2020 & TGRS & 772 images, 77,232 Q\&As & RGB & 10 & Auto & FRE, T/F & 1.0\\ 
RSVQA-HR \cite{lobry2020rsvqa}& 2020 & TGRS & 10,659 images, 1,066,316 Q\&As & RGB & 0.15 & Auto & FRE, T/F  & 1.0\\ 
RSVQAxBEN \cite{lobry2021rsvqaxben}& 2021 & IGARSS & 590,326 images, 14,758,150 triples & RGB & Diverse & Auto & FRE, T/F  & 1.0\\ 
FloodNet \cite{maryam2021floodnet}& 2021 & Access & 2,343 images, 4,500 Q\&As & RGB & 0.015 & Manual & FRE, T/F  & 1.0\\ 
RSIVQA \cite{zheng2021mutual}& 2021 & TGRS & 37,264 images, 111,134 Q\&As & RGB & 0.1$\sim$8 & Semi-Auto & T/F  & 0.1\\ 
CDVQA \cite{yuan2022cdvqa}& 2022 & TGRS & 2,968 Q\&As & RGB, MT & 0.5$\sim$3 & Auto & FRE, T/F  & 1.0\\ 
VQA-TextRS \cite{al2022open}& 2022 & IJRS & 2,144 images, 6,245 Q\&As & RGB & 0.3 & Manual & FRE, T/F  & 1.0\\ 
CRSVQA \cite{zhang2023crsvqa}& 2023 & TGRS & 4,639 images, 4,644 Q\&As & RGB & 0.5$\sim$3 & Manual & FRE  & 1.0\\ 
HRVQA \cite{li2024hrvqa}& 2023 & ISPRS & 53,512 images, 1,070,240 Q\&As & RGB & 0.08 & Auto & FRE  & 1.0\\ 

EarthVQA \cite{wang2024earthvqa}& 2023 & AAAI & 6,000 images, 208,593 Q\&As & MS, HS & 0.3$\sim$0.5 & Manual & FRE, T/F  & 0.6\\ 

QAG-360K \cite{li2024qvg}& 2024 & arXiv & 6,810 images, 360K Q\&As & MT & 0.1$\sim$3.0 & Semi-Auto & FRE, T/F  & 1.0\\ 
RemoteCount \cite{liu2024remoteclip}& 2024 & TGRS & 165,745 images, 828,725 Q\&As & RGB & Diverse & Auto & FRE  & 1.0\\ 
OSVQA \cite{zhao2025osvqa}& 2024 & ISPRS & 14,216 images, 1,131,730 Q\&As & SAR, RGB & Diverse & Semi-Auto & CHO, T/F  & 1.0\\ 
GeoLLaVA \cite{elgendy2024geollava}& 2025 & arXiv & 100,000 train, 6,042 test Q\&As & RGB, MS, MT & 0.3$\sim$3.7 & Semi-Auto & FRE  & 1.0\\ 
MME-RealWorld \cite{zhang2024mme}& 2025 & ICLR & 1,298 images, 3,738 Q\&As & RGB & 0.07$\sim$38.22 & Manual & CHO  & 1.0\\ 
LRS-VQA \cite{luo2025large}& 2025 & arXiv & 1,657 images, 7,333 Q\&As & RGB & Diverse & Semi-Auto & FRE, T/F  & 1.0\\ 
AgroMind \cite{li2025agromind}& 2025 & arXiv & 20,850 images, 28,482 Q\&As & RGB, NIR, MS & Diverse & Auto & FRE, T/F, CHO  & 1.0\\ 
AVI-Math \cite{zhou2025avimath}& 2025 & ISPRS & 360 images, 3,773 Q\&As & RGB & Diverse & Semi-Auto & FRE, CHO, T/F  & 1.0\\ 
Earth-Bench \cite{feng2025earthagent}& 2025 & arXiv & 13,729 images, 248 Q\&As & RGB & Diverse & Auto & CHO  & 1.0\\ 
VICot-HRSC \cite{wang2025vicotagent}& 2025 & NerulRS & 364 images, 4,400 Q\&As & RGB & Diverse & Manual & FRE  & 1.0\\ 
GeoPlan-Bench \cite{li2025geoplan}& 2025 & arXiv & 1,244 Q\&As & MS, SAR, RGB, MT & Diverse & Semi-Auto & FRE  & 1.0\\ 
LRS-GRO \cite{liu2025zoomearth}& 2025 & arXiv & 1,224 images,  13,245 Q\&As & RGB & Diverse & Semi-Auto & FRE  & 1.0\\ 
Landsat30-AU-VQA \cite{ma2025landsat30au}& 2025 & AAAI & 17,725 Q\&As & RGB & 30 & Semi-Auto & CHO  & 1.0\\ 
CitySet \cite{wang2026cityset}& 2025 & ISPRS & 20,589,110, 800K Q\&As & RGB & 0.5 & Manual & FRE  & 1.0\\ 
UnivEARTH \cite{kao2025univearth}& 2025 & ICML & 140 Q\&As & Diverse & RGB & Semi-Auto & T/F  & 1.0\\ 

Geo3DVQA \cite{tsujimoto2025geo3dvqa}& 2025 & arXiv & 110K Q\&As & RGB & Diverse & Auto & FRE, CHO  & 1.0\\ 
UrbanVideo-Bench \cite{zhao2025urbanvideobench}& 2025 & arXiv & 5.2K Q\&As & Video & Diverse & Semi-Auto & CHO  & 1.0\\ 
BEDI \cite{guo2025bedi}& 2025 & arXiv & 185 images, 3,097 Q\&As & RGB, Video & Diverse & Manual & CHO, FRE  & 1.0\\ 
KnowFlow-Bench \cite{chen2025canglingknowflow} & 2025 & arXiv & 324 real RS workflows & RGB & Diverse & Auto & FRE  & 1.0\\
MM-UAVBench \cite{dai2025mmuavbench} & 2025 & arXiv & 1.5K videos, 2.8K images, 5.7K instances & RGB, Video & Diverse & Manual & CHO  & 0.8\\
LRS-VQA-Zoom \cite{li2025rscovlm} & 2025 & arXiv & 302K Q\&As & RGB & Diverse & Semi-Auto & FRE, CHO  & 1.0\\
\bottomrule
\end{tabular}
}
\end{table*}

\subsection{Evolution of RS VQA Datasets}
\label{sec:vqa}

VQA is one of the most common tasks in vision-language multimodal research. It is generally categorized into three types: multiple-choice, true/false (binary), and open-ended questions. Among these, multiple-choice and true/false questions are objective and can be evaluated using accuracy as the primary metric. In contrast, open-ended questions often have non-unique answers, necessitating a variety of evaluation methods. For short-form answers consisting of only one or two words, content-based accuracy can be used, i.e., the prediction is considered correct if it includes any ground-truth word; otherwise, it is marked incorrect. In addition, synonym-based similarity metrics, such as those used in Next-QA \cite{xiao2021next}, can be employed to account for semantic equivalence. However, for longer open-ended responses, these metrics become less reliable. In such cases, it is common to resort to evaluation using LLMs \cite{hu2023rsgpt} or human judges \cite{pang2024vhm} to assess the quality and relevance of the generated answers. We categorize existing RS VQA datasets into five primary types, which are described in detail below. This categorization aligns with the taxonomy of vision-language data shown in Figure \ref{fig:evo_vqa}. The mainstream datasets for these categories are summarized in Table \ref{tab:vqa_data}, showing a clear evolution in scale, modality, and task complexity over the years. Below, we describe each category in detail, referencing key datasets from the table.

\subsubsection{Scene Classification}
\label{sec:class}

Scene classification is a domain-specific task in RS, aimed at identifying the land cover type or land use category of an entire image. In the context of vision-language data, scene classification datasets can be viewed as a special form of VQA, where each image is associated with a predefined question—typically fixed across the dataset (e.g., What is the scene type?) —and a corresponding answer label. Unlike general-purpose VQA tasks that involve diverse and image-specific questions, RS scene classification VQA is characterized by its standardized question format and categorical answer space. This structure enables the integration of scene classification into multimodal learning frameworks under the vision-language paradigm. While not all datasets in Table II are exclusively for scene classification, several early and foundational datasets, such as the RSVQA series (RSVQA-LR/HR) \cite{lobry2020rsvqa} and RSIVQA \cite{zheng2021mutual}, incorporate true/false questions that can be framed as binary classification queries about high-level scene attributes (e.g., Is there a residential area?). This serves as a bridge connecting traditional scene classification to the VQA paradigm.

\subsubsection{VQA}
\label{sec:vqa_sub}

In RS image-level VQA tasks, each data sample typically consists of an image, a question, and a corresponding answer. This encompasses the most common form of VQA, dealing with questions about the image content without explicit region reference. As shown in Table II, a vast majority of datasets focus on this task. The field has evolved significantly from early datasets like RSVQA-LR (77k Q\&As) \cite{lobry2020rsvqa} to recent large-scale ones such as HRVQA (1.07M Q\&As) \cite{li2024hrvqa} and OSVQA (1.13M Q\&As) \cite{zhao2025osvqa}, demonstrating a clear trend towards increased data volume. The questions are typically free-form (FRE), true/false (T/F), or multiple-choice (CHO). Another notable trend is the expansion of data modalities, from primarily RGB in early works to incorporating multispectral (MS), synthetic aperture radar (SAR), and multi-temporal (MT) data in datasets such as EarthVQA \cite{wang2024earthvqa}, OSVQA \cite{zhao2025osvqa}, and CDVQA \cite{yuan2022cdvqa}. Automated and semi-automated data generation methods have been crucial in scaling up dataset sizes, as seen in RSVQAxBEN (14.76M triples) \cite{lobry2021rsvqaxben}. A prominent and challenging subtype within image-level VQA is counting, which involves estimating the number of objects (e.g., vehicles, airplanes) in a scene. Dedicated datasets like RemoteCount (828k Q\&As) \cite{liu2024remoteclip} have been created for this purpose. Since the answers are integers, counting is considered a form of short-form open-ended VQA. However, standard accuracy is an inadequate metric for evaluation, as it treats all incorrect counts equally, failing to reflect the magnitude of error. Therefore, regression-based metrics like Root Mean Squared Error (RMSE) and Mean Absolute Error (MAE) are more appropriate, as used in REO-VLM \cite{xue2024reo}.

\subsubsection{Region VQA}
\label{sec:r_vqa}
Region-level VQA differs from image-level VQA in that the question explicitly refers to a specific region within the image. This explicit reference can be conveyed in two primary ways: (1) through natural language questions that include coordinate references in textual form \cite{yao2025falcon}, or (2) via visual prompts such as bounding boxes or markers overlaid on the image \cite{zhang2025earthmaker}. The explicit specification of regions acts as a strong visual cue, which can mitigate the referential ambiguity present in standard image-level VQA. Recent datasets are increasingly incorporating this capability. For instance, GeoLLaVA \cite{elgendy2024geollava} and GeoPlan-Bench \cite{li2025geoplan} are constructed to support region-level understanding and reasoning, where questions or instructions are tied to specific spatial areas within the RS imagery, enabling more precise and interpretable queries.

\subsubsection{Grounding VQA}
\label{sec:g_vqa}
Grounding VQA refers to a task that combines VG and VQA. The model is required not only to provide an answer to a given question but also to localize the corresponding answer region within the image, as illustrated in Figure \ref{fig:evo_vqa}. Compared to conventional VQA, this imposes greater demands on the model, as it must both infer the correct answer and accurately identify the specific image region (bounding boxes) that supports it. Models trained on such data exhibit enhanced interpretability, as the localized regions provide visual evidence. In RS, datasets such as LRS-GRO \cite{liu2025zoomearth} and QAG-360K \cite{li2024qvg} are designed explicitly for this joint task, requiring the model to generate a bounding box or mask corresponding to the answer for a given question. This represents a more advanced and spatially-grounded form of VQA.

\subsubsection{Temporal VQA}
\label{sec:t_vqa}
With the growing capability of vision-language models to process sequential data, temporal VQA tasks have emerged. These tasks take multiple images (often forming a video or a temporal sequence) as input and require models to derive answers by analyzing temporal relationships, changes, or dynamics. In RS, most existing temporal VQA datasets are constructed from change detection data, focusing on queries about pre- and post-disaster comparisons or urban development. Reflecting this trend, recent datasets begin to incorporate temporal or video data. For example, CDVQA \cite{yuan2022cdvqa} uses bi-temporal images for change-related Q\&A. More recent datasets, such as BEDI \cite{guo2025bedi}, are built upon aerial video imagery, requiring mathematical reasoning over dynamic scenes. Similarly, UrbanVideo-Bench \cite{zhao2025urbanvideobench} and MM-UAVBench \cite{dai2025mmuavbench} utilize video data to support tasks that require temporal understanding, such as tracking and action recognition, thereby pushing the boundaries of RS VQA from static imagery to spatiotemporal analysis.

Overall, the development of RS VQA datasets is characterized by increased scale, richer modalities, and greater task complexity. The field has progressed from foundational, manually-constructed datasets to large-scale, automatically generated benchmarks, as evidenced by the growth in the number of samples. There is a significant diversification in data modalities, expanding from standard RGB to include multispectral, SAR, and video data. Concurrently, the VQA task formulations have evolved beyond basic image-level question-answering to incorporate more complex types such as region-referenced, grounded, and temporal VQA.

\begin{figure*}[!t]
	\begin{center}             
        \includegraphics[width=1\linewidth]{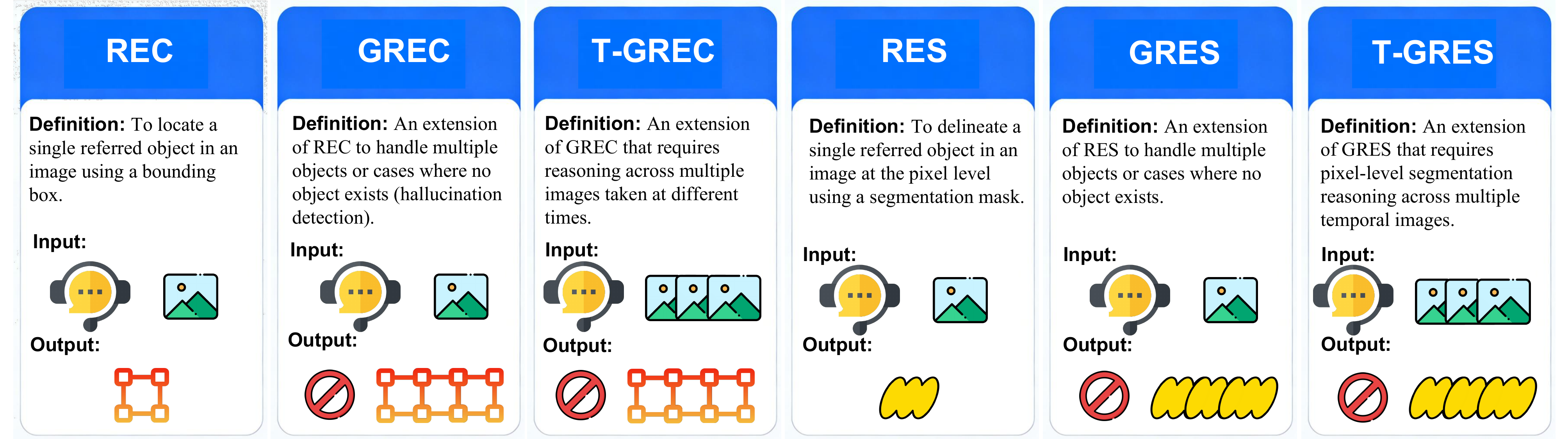}
    \end{center}
    \caption{Rapid Distinction of RS VG Tasks. The three tasks on the left operate at the region level, while those on the right are at the pixel level, with difficulty increasing progressively from left to right on both sides.}
    \label{fig:evo_vg}        
    \vspace{-0.3cm}
\end{figure*}

\begin{table*}[!t]
\centering
\setlength{\tabcolsep}{6pt}
\caption{Mainstream RS VG Datasets. Abbreviations: Horizontal Bounding Box (HBB), Segmentation Mask (Mask), Manually Data Pipeline (Manual), Automatically Data Pipeline (Auto), Semi-automatically Data Pipeline (Semi-Auto), Ground Sample Distance (GSD), Novel Annotation Ratio (NAR)}.
\label{tab:vg_data}
\resizebox{\linewidth}{!}{
\begin{tabular}{lcc|cccccc}
\toprule
\textbf{Name} & \textbf{Year} & \textbf{Venue} & \textbf{\#Samples} & \textbf{Modality} & \textbf{GSD (m)} & \textbf{Method} & \textbf{Type}  & \textbf{NAR} \\ 
\midrule
RSVG \cite{sun2022rsvg}& 2022 & MM & 4,329 images, 7,933 refers  & RGB & Diverse & Semi-Auto & HBB & 1.0\\

DIOR-RSVG \cite{zhan2023dior-rsvg}& 2022 & TGRS & 17,402 images, 38,320 refers & RGB & 0.5$\sim$30 & Semi-Auto & HBB  & 1.0\\

GeoText-1652 \cite{chu2024towards}& 2023 & ECCV & 316,335 refers & RGB & Diverse & Semi-Auto & HBB  & 1.0\\

RSVG-HR \cite{lan2024lqvg}& 2024 & TGRS & 2,650 refers & RGB & Diverse & Manual & HBB  & 1.0\\

OPT-RSVG \cite{li2024opt-rsvg}& 2024 & TGRS & 25,452 images, 48,952 refers & RGB & Diverse & Semi-Auto & HBB  & 1.0\\

RefSegRS \cite{yuan2024rrsis}& 2024 & TGRS & 4,420 refers across 285 scenes & RGB & 0.13 & Auto & Mask  & 1.0\\

RRSIS-D \cite{liu2024rotated}& 2024 & CVPR & 17,402 refers & RGB & 0.5$\sim$30 & Semi-Auto & Mask  & 1.0\\

AAVG \cite{zhou2024geoground}& 2025 & arXiv & 26,465 refers & RGB & 0.007$\sim$0.04 & Auto & Mask  & 1.0\\

RISBench \cite{dong2024risbench}& 2025 & arXiv & 52,472 refers & RGB & 0.1$\sim$30 & Semi-Auto & Mask  & 1.0\\

AerialVG \cite{liu2025aerialvg}& 2025 & ICCV & 50,000 images, 50,000 refers & RGB & Diverse & Manual & HBB  & 1.0\\

EarthReason \cite{li2025segearth}& 2025 & arXiv & 5,434 masks, 30,000 refers & RGB & 0.5$\sim$153 & Semi-Auto & Mask  & 1.0\\

GRES \cite{quenum2025lisat}& 2025 & arXiv & 9,205 images, 1M refers & RGB & 0.3 & Semi-Auto & Mask  & 1.0\\

RefDrone \cite{sun2025refdrone}& 2025 & arXiv & 8,536 images, 17,900 refers & RGB & Diverse & Semi-Auto & HBB  & 1.0\\

RemoteSAM-270K \cite{yao2025remotesam}& 2025 & MM & 270K refers & RGB & Diverse & Auto & Mask  & 0.7\\

NWPU-Refer \cite{yang2025nwpurefer}& 2025 & arXiv & 15,003 images, 49,745 refers & RGB & 0.12$\sim$0.5 & Manual & Mask  & 1.0\\

GeoSeg-1M \cite{ni2025unigeoseg}& 2025 & arXiv & 590K images, 1.1M refers & RGB & 0.05$\sim$153 & Auto & Mask  & 1.0\\

RIS-LAD \cite{ye2025rislad}& 2025 & AAAI & 13,871 refers & RGB & Diverse & Semi-Auto & Mask  & 1.0\\

GRASP-1K \cite{jiang2025grasp}& 2025 & arXiv & 1,071 refers & RGB & 0.08$\sim$30 & Semi-Auto & Mask  & 1.0\\

LaSeRS \cite{xin2025segearthr2}& 2025 & arXiv & 40K masks, 30K refers & RGB & Diverse & Semi-Auto & Mask  & 1.0\\
\bottomrule
\end{tabular}

}
\end{table*}

\subsection{Evolution of RS Visual Grounding Datasets}
\label{sec:vg}

RS VG aims to use natural language expressions to locate specific objects (in the form of bounding boxes or segmentation masks) in RS images, thereby enhancing human interaction with intelligent RS interpretation systems. As shown in Figure \ref{fig:evo_vg}, the evolution of datasets has been pivotal in driving progress in this field. In practical applications, different objects and scenarios necessitate different grounding representations: an HBB can coarsely localize an object's position, and a mask precisely depicts its shape. As summarized in Table~\ref{tab:vg_data}, a variety of large-scale RS visual grounding datasets have emerged recently, spanning from 2022 to 2025. Early research primarily focused on HBBs, as seen in datasets like RSVG~\cite{sun2022rsvg} and DIOR-RSVG~\cite{zhan2023dior-rsvg}. However, as the field advances, more diverse grounding types are being adopted to meet complex application needs. Newer datasets increasingly support segmentation masks, such as RefSegRS~\cite{yuan2024rrsis}, RRSIS-D~\cite{liu2024rotated}, and the large-scale GeoSeg-1M~\cite{ni2025unigeoseg}. This shift reflects the growing demand for more precise, pixel-level localization. The annotation methodologies have also evolved, ranging from manual and semi-automatic to fully automatic data pipelines, significantly improving annotation efficiency and scale. Concurrently, the diversity in GSD and the sheer volume of samples in these datasets have greatly enhanced the robustness and applicability of the developed models.

\subsubsection{Referring Expression Comprehension (REC)}
\label{sec:rec}

The input to REC is an RS image paired with a referring expression, and the output is a single corresponding bounding box of the uniquely referred object in the image. Most early datasets (e.g., RSVG~\cite{sun2022rsvg}, DIOR-RSVG~\cite{zhan2023dior-rsvg}, and GeoText-1652~\cite{chu2024towards}) typically describe objects using their relative positions within the image, as well as appearance attributes such as color, size, and shape. Subsequently, AVVG \cite{zhou2024geoground} has shifted its focus toward referential description based on 3D spatial locations and fine-grained attributes of vehicles. This evolution addresses the need for precise object localization and identification in complex urban environments. For the REC task, the IoU between the predicted bounding box and the ground truth box is directly employed as the evaluation metric. Alternatively, a threshold (e.g., IoU \> 0.5) can be set to compute accuracy, where a prediction is correct if its IoU exceeds the threshold.

\subsubsection{Generalized REC (GREC)}
\label{sec:grec}
GREC extends REC to multi-object localization and hallucination detection, where no referred object exists. This formulation is more aligned with real-world applications. GeoChat-Instruct \cite{kuckreja2024geochat} represents an early example of a GREC dataset in the RS domain. Notably, conventional RS object detection can be viewed as a specialized GREC task where the referring expressions are predefined object categories. Common object detection metrics, such as Mean Average Precision (mAP), require confidence scores, which many MLLMs do not provide natively. To address this, LMMRotate~\cite{li2025simple} introduces mAP$_{nc}$ and mF$_1$ metrics designed for models without confidence scores, which are recommended for evaluating GREC. The development of GREC reflects the field's progression towards handling more complex and realistic scenarios, where models must not only identify relevant objects but also recognize when no object matches the description, thereby reducing false positives in practical applications.

\subsubsection{Temporal GREC}
\label{sec:tgrec}

T-GREC extends the input from a single image to multi-image sequences. A canonical example is change detection, where given bi-temporal images, the model must localize changed regions by reasoning across time. This demands spatio-temporal understanding, making it more challenging than standard GREC. To our knowledge, TEOChatlas \cite{irvin2024teochat} is the first and only RS dataset designed to support this specific task. This task requires the model to understand both spatial relationships within individual images and temporal dynamics across image sequences, enabling applications such as urban development monitoring, disaster assessment, and agricultural growth tracking. The complexity of T-GREC lies in the need for joint spatial-temporal reasoning, where models must distinguish actual changes from seasonal variations, illumination differences, and other confounding factors.

\subsubsection{Referring Expression Segmentation (RES)}
\label{sec:res}
Similar to REC, RES aims to localize a uniquely referred object, but it outputs a precise segmentation mask instead of a bounding box. The development of datasets supporting mask output, as shown in Figure \ref{fig:evo_vg}, has facilitated research in this direction. Representative datasets include RefSegRS~\cite{yuan2024rrsis}, AAVG~\cite{zhou2024geoground}, and RRSIS-D~\cite{liu2024rotated}. Influenced by the reasoning segmentation tasks \cite{lai2023lisa} in the field of general MLLMs, research in this domain has begun to exhibit a marked shift from simple explicit reference towards tasks that require reasoning for implicit reference, with notable examples including GRASP-1K \cite{jiang2025grasp}, and GeoSeg-1M \cite{ni2025unigeoseg}. Evaluation can use an IoU-based accuracy threshold. Standard segmentation metrics such as mean IoU and overall IoU are also applicable for a comprehensive assessment, where IoU is calculated between masks. The transition from bounding boxes to segmentation masks represents a significant advancement in localization precision, enabling applications that require detailed shape information, such as building footprint extraction, crop boundary delineation, and infrastructure monitoring.

\subsubsection{Generalized RES (GRES)}
\label{sec:gres}

GRES extends RES to support multiple objects and empty referrals, corresponding to a more generalized node in the task taxonomy. Analogous to how object detection is a form of GREC, RS semantic segmentation can be viewed as a specialized GRES task. The recently proposed RemoteSAM-270K dataset~\cite{yao2025remotesam} is a representative benchmark for this task. Since it involves potentially multiple disconnected regions, it generalizes beyond the single-mask output of RES. GRES addresses the practical need to localize multiple objects described in complex natural language expressions, which is essential for applications like inventory counting, damage assessment, and environmental monitoring. The development of GRES datasets and methods reflects the field's growing maturity in handling more sophisticated referring expressions that may involve multiple objects, complex relationships, and potentially no relevant objects in the image.

\subsubsection{Temporal GRES}
\label{sec:tgres}
T-GRES advances the paradigm by requiring segmentation outputs from multiple input images. The primary application is in change detection, where the goal is to segment the changed regions between images taken at different times. This task combines temporal reasoning with precise mask prediction, presenting greater complexity than T-GRES, which only outputs bounding boxes for changes. Datasets ChangeChat-87K \cite{deng2025changechat} and Falcon\_SFT \cite{yao2025falcon} are currently identified as the two primary benchmarks supporting the T-GRES task.  This task requires models to perform fine-grained pixel-level segmentation of changes while ignoring irrelevant variations caused by factors like illumination, season, or sensor characteristics. The emergence of T-GRES represents the cutting edge of RS visual grounding, integrating temporal analysis with precise segmentation to support applications such as monitoring construction progress, assessing disaster impacts, and tracking land use changes over time.

In summary, the field of RS VG has undergone rapid evolution. As evidenced by the survey of mainstream datasets, there is a clear trend from early models using horizontal bounding boxes towards recent, more precise segmentation masks, alongside advancements in annotation methods from manual to automatic pipelines. Correspondingly, the task taxonomy has expanded from the foundational REC to include more complex and practical scenarios, such as generalized, temporal, and segmentation-based variants, reflecting the growing sophistication and breadth of application in the research domain.

\begin{table*}[!t]
\centering
\setlength{\tabcolsep}{2pt}
\caption{Mainstream RS Comprehensive Data. Abbreviations: Multitemporal (MT), Infrared (IR),  Multispectral (MS), Synthetic Aperture Radar (SAR), Manually Data Pipeline (Manual), Automatically Data Pipeline (Auto), Semi-automatically Data Pipeline (Semi-Auto), Ground Sample Distance (GSD), Novel Annotation Ratio (NAR)}.
\label{tab:com_data}
\resizebox{\linewidth}{!}{
\begin{tabular}{lcc|ccccc}
\toprule
\textbf{Name} & \textbf{Year} & \textbf{Venue} & \textbf{\#Samples} & \textbf{Modality} & \textbf{Method} & \textbf{GSD (m)} & \textbf{NAR} \\ 
\midrule
GeoChat-Instruct \cite{kuckreja2024geochat}& 2023  & CVPR & 318,000 instructions & RGB & Auto & Diverse & 1.0\\

MMRS-1M \cite{zhang2024earthgpt} & 2024 & TGRS & 1M instructions & RGB, SAR, IR & Auto & Diverse & 0.0\\
VRSBench-Train \cite{li2024vrsbench}& 2024  & NeurIPS & 142K instructions  & RGB & Semi-Auto & Diverse & 1.0\\
RS-GPT4V \cite{xu2024rs}& 2024  & arXiv & 107,936 images & RGB & Semi-Auto & Diverse & 1.0\\
RSVP \cite{zhang2025earthmaker}& 2024  & TGRS & 3.65M instructions & RGB & Auto & Diverse & 1.0\\
LHRS-Align/Instruct \cite{muhtar2024lhrs}& 2024  & ECCV & 1.1898M instructions & RGB & Auto & Diverse & 1.0\\
FIT-RS \cite{luo2024skysensegpt}& 2024  & arXiv & 1,800,851 instructions & RGB, IR, NIR, HS & Auto & Diverse & 1.0\\
VersaD/HnstD/VariousRS \cite{pang2024vhm}& 2024  & AAAI & 1.5214M instructions & RGB & Semi-Auto & Diverse & 1.0\\
LHRS-Align-Recap/Instruct \cite{li2025lhrsbotnova}& 2024  & ISPRS  & 1.18M instructions & RGB & Auto & Diverse & 1.0\\
ChangeChat-87K \cite{deng2025changechat}& 2024 & ICASSP  & 87,195 instructions & MT & Semi-Auto & Diverse & 0.4\\
EarthDial-Instruct \cite{soni2024earthdial}& 2025  & CVPR & 11.11M instructions & RGB, SAR, MS & Auto &  Diverse & 1.0\\
SkyEye-968K \cite{zhan2025skyeyegpt}& 2025  & ISPRS & 968K instructions & RGB & Semi-Auto & Diverse & 0.03\\
DDFAV \cite{li2025ddfav}& 2025  & RS & 1,755 images, 27,650 instructions & RGB & Semi-Auto & Diverse & 1.0\\
GeoPixelD \cite{shabbir2025geopixel} & 2025  & ICML & 53,816 instructions & RGB & Semi-Auto & Diverse & 1.0\\
Falcon\_SFT \cite{yao2025falcon}& 2025  & arXiv & 5.6M images, 78M instructions & RGB, MS, SAR & Auto & Diverse & 1.0\\

TEOChatlas \cite{irvin2024teochat}& 2025  & ICLR & 1,244,393 images, 554,071 instructions & MT & Auto & Diverse & 0.6\\
GeoPixInstruct \cite{ou2025geopix}& 2025  & GRSM & 65,463 images, 140,412 instructions & RGB & Semi-Auto & Diverse & 0.7\\
SARLANG-1M \cite{wei2025sarlang}& 2025  & ICLR & 118,331 images, 1,126,277 instructions & SAR & Semi-Auto & 0.1$\sim$25 & 1.0\\
M-RSVP \cite{zhang2025earthgptx}& 2025  & arXiv & 650,000 instructions & RGB, SAR, IR & Semi-Auto & Diverse & 1.0\\
RS-VL3M \cite{hu2025ringmoagent}& 2025  & arXiv & 3M instructions & SAR, IR, RGB & Auto & Diverse & 1.0\\

FUSAR-GEOVL-1M \cite{yang2025fusarklip}& 2025  & arXiv & 120K images, 1M instructions & SAR & Auto & 0.2$\sim$3 & 1.0\\
ScoreRS \cite{muhtar2025scorers}& 2025  & NeurIPS & 131K instructions & RGB & Semi-Auto & Diverse & 1.0\\
DisasterM3 \cite{wang2025disasterm3}& 2025  & NeurIPS & 26,988 image pairs & RGB, SAR & Semi-Auto & Diverse & 1.0\\
RS-Eot-4K \cite{shao2025socrates}& 2025 & arXiv & 4.3K instructions & RGB, SAR, IR & Auto & Diverse & 1.0\\ 
\bottomrule
\end{tabular}
}
\end{table*}

\begin{table*}[!t]
\centering
\setlength{\tabcolsep}{3pt}
\caption{Mainstream RS Comprehensive Benchmarks. Abbreviations: Multitemporal (MT), Multispectral (MS), Synthetic Aperture Radar (SAR), Bitemporal (BT), Panchromatic (PAN), Manually Data Pipeline (Manual), Automatically Data Pipeline (Auto), Semi-automatically Data Pipeline (Semi-Auto), Ground Sample Distance (GSD), Novel Annotation Ratio (NAR)}.
\label{tab:com_bmk}
\resizebox{\linewidth}{!}{
\begin{tabular}{lcc|ccccc}
\toprule
\textbf{Name} & \textbf{Year} & \textbf{Venue} & \textbf{\#Samples} & \textbf{Modality} & \textbf{Method} & \textbf{GSD (m)} & \textbf{NAR} \\ 
\midrule
RSIEval \cite{hu2023rsgpt}& 2023 & ISPRS & 100 images, 100 captions, 936 Q\&As & RGB & Manual & Diverse & 1.0\\ 
GeoChat-Bench \cite{kuckreja2024geochat}& 2023 & CVPR & 105K images, 318K instructions & RGB, MS & Auto & Diverse  & 1.0\\ 
LHRS-Bench \cite{muhtar2024lhrs}& 2024 & ECCV & 108 questions, 690 Q\&As & MS, PAN & Manual & Diverse  & 1.0\\ 
GEOBench-VLM \cite{danish2024geobench}& 2025 & ICCV &  10,000 instructions & RGB, MS, SAR, BT, MT & Semi-Auto & Diverse  & 1.0\\ 
UrBench \cite{zhou2024urbench}& 2025 & AAAI & 11,600 instructions & RGB & Semi-Auto & Diverse  & 1.0\\ 
XLRS-Bench \cite{wang2025xlrs}& 2025 & CVPR & 1,400 images, 91K instructions & RGB & Semi-Auto & Diverse  & 1.0\\ 
CHOICE \cite{an2025choice}& 2025 & NeurIPS & 10,507 instructions & RGB, MT, BT, MI & Semi-Auto & 0.1$\sim$30  & 1.0\\ 
AirSpatial\cite{zhou2025airspatialbot}& 2025 & TGRS & 14K refers, 17K Q\&As  & RGB & Auto & 0.007$\sim$0.04  & 1.0\\ 
SARChat-Bench-2M \cite{ma2025sarchat}& 2025 & arXiv & 2,063,548 instructions & SAR & Auto & 0.3$\sim$10  & 1.0\\ 
REOBench \cite{li2025reobench} & 2025 & arXiv & 37,400 instructions & RGB & Auto & Diverse  & 1.0\\ 
ThinkGeo \cite{shabbir2025thinkgeo}& 2025 & arXiv & 486 tasks, 1773 instructions & RGB, SAR & Semi-Auto & 0.015$\sim$15  & 1.0\\ 
A2Seek \cite{mo2025a2seek}& 2025 & arXiv & 542 videos & Video, RGB, IR & Semi-Auto & Diverse  & 1.0\\ 
\bottomrule
\end{tabular}
}
\end{table*}

\subsection{Evolution of Comprehensive Datasets}
\label{sec:com}

Beyond the aforementioned single-task RS vision-language datasets, numerous studies have proposed comprehensive datasets and benchmarks supporting multiple tasks to enhance the overall capabilities of MLLMs, as exemplified in Table \ref{tab:com_data} and Table \ref{tab:com_bmk}. Training data are undergoing explosive growth in scale, rapidly escalating from hundreds of thousands of instructions in 2023 to tens of millions (e.g., 11.11 million instructions in EarthDial-Instruct) and even nearly a hundred million (e.g., 78 million instructions in Falcon\_SFT) by 2025. The construction methodologies heavily rely on automated pipelines to support this massive scale. Concurrently, data modalities are rapidly diversifying from the early dominance of RGB, with SAR, infrared, multispectral, and even hyperspectral data being widely integrated, fostering specialized datasets for domains like change detection and disaster assessment .

Currently, RS vision-language data stands on the cusp of a transition from quantitative accumulation to qualitative breakthrough. Its development trajectory is clear: continuous expansion in scale, increasing modality fusion, and more specialized tasks. Future evolution directions are even more defined: First, a balanced emphasis on both scale expansion and quality improvement, enhancing the complexity, diversity, and accuracy of instructions while shifting the paradigm from "scale-first" to "quality-and-efficiency parity." Second, deep multimodal fusion, moving beyond simple juxtaposition of data types towards pursuing cross-modal alignment and synergistic understanding, enabling complementary use of multi-source data in complex tasks like disaster response and land cover change detection. Third, task design is trending towards complexity and practicality, evolving from basic description and Q\&A to comprehensive tasks requiring spatiotemporal reasoning, causal analysis, and decision support.

These trends collectively propel comprehensive datasets to become the cornerstone for training the next generation of powerful and general-purpose RS MLLMs. The core of future evolution will shift from pursuing sheer data volume towards building a new generation of data infrastructure characterized by high quality, deep fusion, strong reasoning capabilities, and effective evaluation. This necessitates closer collaboration among data builders, algorithm researchers, and domain experts to jointly forge a next-generation data system capable of empowering large RS models to achieve deep visual understanding, complex spatiotemporal reasoning, and reliable geospatial decision-making.

\begin{figure*}[!t]
	\begin{center}             
        \includegraphics[width=1\linewidth]{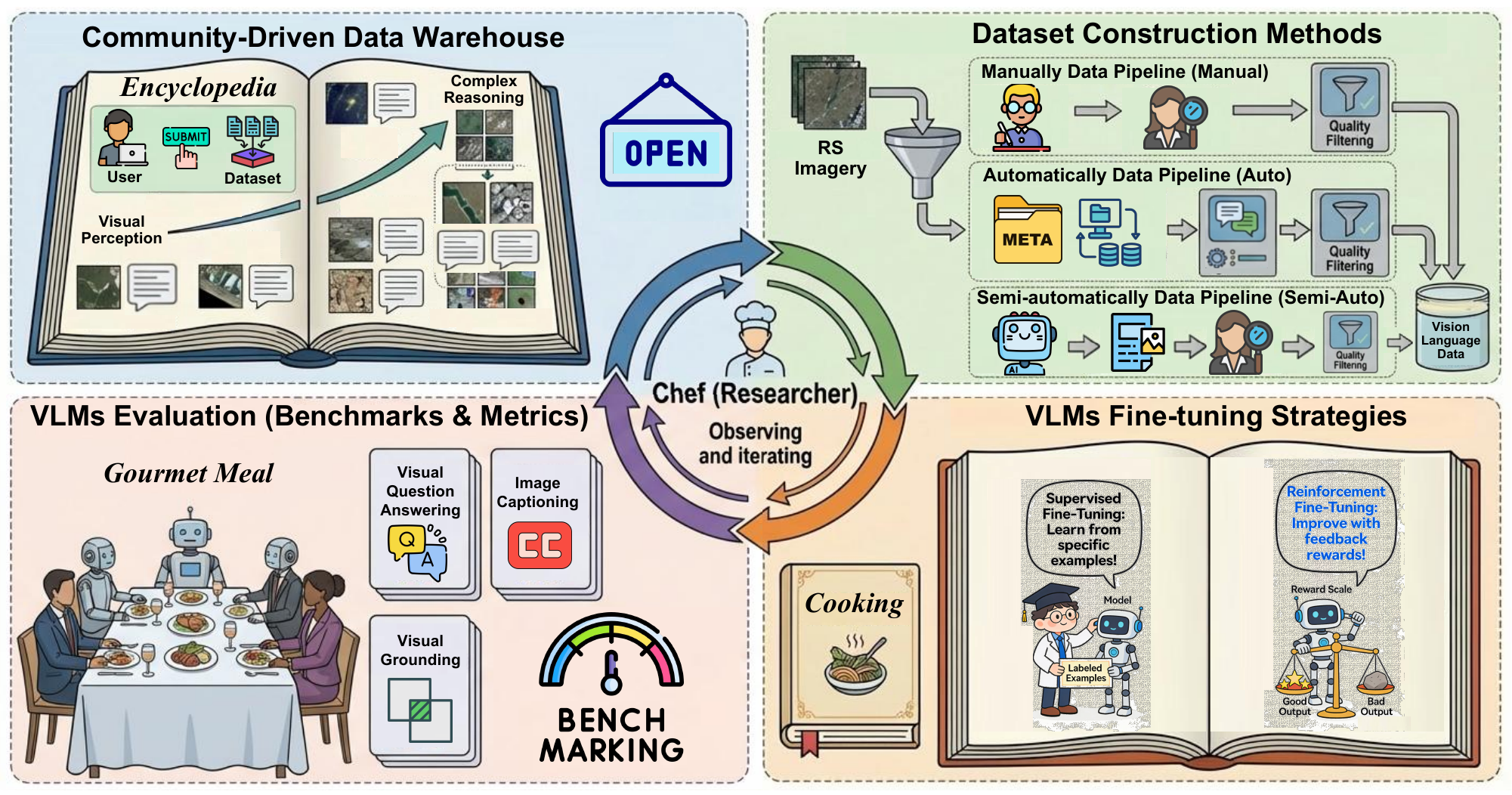}
    \end{center}
    \caption{A Practical Guide to Data Choice, Curation, Fine-Tuning, and Evaluation. With GeoChef, researchers can master an iterative observe-refine cycle to train models that excel in practical RS scenarios. }
    \label{fig:main_4step}        
\end{figure*}

\subsection{Demarcation Between Atomic and Composite Datasets}

With the rapid advancement of MLLMs, a growing number of composite datasets have been introduced, which are constructed by aggregating existing image-text datasets and reformatting them into standardized instruction-following formats. In contrast to atomic datasets, which are built entirely with de novo novel annotations from scratch, composite datasets represent a distinctive data paradigm unique to the era of MLLM.

For image captioning datasets, LAION-EO \cite{czerkawski2023laion} is constructed by filtering RS images from general image-text pairs, resulting in a Novel Annotation Ratio (NAR) of 0; it is a purely composite dataset. For the LRS2M \cite{chen2025dgtrsd} dataset, half of its samples are obtained by integrating existing datasets, while the other half are generated via model-based augmentation from short captions, giving it a NAR of 0.5. For VQA datasets, a large number of questions in the RSIVQA \cite{zheng2021mutual} task are generated via template-based transformation—for example, by directly converting RS image classification tasks into VQA tasks—resulting in a NAR of only 0.1. For VG datasets, 30\% of the samples in RemoteSAM-270K \cite{yao2025remotesam} are derived from the integration of existing datasets, yielding a NAR of 0.7. For comprehensive multi-modal datasets, the MMRS-1M \cite{zhang2024earthgpt} and SkyEye-986k \cite{zhan2025skyeyegpt} datasets are constructed almost entirely through the integration or transformation of existing datasets. In contrast, all corresponding comprehensive benchmarks have a NAR of 1.

Notably, these two dataset categories are neither mutually exclusive nor mutually antagonistic; instead, they constitute a synergistic, complementary ecosystem that underpins the full development lifecycle of RS VLM. When evaluated solely on their intrinsic contribution at the dataset level, composite datasets do not outperform atomic datasets, yet low-NAR composite datasets deliver a scalable data foundation for large-scale pre-training and model initialization.

\section{Community-Driven Data Warehouse}
\label{sec:warehouse}

\begin{figure}[!t]
	\begin{center}             
        \includegraphics[width=1\linewidth]{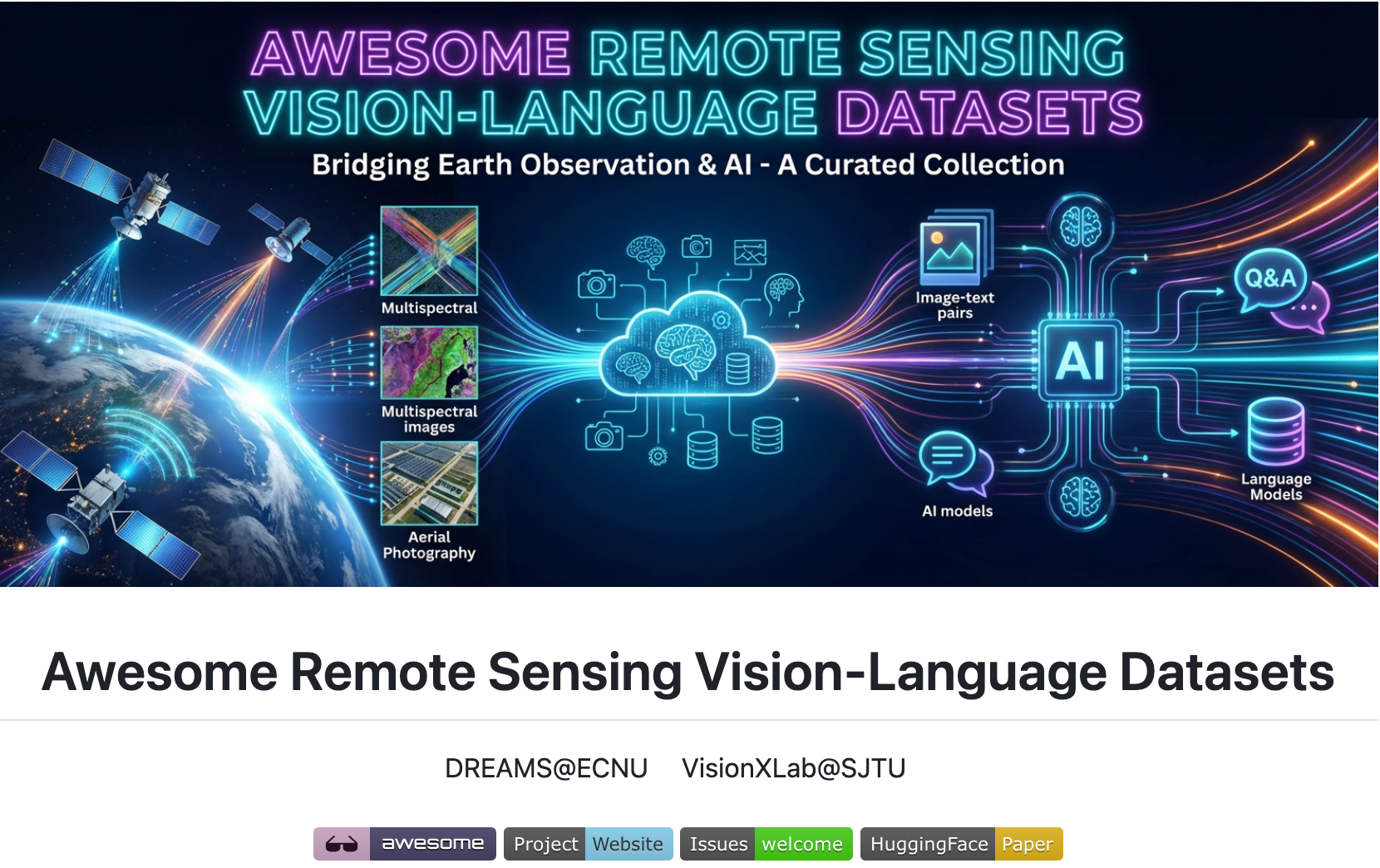}
    \end{center}
    \vspace{-0.3cm}
    \caption{Our Open-source Project  maintained on Github.}
    \label{fig:homepage}        
\end{figure}

After familiarizing ourselves with existing RS vision-language datasets, we devised a four‑step framework, illustrated in Figure \ref{fig:main_4step}, to support researchers in iterative development, culminating in the creation of GeoChat. For a chef, the first and foremost step is to select the appropriate ingredients. Similarly, for a researcher, the first step is to select the appropriate data according to their specific needs. As shown in Figure \ref{fig:homepage}, this section
begins by presenting the foundational component of the project: a community‑driven data warehouse.

To address the challenges of fragmented data resources and the discontinuation of maintenance for existing open-source repositories in the field of RS vision-language research, this paper introduces a community-driven open-source data warehouse, \texttt{Awesome-RS-VL-Data}\footnote{\href{https://github.com/zytx121/Awesome-RS-VL-Data}{Awesome-RS-VL-Data}}, to provide sustained and comprehensive data support for related studies. As an emerging paradigm for data management, the core of a community-driven data warehouse lies in its approach to aggregating, maintaining, and updating data resources through open collaboration, thereby overcoming the limitations in sustainability and scalability inherent in traditional closed databases.

The open source of this data warehouse followed a prolonged period of preparation and internal refinement, including an initial trial operation within internal scenarios for over six months. The primary objective was to accumulate a substantial volume of high-quality dataset resources. Through continuous expansion, the repository now hosts more than 120 datasets, establishing a solid foundation for open sharing with the community. Furthermore, recognizing that other open-source repositories\footnote{\href{https://github.com/Jack-bo1220/Awesome-Remote-Sensing-Foundation-Models}{Awesome-Remote-Sensing-Foundation-Models}}\footnote{\href{https://github.com/lzw-lzw/awesome-remote-sensing-vision-language-models}{awesome-remote-sensing-vision-language-models}}\footnote{\href{https://github.com/geoaigroup/awesome-vision-language-models-for-earth-observation}{awesome-vision-language-models-for-earth-observation}}\footnote{\href{https://github.com/ZhanYang-nwpu/Awesome-Remote-Sensing-Multimodal-Large-Language-Model}{Awesome-Remote-Sensing-Multimodal-Large-Language-Model}} on similar topics have suffered from update stagnation -- often due to shifts in research team focus, graduation of core maintainers, or other commitments -- leading to data resources failing to keep pace with research advancements, we decided to formally open-source our internally curated and mature dataset inventory. This initiative aims to fill the gap for a sustainably maintained data warehouse in this domain. Moving forward, the core development team will continue to participate in the repository's updates and improvements, while also leveraging the power of the community to achieve long-term, iterative data refinement.

\begin{figure}[!t]
	\begin{center}             
        \includegraphics[width=1\linewidth]{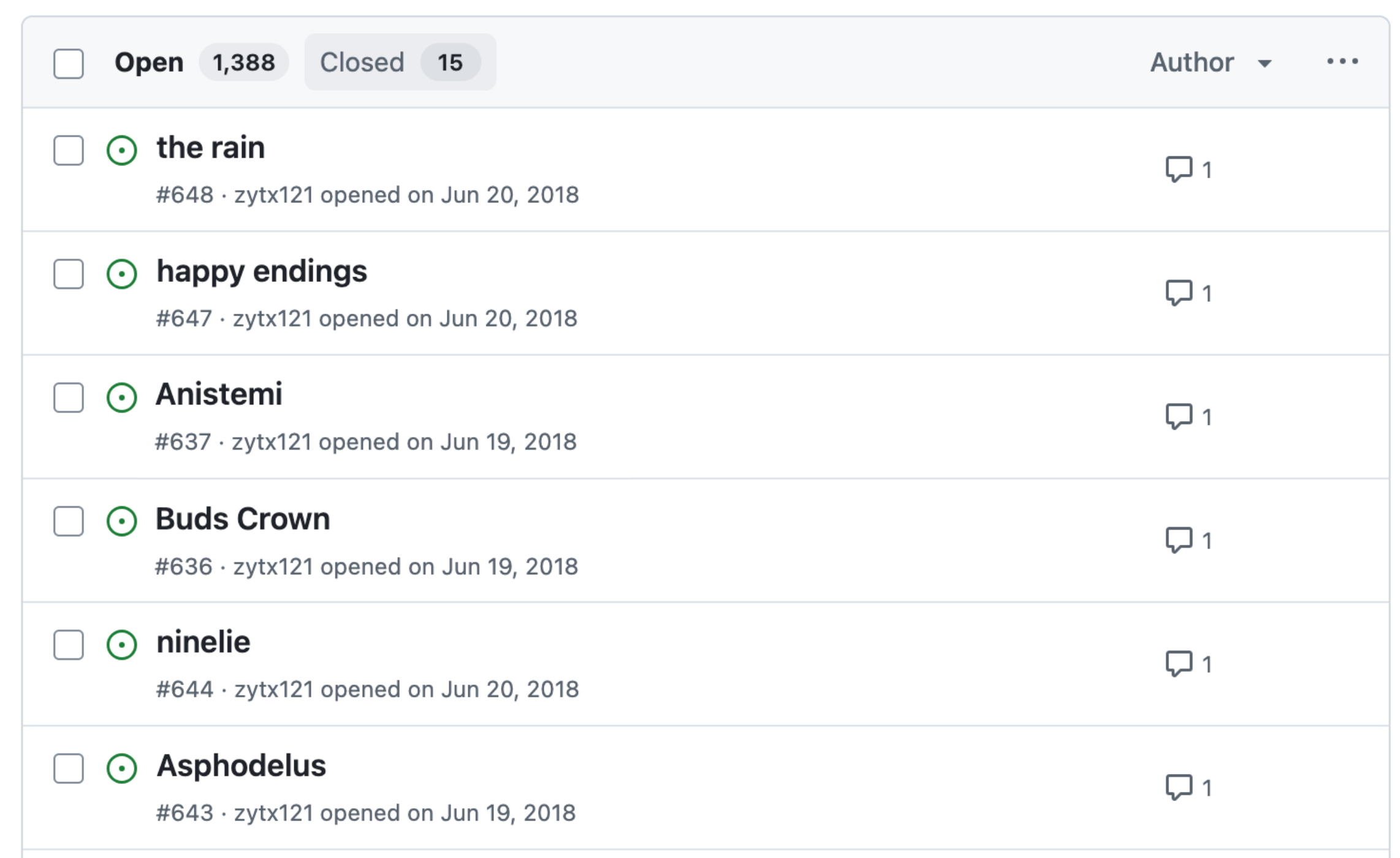}
    \end{center}
    \caption{A community-driven harmonica score library. Currently, 1,388 harmonica scores have been collected, 99\% of which were contributed by community users.}
    \label{fig:je}        
\end{figure}

Compared to traditional private data warehouses, the community-driven model represents the core advantage of this repository. Private data warehouses typically rely on maintenance by specific research teams or individual personnel, making their continuity and updates highly vulnerable to factors such as changes in research direction or turnover of key personnel, often resulting in dataset obsolescence and project abandonment. In contrast, our data warehouse adopts a community-self-maintenance model, opening both the permissions and responsibilities for data upkeep to the entire community. This fundamentally mitigates the sustainability risks associated with a single maintaining entity. The feasibility of this model has been successfully validated through another open-source project we initiated: a harmonica score library\footnote{\href{https://github.com/zytx121/je}{JE Harmonica Score Library}}. By establishing a unified data format and initially manually extracting and standardizing tablature information before transitioning to community self-maintenance after open-sourcing, this project has stably operated for eight years. The community has spontaneously contributed and maintained nearly 1,400 harmonica tablatures, fully demonstrating the significant advantages of the community-driven model in resource accumulation and project sustainability. Even if the core development team suspends direct maintenance due to unforeseen circumstances, the community can still propel the continuous development of the data warehouse.


\begin{figure}[!t]
	\begin{center}             
        \includegraphics[width=1\linewidth]{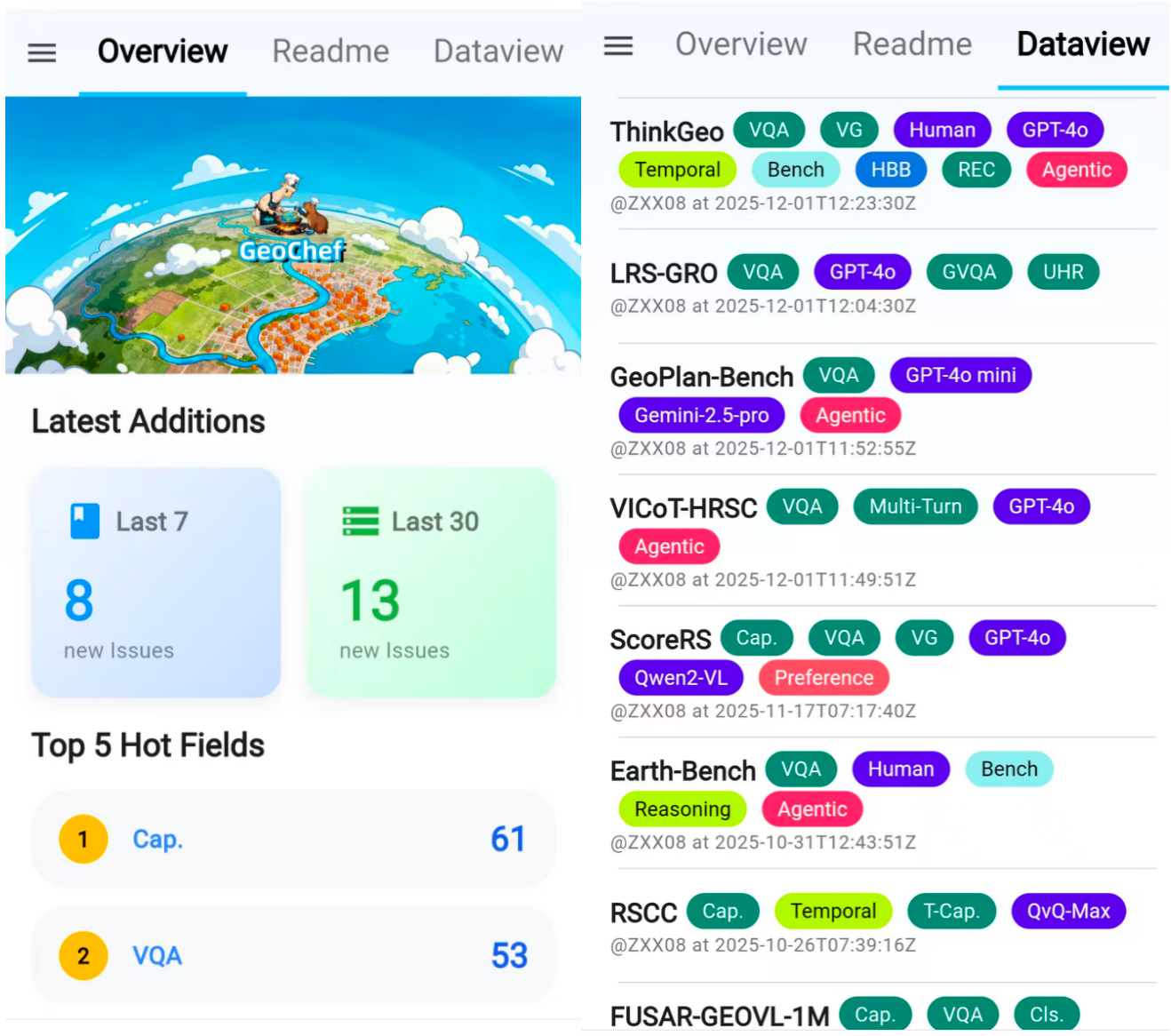}
    \end{center}
    \caption{A mobile app developed based on our data warehouse. It can help researchers real-time grasp the dynamics of RS vision-language datasets.}
    \label{fig:app}        
\end{figure}

In terms of functional design and user experience, this data warehouse incorporates several distinctive features. First, it employs a multi-dimensional label classification system to finely annotate the included datasets. It enables researchers to quickly locate resources that meet their needs. Second, it adheres to a mechanism of continuous updates, closely tracking research trends in the RS vision-language field and consistently incorporating new datasets and related information. This ensures the timeliness and comprehensiveness of the data resources, providing up-to-date support for research and applications. Third, leveraging the GitHub Issues ecosystem and the official GitHub REST API v3, the warehouse possesses powerful extensibility, supporting third-party platform development. The adaptive website GeoCef, built upon this data warehouse, achieves cross-device compatibility and can be seamlessly used on computers, mobile phones, tablets, and other terminals, as shown in Figure \ref{fig:app}. This allows researchers to utilize fragmented time to track the latest dataset dynamics and search for target resources, thereby providing inspiration and high-quality corpus support for research topic selection and model development.

In addition to the GitHub-based open-source data repository functionality, the README interface of this project aggregates resources for RS visual language datasets, including access links to papers and data, as well as corresponding references and data links to the original RS datasets. Furthermore, we have compiled a list of recently published large-scale RS vision-language models and incorporated other high-quality Awesome resource lists in this field, aiming to provide researchers with one-stop information support. This community-driven data warehouse not only establishes an efficient data-sharing platform for RS vision-language research but also fosters a virtuous cycle of community contribution, resource accumulation, research empowerment. In the future, we anticipate broader participation from researchers in maintaining and refining the data warehouse, collectively enriching the data ingredients to provide solid support for the innovative development of next-generation RS intelligent models.

\section{Construction of RS Vision-Language Datasets}
\label{sec:construction}

Dataset construction is the second step to becoming a GeoChef. After selecting the suitable raw data materials, preliminary processing must be performed to prepare for the subsequent cooking process. The construction methodology of RS vision-language datasets is pivotal to their quality and applicability. Based on the level of human involvement and algorithmic automation, mainstream data construction pipelines can be broadly categorized into three paradigms: the completely manual pipeline \cite{wang2025xlrs,yuan2022rsitmd,cheng2022nwpu-cap,liu2022levir-cc}, the completely automatic pipeline \cite{qu2016cap,schuhmann2022laion,liu2024remoteclip,ge2024rsteller,xiong2025geolangbind,zavras2025gaia,si2025spex}, and the semi-automatic pipeline \cite{bashmal2023capera,wang2024skyscript,marimo2025ssl4eos,lu2025gleam,ye2025street2sat,chen2025rscc,ma2025landsat30au}. The choice among these paradigms involves a fundamental trade-off between data quality, production efficiency, diversity, and the requirement for expert knowledge. The following subsections detail the characteristics, advantages, disadvantages, and typical use cases of each approach.

\begin{figure*}[!t]
	\begin{center}             
        \includegraphics[width=1\linewidth]{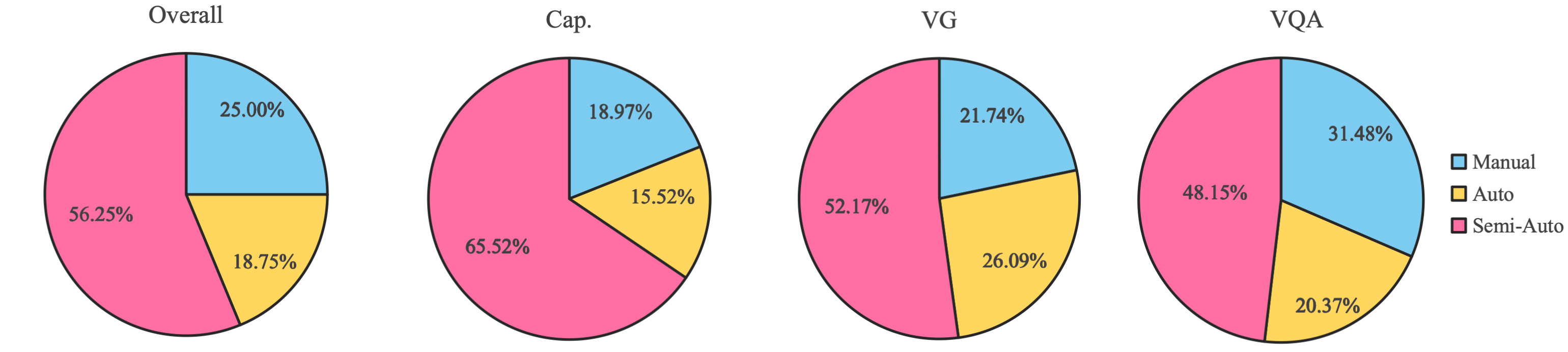}
    \end{center}
    \caption{Distribution of Data Construction Methods Across the Entire and Different Types of Datasets.}
    \label{fig:data_ratio}        
    \vspace{-0.3cm}
\end{figure*}

\subsection{Manually Data Pipeline}

The manually data construction pipeline relies entirely on human experts throughout the entire process. In this paradigm, there is no algorithmic involvement; the generation, cross-validation, and iterative optimization of data samples are all performed by humans. This approach is particularly suited for constructing high-precision, small-sample datasets that demand deep domain expertise, such as benchmark datasets designed for rigorous evaluation. Its foremost advantage lies in the exceptional diversity and high quality of the curated data, as expert knowledge ensures accuracy and richness. However, the primary drawback is its low efficiency and high dependency on skilled human resources, making it prohibitively expensive and impractical for large-scale captioning or generative tasks. A representative successful case of this paradigm is the XLRS-Bench \cite{wang2025xlrs}, a high-quality benchmark for evaluating RS vision-language models.

\subsection{Automatically Data Pipeline}

In contrast, the automatic data construction pipeline operates without any human intervention. It typically starts from existing datasets and employs predefined rules or algorithms to automatically generate templated data samples. This method is highly suitable for building large-scale datasets for repetitive or structured tasks. Its core advantage is high efficiency; once the generation rules and algorithms are verified to be correct, the produced data theoretically requires no manual inspection, enabling rapid scaling. The main disadvantage is the potential lack of data diversity, as the output is constrained by the initial templates and rules. Nevertheless, this limitation can be mitigated by employing LLMs to paraphrase and enrich the generated text. A notable example is the GeoChat \cite{kuckreja2024geochat} dataset, constructed automatically for geo-localized dialogue.

\subsection{Semi-automatically Data Pipeline}

The semi-automatic pipeline represents a synergistic approach that balances human expertise and algorithmic efficiency, making it the most commonly adopted method in industrial practice. In this paradigm, a model (e.g., a large vision-language model) is primarily responsible for the initial data generation, while human annotators are tasked with verification, correction, and optimization. This collaborative workflow effectively balances speed with quality control. The advantages are significant: it yields diverse data quickly by leveraging the generative capacity of models, while human oversight ensures a reliable quality floor. The key disadvantage is the continued requirement for manual verification, which can be applied either to the full dataset or through sampling, thus incurring non-negligible but manageable labor costs. This approach is exemplified by the construction of the VRSBench \cite{li2024vrsbench} dataset, where model generation was coupled with human review.

\begin{figure*}[!t]
	\begin{center}             
        \includegraphics[width=1\linewidth]{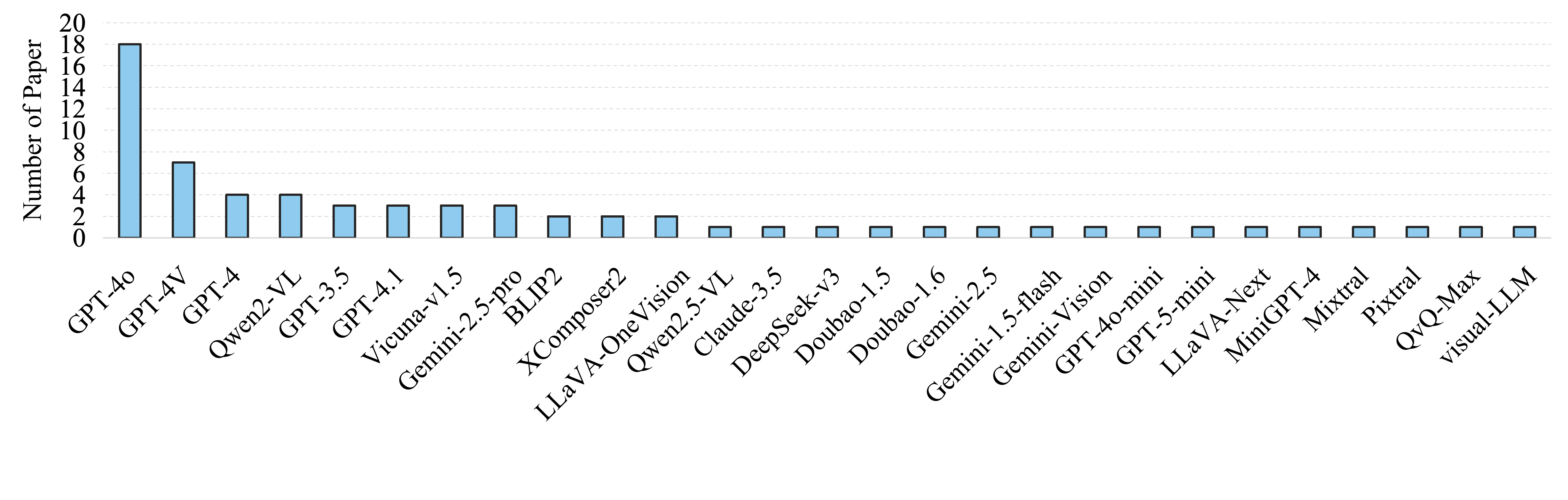}
    \end{center}
    \vspace{-1.cm}
    \caption{Distribution of Data Generation Models in Papers Using Semi-Automatic Data Construction.}
    \label{fig:model_ratio}        
\end{figure*}

\begin{figure*}[!t]
	\begin{center}             
        \includegraphics[width=1\linewidth]{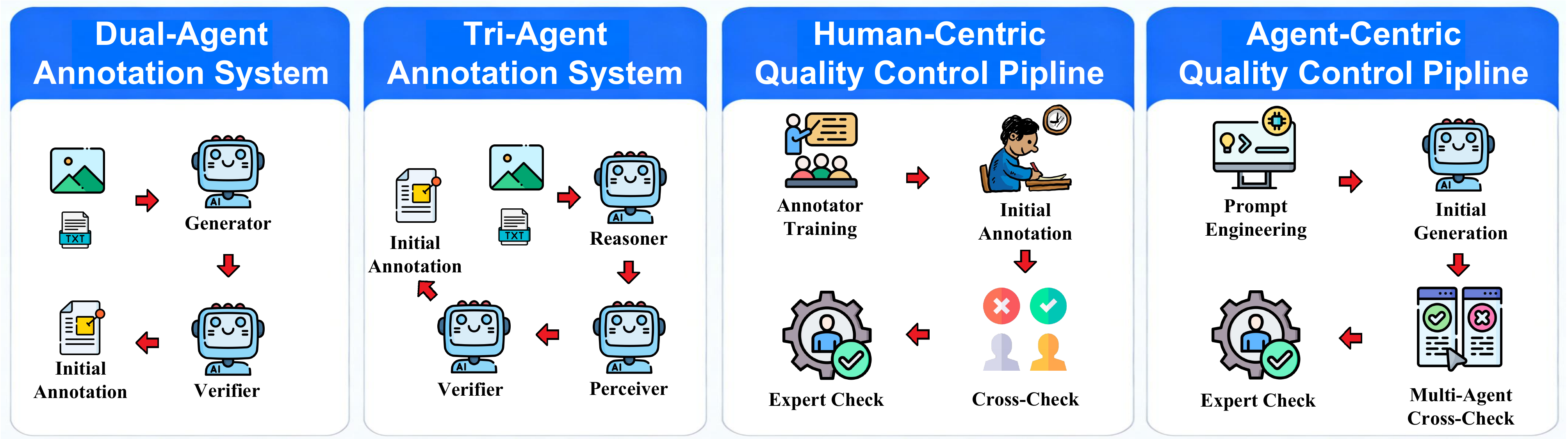}
    \end{center}
    \caption{Multi-Agent Annotation Systems and Quality Control Paradigms.}
    \label{fig:agent_quality_control}        
\end{figure*}

\subsection{Analysis about Data Construction Methods}

Figure \ref{fig:data_ratio} quantifies the distribution of multimodal data construction methods. Semi-automation dominates overall at 56.25\%, peaking at 65.52\% in captioning datasets. Manual annotation varies drastically, reaching its highest share (31.48\%) in VQA datasets, driven by the task’s inherent reasoning complexity and demand for precise visual-text alignment. Fully automatic methods remain secondary and stable (15.52\%–26.09\%), with maximum adoption in the VG dataset, indicating structured-output tasks are more amenable to automation. These patterns reveal a clear correlation between task nature and construction paradigm: description-heavy tasks prioritize semi-automation for scalability, while reasoning-intensive tasks require greater manual intervention. The persistent minority status of automatic methods underscores the fundamental challenge of high-quality unsupervised multimodal data generation, highlighting the need for advanced human-in-the-loop frameworks or generative models to improve efficiency for complex tasks.

Figure \ref{fig:model_ratio} further analyzes model adoption in recent semi-automatic data construction studies. GPT-4o \cite{achiam2023gpt} emerges as the undisputed dominant tool, used in nearly 20 surveyed papers, owing to its robust multimodal capabilities, high instruction fidelity, and reliable API performance—critical for generating coherent, task-aligned synthetic data. Beyond this dominant model, adoption becomes highly fragmented. A secondary tier includes GPT-4V, GPT-4 \cite{achiam2023gpt}, Qwen2-VL \cite{wang2024qwen2}, and GPT-3.5 \cite{lin2023comparison}. The broader ecosystem remains in an active exploratory phase, with researchers evaluating tradeoffs between the cost-effectiveness of smaller models, customizability of open-source options, and specialized capabilities for specific data types. This has led to adoption of alternatives including Claude-3.5 \cite{benzon2025llm}, Gemini variants \cite{team2023gemini}, and leading open-source vision-language models \cite{liu2023llava, chen2023minigpt, bai2023qwen, chen2023internvl}.

\subsection{AI Agent-Driven Data Construction Paradigm and Quality Control}

The AI agent-driven paradigm represents the next evolutionary stage of semi-automatic data construction, replacing the traditional ``single-model generation + human verification'' workflow with a collaborative multi-agent system \cite{luo2025large, muhtar2025scorers} that automates complex, multi-step reasoning and curation processes. Unlike single-model approaches plagued by inherent biases and limited reasoning capabilities, multi-agent systems leverage specialized division of labor to achieve higher annotation quality and scalability. As shown in the Figure \ref{fig:agent_quality_control}, the most prevalent and practical implementation is the dual-agent architecture \cite{luo2025large}, consisting of a generator agent producing initial annotations and an independent evaluator agent filtering low-quality or incorrect samples. This separation of generation and evaluation effectively mitigates confirmation bias and overconfidence issues.

Building on dual-agent systems, recent research has proposed tri-agent architectures to address the challenging task of constructing high-quality chain-of-thought (CoT) reasoning data, which is critical for training RS-VL models with advanced spatial and logical reasoning abilities. A representative example is the RS-EoT-4K dataset pipeline \cite{muhtar2025scorers}, which employs a Socratic questioning framework involving three specialized agents: Reasoner, Perceiver, and Verifier. The Reasoner is blinded to image content and prompted that its collaborator has "weak perceptual capabilities", forcing it to decompose complex spatial problems into simple incremental questions based on local visual features. Conversely, the Perceiver with full image access is prompted that its collaborator has "weak reasoning capabilities", compelling it to provide precise, concise answers without extraneous information. Their iterative Q\&A exchange naturally unfolds into step-by-step reasoning trajectories, which are finally validated for logical consistency and factual accuracy by the Verifier agent. This self-play-based Socratic approach produces more interpretable and logically sound reasoning traces than direct single-model CoT generation, setting a new benchmark for reasoning-focused RS-VL dataset construction.

Despite these advancements, MLLMs are not infallible. In the RS domain with highly specialized knowledge and ambiguous visual patterns, model-generated annotations inevitably contain errors. Thus, human-in-the-loop quality control remains indispensable for high-quality RS-VL datasets. The traditional human-centric quality control pipeline \cite{wang2026cityvlm} consists of four sequential stages: annotator training, initial human annotation, cross-validation by multiple annotators, and expert final confirmation, which relies almost entirely on human labor and is extremely time-consuming and expensive. In contrast, the agent-augmented pipeline shifts the majority of the workload to AI agents, evolving into four new stages: prompt engineering, MLLM initial generation, multi-agent cross-validation, and expert final confirmation, with human intervention limited to only the first and last critical stages. The required intensity of human verification varies by data usage. For training sets where minor noise is tolerable, random sampling checks are sufficient; failed samples can be manually corrected or discarded. However, for test sets used to evaluate model performance and draw scientific conclusions, 100\% full manual verification by domain experts is mandatory, as any errors can lead to misleading performance comparisons and invalid research findings.

Nevertheless, the AI agent-driven paradigm still faces several limitations: current multi-agent systems struggle with highly specialized RS knowledge, their performance is highly dependent on prompt engineering quality, and standardized metrics for evaluating agent-generated reasoning data are lacking. Future research should focus on developing domain-specialized agents, exploring dynamic role assignment architectures, and establishing rigorous benchmarks for agent-constructed datasets, which will further accelerate the development of next-generation RS MLLMs.


\begin{figure}[!t]
	\begin{center}             
        \includegraphics[width=1\linewidth]{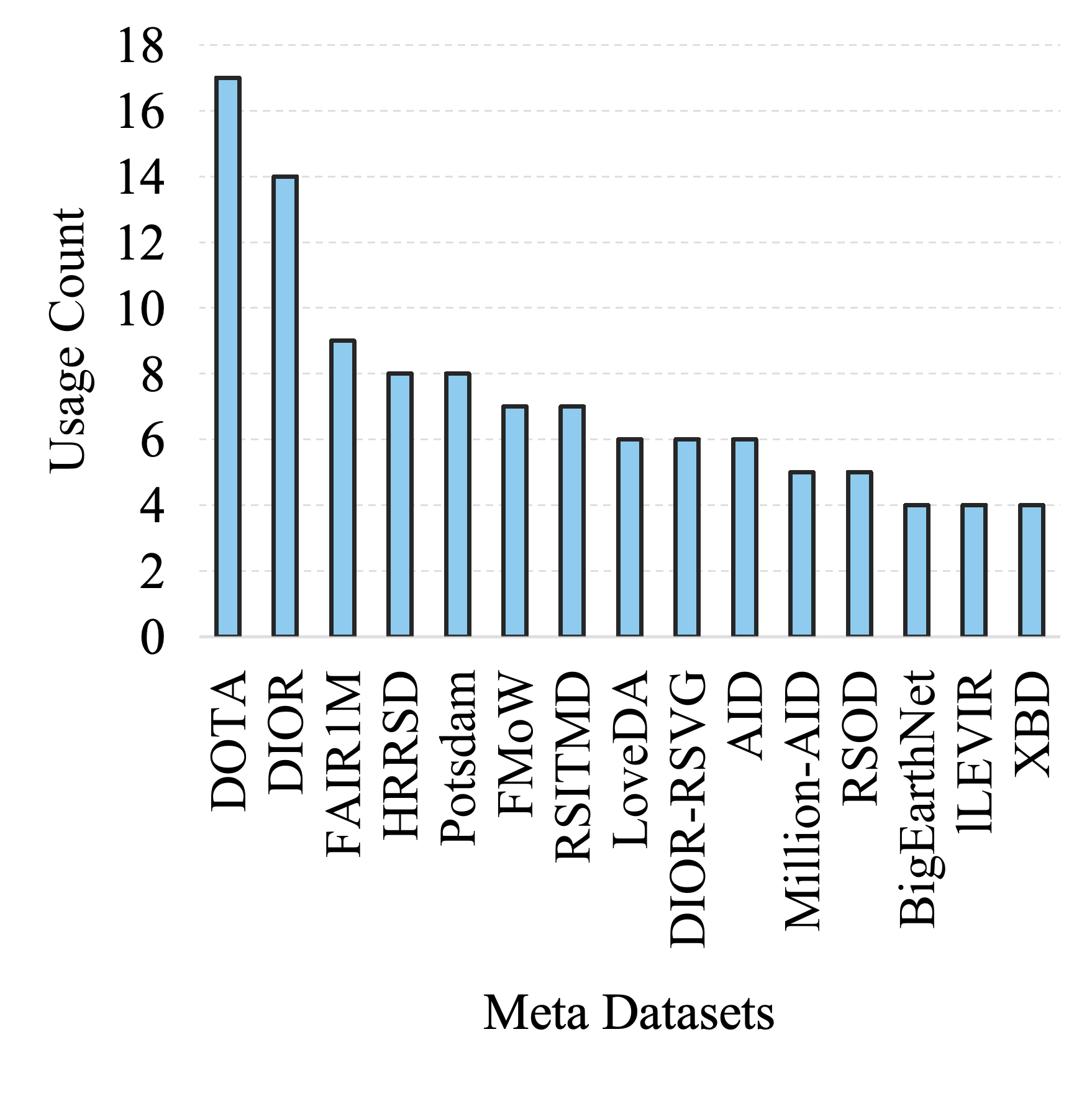}
    \end{center}
    \vspace{-0.8cm}
    \caption{Frequency of Major RS Meta Datasets Used to Construct Vision-Language Datasets.}
    \label{fig:usage_count}        
\end{figure}

\subsection{Best Practices for Constructing High-Quality RS Vision-Language Dataset}

In RS vision-language research, redundant dataset construction and inconsistent quality standards waste valuable resources and hinder model development. As illustrated in Figure \ref{fig:usage_count}, a vast amount of RS metadata is repeatedly repurposed to build RS vision-language datasets. Among them, the classic RS object detection dataset DOTA \cite{xia2018dota} has been repeatedly constructed as a vision-language dataset in 17 studies, resulting in significant redundancy. To address these pervasive challenges, this guide introduces a streamlined three-step workflow to help researchers efficiently construct high-quality datasets.

\begin{table*}[!t]
\centering
\caption{A Quick Lookup Table for Pre-construction Redundancy Check.}
\label{tab:dataset_count}
\setlength{\tabcolsep}{6pt}
\renewcommand{\arraystretch}{1.4}

\begin{tabular}{
l                   
>{\centering\arraybackslash}p{1.6cm}  
>{\centering\arraybackslash}p{2.1cm}  
p{3.2cm}            
p{4cm}              
}
\toprule
\textbf{RS Dataset} & \textbf{Cap} & \textbf{VQA} & \textbf{VG} & \textbf{Comprehensive Data} \\
\midrule
DOTA \cite{xia2018dota}& LRS2M, RSICap & RSIVQA, RemoteCount & GRSAP-1K, GeoSeg-1M, RemoteSAM-270K, RISBench & RSVP, LHRS-Align-Recap \& LHRS-Instruct-Plus, SkyEye-968K, VRSBench-Train, DDFAV, VersaD \& HnstD \& VariousRS-Instruct, MMRS-1M, UrBench, REOBench \\
\addlinespace
DIOR \cite{dior}& LRS2M, Git-10M & RemoteCount & RemoteSAM-270K, RISBench, GRSAP-1K, GeoSeg-1M, OPT-RSVG, DIOR-RSVG & VRSBench-Train, GeoChat-Instruct, DDFAV, MMRS-1M, SARCHat-Bench-2M \\
\addlinespace
FAIR1M \cite{sun2022fair1m}& - & LRS-VQA, MME-RealWorld & GeoSeg-1M, GRSAP-1K & LHRS-Align-Recap \& LHRS-Instruct-Plus, DDFAV, GeoChat-Instruct, VersaD \& HnstD \& VariousRS-Instruct, MMRS-1M  \\
\addlinespace
HRRSD \cite{zhang2019hierarchical}& LRS2M & RSIVQA, RemoteCount & OPT-RSVG, RemoteSAM-270K & RSVP, MMRS-1M  \\
\addlinespace
Potsdam \cite{song2020semantic} & LRS2M & MME-RealWorld & GeoSeg-1M, RemoteSAM-270K & RSVP, VersaD \& HnstD \& VariousRS-Instruct, SARCHat-Bench-2M, UrBench \\
\addlinespace
FMoW \cite{christie2018functional} & RS5M, MMM-RS & GeoLLaVA & EarthReason & VersaD \& HnstD \& VariousRS-Instruct, EarthDial-Instruct, TEOCHatlas  \\
\addlinespace
RSITMD \cite{yuan2022rsitmd} & - & RemoteCount & - & RSVP, SkyEye-968K, LHRS-Align \& LHRS-Instruct, RS-GPT4V, VersaD \& HnstD \& VariousRS-Instruct, MMRS-1M  \\
\addlinespace
LOVEDA \cite{2021_4e732ced}& LRS2M & EarthVQA, RemoteCount & GeoSeg-1M, RemoteSAM-270K & VersaD \& HnstD \& VariousRS-Instruct  \\
\bottomrule
\end{tabular}

\end{table*}

\begin{itemize}
    \item \textbf{Step 1: Pre-construction Redundancy Check:} Before construction, consult Table \ref{tab:dataset_count} to map base RS datasets to their existing vision-language extensions. If similar annotations exist, extend the current dataset or focus on underrepresented domains instead of rebuilding from scratch to reduce redundant workload.
    \item \textbf{Step 2: Task-Specific Semi-Automatic Annotation:} Full manual annotation is costly and unscalable, so adopt a task-specific semi-automatic pipeline: use RS-fine-tuned VLMs for image-level captioning with human verification; combine pre-trained RS object detectors with region-specific caption generators for region-level descriptions; use template-based methods for factual QA and general MLLMs for reasoning QA, with mandatory human validation for all annotations.
    \item \textbf{Step 3: Rigorous Quality Control :} Please follow the two pipelines shown in Figure \ref{fig:agent_quality_control} for quality management. All datasets should include a transparent public quality report recording construction details, standards and limitations.
\end{itemize}

This standardized workflow minimizes redundancy, ensures dataset reliability, and accelerates RS vision-language model development. For users who find manual table consultation cumbersome, our GeoChef agent provides query services: after receiving your needs, it generates personalized dataset construction recommendations.

\section{Fine-tuning of RS Vision-Language Datasets}
\label{sec:fine-tuning}

The third step towards becoming a GeoChef involves fine-tuning the model using the prepared vision-language data. This step can be likened to a chef cooking a dish, as it critically determines the final performance of the model. Although suitable ingredients have been carefully selected and preprocessed, improper control of the culinary process could undermine previous efforts. Thus, this phase is of great importance.

Given that this paper primarily targets universities and small-to-medium-sized research institutions with limited computational resources, which often struggle to bear the high costs of training large models from scratch, it does not cover large-scale pre-training \cite{zoph2020rethinking,chen2023vlp}. Instead, it focuses on deploying effective fine-tuning strategies based on pre-trained general-purpose MLLMs \cite{ma2025exploring,lin2024vila}. As an efficient transfer learning approach, fine-tuning can rapidly adapt large models to specific domains or tasks under constrained resources \cite{yang2024cross,li2023lift}. This chapter will first clarify the core differences and applicable scenarios between two technical paths: full-parameter fine-tuning (FPFT) \cite{lv2024full,liu2024hift} and parameter-efficient fine-tuning (PEFT) \cite{ding2023parameter,wang2025parameter}. Subsequently, it systematically introduces the basic methods of supervised fine-tuning (SFT) \cite{dong2024abilities,wang2023self} and discusses the principles and advantages of reinforcement fine-tuning (RFT) \cite{fan2023dpok,zhai2024fine} techniques that have emerged in this year. By systematically reviewing different fine-tuning strategies, this chapter aims to guide researchers in flexibly selecting appropriate methods based on data characteristics, fully leveraging the potential of data to construct higher-performance MLLMs. 

\subsection{FPFT v.s. PEFT}

FPFT represents the most classical approach to model adaptation. Its core principle involves unlocking all parameters of a pre-trained model and performing end-to-end training updates on a task-specific dataset \cite{ziegler2025craft}. This method aims to meticulously adjust the weights of each layer according to the requirements of the new task, thereby maximizing the model's potential to achieve optimal task-specific performance. In the domain of RS, VHM \cite{pang2024vhm} performed FPFT on 16 NVIDIA A100-80G GPUs, while EarthDial \cite{soni2024earthdial} employed 8 identical GPUs for its training. However, this comprehensive adjustment comes at a significant cost in computational resources and time, making it particularly unsuitable for environments with limited resources, such as universities and small-to-medium-sized research institutions. Furthermore, each model fine-tuned in this manner requires storing a complete parameter copy, leading to substantial storage overhead. It also tends to be prone to overfitting when the dataset is small.

To address the challenges above, PEFT techniques have emerged. The core idea is to keep the main parameters of the pre-trained model fixed, introducing and training only a minimal adapter to learn task-specific knowledge. Among various PEFT techniques, LoRA (Low-Rank Adaptation) \cite{hu2022lora} stands out as one of the most mainstream and practical methods. 
For example, GeoChat \cite{kuckreja2024geochat} freezes the visual encoder and projection layer, fine-tuning only the LLM with LoRA to inject RS knowledge into LLaVA. SkyEyeGPT \cite{zhan2025skyeyegpt} freezes the visual encoder and fine-tunes both the LLM and projection layer with LoRA. RS-LLaVA \cite{yakoub2024rsllava}, by contrast, fine-tunes all three components simultaneously with LoRA. Some methods combine both approaches, such as LHRS-Bot \cite{muhtar2024lhrs} which uses FPFT for connection layers and LoRA for the LLM. 



FPFT and PEFT offer complementary strengths. FPFT is preferred for peak-performance tasks, GPU-rich environments, large domain shifts, and data-abundant scenarios. PEFT, by contrast, is optimal for resource-limited settings, rapid deployment, multi-task learning, and data-scarce applications where overfitting is a major concern \cite{pan2025medvlm,rodriguez2025integrating,wu2025fsvlm}.


\subsection{SFT v.s. RFT}

SFT is fundamentally an imitative learning paradigm\cite{boisseau2024imitation,alvarado2025turing}. Its core mechanism involves guiding models to learn task-specific answer distributions by minimizing the loss between predictions and ground-truth labels using large-scale high-quality input-output annotated data, analogous to students learning from examples. In RS applications, the FPFT and PEFT methods, which were introduced earlier, all fall under the category of SFT. It excels in simple implementation, stable training, low computational cost, rapid achievement of excellent stable performance on well-defined tasks with sufficient high-quality data, and strong output style controllability. Its main limitations include heavy reliance on costly annotated data, confinement to training data distributions, lack of active innovation, poor generalization to dynamically changing or open-ended tasks, and significant overfitting risks.

RFT is an advanced fine-tuning technique built on SFT \cite{agnihotri2025efficient}. It introduces a reinforcement learning (RL) framework that uses reward signals—rather than fixed input-output annotations—to drive model optimization. Recent RS research shows substantial gains from RFT. The Geo-R1 \cite{zhang2025geor1} paradigm enforces a "reason first, then act" process for few-shot geospatial referring expression understanding, where the model first generates explicit reasoning chains before localizing target objects—consistently and substantially outperforming SFT baselines across three few-shot benchmarks while demonstrating strong cross-dataset generalization. DVGBench employs RFT to train an implicit-to-explicit reasoning mechanism for UAV RS imagery, significantly enhancing visual reasoning grounding capabilities. RFT's core mechanism is trial-and-error learning: a preset reward function quantifies how well the model output meets task objectives, and RL algorithms like policy gradients iteratively update the model to maximize rewards. RFT overcomes the annotation limitations of SFT, improving output quality, consistency, task alignment, generalization, and reliability. However, it is ineffective for samples that are always wrong or always correct; its main benefit is on near-decision-boundary samples that are sometimes correct, sometimes not. Therefore, selecting such samples as training data maximizes RFT’s optimization gains.

In practical application scenarios, SFT and RFT are not opposing optimization options but form a progressive relationship of basic laying and advanced improvement in the large model iteration process. For most large model deployment scenarios, it is recommended to first complete SFT training using high-quality annotated data to quickly drive the model to adapt to the core needs of the task. GeoZero \cite{wang2025geozero} employs a two-stage fine-tuning strategy: first SFT, then RFT. It first applies SFT to simple problems to build a solid foundation for the model, and then uses RFT on challenging problems to further unlock the model’s potential. 


However, if it is necessary to carry out research related to Chain-of-Thought (CoT) \cite{wang2025vicotagent}, building on SFT is an inevitable path. Taking exploration research on RS CoT as an example, the general technical route usually first uses a large amount of high-quality CoT data to perform SFT, enabling the model to master basic question-answering capabilities and the ability to output thinking processes, thus becoming a qualified model with basic task adaptability. On this basis, the RFT mechanism is introduced to further optimize the model’s answer quality and CoT logic through a carefully designed reward function \cite{stechly2024chain}. Nevertheless, the lack of high-quality CoT-annotated data in the RS domain poses a significant challenge, as directly employing data without CoT annotations for SFT tends to diminish the MLLM’s ability to produce coherent reasoning chains \cite{xia2025beyond}. To mitigate this issue, two primary strategies have been explored: one leverages advanced closed-source models to automatically generate CoT-annotated SFT samples, while the other applies RFT directly to general-purpose MLLMs, bypassing domain-specific SFT entirely \cite{lyu2023faithful}. SocraticAgent \cite{shao2025socrates} developed a self-play system to generate high-quality CoT-annotated SFT data. The system comprises two AI MLLMs: the Reasoner and the Perceiver, which produce training data by simulating Socratic dialogues. DroneVG-R1 \cite{2026DVGBench} did not perform SFT and directly used manually annotated implicit-to-explicit CoT for RFT.


\section{Evaluation of RS MLLMs}
\label{sec:evaluation}

Model evaluation serves as the final step in the GeoChef pipeline, marking both the completion of a new iterative cycle and the commencement of a new one. Just as the quality of culinary dishes must be judged by customer feedback, the performance of a trained model must be evaluated against standardized benchmarks. Based on the evaluation results, researchers can refine data selection strategies, data preprocessing procedures, and fine-tuning approaches, thereby continuously improving the model—much like honing one’s culinary skills. We summarize the most common public benchmarks and their corresponding evaluation metrics for 4 RS vision-language tasks in Tables \ref{tab:cap_bench}, \ref{tab:vqa_bench}, \ref{tab:vg_region}, and \ref{tab:pixel_vg}, which can serve as a reference for newcomers to the field. In addition to utilizing existing benchmarks, researchers may also construct specialized benchmarks to verify the superiority of their proposed methods for specific problems under fair testing conditions.

\begin{table*}[!t]
\centering
\setlength{\tabcolsep}{14pt}
\caption{Comparison of mainstream RS MLLMs on 3 RS image captioning benchmarks.}
\label{tab:cap_bench}
\resizebox{\linewidth}{!}{
\begin{tabular}{c|cccccc}
\toprule
Dataset & Model & BLEU-4 & METEOR & ROUGE\_1 & ROUGE\_L & CIDEr \\
\midrule
\multirow{6}{*}{RSICD} & RSGPT \cite{hu2023rsgpt} & 36.83 & 30.10 & - & 53.34 & 102.94 \\

 & GeoChat \cite{kuckreja2024geochat}& - & 13.48 &  11.59 &- & 12.39 \\

 & SkyEyeGPT \cite{zhan2025skyeyegpt}& \textbf{59.99} & 35.35 & - & 62.63 & 83.65 \\

 & RS-CapRet \cite{silva2024rscapret}& 43.40 & 37.00 & - & \textbf{63.30} & \textbf{250.20} \\

 & Earthdial \cite{soni2024earthdial}& - & \textbf{56.18} & 33.77 & 27.61 & - \\

 & GeoZero \cite{wang2025geozero}& 35.35 & 52.31 & \textbf{60.16} & 54.47 & 114.82 \\
\midrule
\multirow{8}{*}{UCM-Captions} & RSGPT \cite{hu2023rsgpt}& 65.74 & 42.21 & - & 78.34 & 333.23 \\

 & GeoChat \cite{kuckreja2024geochat}& - & 14.40 &  & 13.22 & 14.27 \\
 & SkyEyeGPT \cite{zhan2025skyeyegpt}& \textbf{78.41} & 46.24 & - & 79.49 & 236.75 \\

 & RS-CapRet \cite{silva2024rscapret}& 67.00 & 47.20 & - & 81.70 & 354.80 \\

 & RS-LLaVA \cite{yakoub2024rsllava}& 72.84 & 47.98 & - & 85.17 & 349.43 \\

 & Earthdial \cite{soni2024earthdial}& - & 51.42 & 40.00 & 34.15 & - \\

 & RingMo-Agent \cite{hu2025ringmoagent}& 77.63 & 51.79 & - & \textbf{85.51} & 373.68 \\

 & GeoZero \cite{wang2025geozero}& 75.72 & \textbf{83.75} & \textbf{84.60} & 83.75 & \textbf{393.57} \\

\midrule
\multirow{7}{*}{Sydney-Captions}
 & RSGPT \cite{hu2023rsgpt}& 62.23 & 41.37 & - & 74.77 & 273.08 \\

 & GeoChat \cite{kuckreja2024geochat}& - & 10.63 &  12.00 & 11.26 & -\\

 & SkyEyeGPT \cite{zhan2025skyeyegpt}& \textbf{77.40} & 46.62 & - & \textbf{77.74} & 181.06 \\

 & RS-CapRet \cite{silva2024rscapret}& 56.40 & 38.80 & - & 70.70 & 239.20 \\

 & Earthdial \cite{soni2024earthdial}& - & 57.31 & 49.39 & 41.00 & - \\

 & GeoZero \cite{wang2025geozero}& 64.08 & \textbf{75.12} & \textbf{75.72} & 74.12 & \textbf{293.49} \\
\bottomrule
\end{tabular}}

\end{table*}

\subsection{Evaluation on RS Image Captioning Tasks}

To ensure a reasonable and well-grounded evaluation, we select five complementary metrics tailored to the multi-dimensional requirements of RS image captioning (lexical accuracy, semantic fidelity, fluency, and informational richness). BLEU-4 \cite{2002BLEU} measures 4-gram precision to assess domain terminology accuracy and syntactic correctness, while METEOR \cite{2005METEOR} compensates for its limitations by emphasizing semantic equivalence and aligning with human judgments. ROUGE-1 (unigram coverage) and ROUGE-L (longest common subsequence) \cite{2004ROUGE} evaluate the capture of key geographic features and spatial coherence, respectively. CIDEr \cite{2014CIDEr}, designed specifically for image captioning, uses TF-IDF weighting to highlight informational salience by prioritizing rare, discriminative terms.

Table \ref{tab:cap_bench} reveals a complex and nuanced competitive landscape within the field of RS image captioning. The GeoZero \cite{wang2025geozero} model emerges as a particularly notable contender. It demonstrates performance approaching the performance ceiling on the UCM-Captions \cite{qu2016cap} dataset, exemplified by its remarkably high METEOR score of 83.75. This indicates an exceptional capability in generating fluent descriptions with near-perfect semantic alignment, suggesting a powerful fitting ability in specific, well-structured scenarios. However, a clear specialization bias is observed across the models. The RS-CapRet \cite{silva2024rscapret} model achieves a standout CIDEr score of 250.20 on the most challenging RSICD \cite{lu2018rsicd} dataset, significantly surpassing all other models. This result highlights its unique advantage in producing highly informative and salient captions. Yet, this superiority is not entirely sustained on the Sydney dataset \cite{qu2016cap}, implying potential boundaries to its generalizability.

In summary, no single model establishes absolute dominance across all datasets and evaluation metrics, underscoring the current absence of a universally superior solution. Model performance is highly dependent on the alignment between its training objectives and the distribution of the test data. The future competitive focus will likely center on developing a general-purpose model that exhibits balanced and robust capabilities across diverse and challenging real-world scenarios.

\subsection{Evaluation on RS VQA Tasks}

In RS VQA tasks, accuracy \cite{2015VQA} is generally regarded as the primary evaluation metric. For counting-oriented questions or regression-based tasks, regression metrics such as mean absolute error (MAE) \cite{2015Exploring} are commonly adopted to quantify prediction errors. In this survey, we uniformly use Overall Accuracy (OA) \cite{lobry2020rsvqa} as the core quantitative evaluation metric across all four RS VQA benchmarks, as it has become the de facto standard in the RS vision-language community. Formally, the OA metric is defined as the percentage of test samples where the model’s predicted answer is completely consistent with the human-annotated ground truth.

\begin{table}[!t]
    \centering
    \setlength{\tabcolsep}{2pt}
\caption{Comparison of mainstream RS MLLMs on 4 RS VQA benchmarks. Among them, XLRS-Bench and LRS-VQA belong to the ultra-high-resolution dataset.}
\label{tab:vqa_bench}
\resizebox{\linewidth}{!}{
  \begin{tabular}{l|cccc}
    \toprule
     Model & RSVQA-LR & RSVQA-HR & XLRS-Bench & LRS-VQA \\
    \midrule
    RSGPT \cite{hu2023rsgpt} & 92.29 & 89.78 & - & -\\
    GeoChat \cite{kuckreja2024geochat}& 91.81 & 70.82 & 22.03 & 19.38 \\
    SkyEyeGPT \cite{zhan2025skyeyegpt}& 84.19 & 81.89 & - & -\\
    EarthGPT \cite{zhang2024earthgpt}& - & 71.15 & - & \\
    VHM \cite{pang2024vhm}& 89.12 & 74.35 & - & 26.69 \\
    RSUniVLM \cite{liu2024rsunivlm} & 92.05 & \textbf{90.85} & - & - \\
    RS-LLaVA \cite{yakoub2024rsllava}& \textbf{93.04} & - & - & -\\
    LHRS-Bot-Nova \cite{li2025lhrsbotnova}& - & 88.45 & - & -\\
    Earthdial \cite{soni2024earthdial}& - & 71.00 & - & -\\
    RingMo-Agent \cite{hu2025ringmoagent}& - & 79.58 & - & -\\
    ScoreRS \cite{muhtar2025scorers}& - & 81.55 & - & -\\
    GeoZero \cite{wang2025geozero}& - & 88.92 & \textbf{48.1} & -\\
    ImageRAG \cite{zhang2025imagerag}&- &- &- & 27.92 \\
    ZoomSearch \cite{zhou2025zoom}& -&- &- & \textbf{32.92} \\
    \bottomrule
  \end{tabular}}
\end{table}

As shown in Table \ref{tab:vqa_bench}, leading models have achieved scores approaching or exceeding 90 on the conventional RSVQA-LR and HR datasets \cite{lobry2020rsvqa}, indicating that the VQA task for RS images at standard resolutions is nearing maturity. However, a drastic performance decline is observed on the ultra-high-resolution benchmarks, XLRS-Bench \cite{wang2025xlrs} and LRS-VQA \cite{luo2025large}, where most reported scores fall within the 20–50 range. This significant gap highlights the fundamental challenges posed by ultra-high-resolution imagery, including perceiving fine-grained visual patterns, filtering critical information from massive pixel data, and comprehending complex long-range spatial relationships. These challenges likely push the capabilities of existing mainstream model architectures to their limits.

Furthermore, model development exhibits a clear trend toward specialization. One group of models aims to achieve state-of-the-art performance on specific tasks, serving as domain-specific experts. For instance, RS-LLaVA \cite{yakoub2024rsllava} achieves the highest score on RSVQA-LR, GeoZero \cite{wang2025geozero} leads by a significant margin on XLRS-Bench, and ZoomSearch \cite{zhou2025zoom} performs best on LRS-VQA. 

Notably, GeoZero presents a promising exception. It maintains high competitiveness on the traditional high-resolution task RSVQA-HR while achieving a commanding lead on the ultra-high-resolution XLRS-Bench. This outcome suggests that its technical approach may possess superior generalization and robustness in handling visual information across different scales and complexities, offering valuable insights for developing next-generation, more powerful and general RS vision-language models.

\subsection{Evaluation on RS VG Tasks}

\begin{table}[!t]
  \centering
  \setlength{\tabcolsep}{8pt}
  \caption{Comparison of mainstream RS MLLMs on 4 RS region-level VG benchmarks. Among them, AVVG belongs to the high-resolution dataset.}
  \resizebox{\linewidth}{!}{
  \begin{tabular}{l|cccc}
  \toprule
   Model & RSVG & DIOR-RSVG & VRSBench & AVVG \\
  \midrule
  GeoChat \cite{kuckreja2024geochat} & 2.04 & 24.05 & 11.52 & 0.28\\
    LHRS-Bot \cite{muhtar2024lhrs}& 1.56 & 17.59 & 1.19 & 0.00\\
  SkySenseGPT \cite{luo2024skysensegpt} & - & 12.87 & - & -\\
  LHRS-Bot-Nova \cite{li2025lhrsbotnova}& - & 31.51 & - & -\\
 
  VHM \cite{pang2024vhm}& - & 56.17 & - & - \\
  GeoGround \cite{zhou2024geoground}& 26.65 & 77.73 & 66.04 & 21.58 \\
  RSUniVLM \cite{liu2024rsunivlm} & - & 72.47 & 69.31 & - \\
  ScoreRS \cite{muhtar2025scorers}& - & 64.52 & - & - \\

  GeoZero \cite{wang2025geozero}& 50.04 & 79.43 & 73.32 & - \\
   RSCoVLM \cite{li2025rscovlm}& \textbf{53.79} & \textbf{84.55} & \textbf{79.73} & \textbf{29.40} \\
 \bottomrule
  \end{tabular}}
  \label{tab:vg_region}
\end{table}

\begin{table}[!t]
\centering
\setlength{\tabcolsep}{6pt}
\caption{Comparison of mainstream RS MLLMs on 2 RS pixel-level VG benchmarks. Among them, EarthReason belongs to the dataset of reasoning segmentation.}
\label{tab:pixel_vg}
\resizebox{\linewidth}{!}{
\begin{tabular}{lccccc}
\toprule
\multirow{2}{*}{Model} & \multicolumn{3}{c}{RRSIS-D} & \multicolumn{2}{c}{EarthReason} \\
\cmidrule(lr){2-4} \cmidrule(lr){5-6}
 & P@0.5 & oIoU & mIoU & cIoU & gIoU \\
\midrule
GeoGround \cite{zhou2024geoground} & 67.50 & - & 60.50 & - & - \\
RemoteSAM \cite{yao2025remotesam}& - & 80.04 & 71.75 & - & - \\
GeoPixel \cite{shabbir2025geopixel}& \textbf{83.33} & \textbf{84.90} & 67.30 & 53.90 & 52.53 \\
SegEarth-R1 \cite{li2025segearth}& 76.96 & 78.01 & 66.40 & 68.25 & 70.75 \\
RemoteReasoner \cite{yao2025remotereasoner}& 54.29 & - & 50.97 & 69.13 & 70.96 \\
UniGeoSeg \cite{ni2025unigeoseg}& - & 81.35 & \textbf{72.23} & \textbf{75.67} & 73.12 \\
SegEarth-R2 \cite{xin2025segearthr2}& - & - & 67.90 & 69.50 & \textbf{73.50} \\
\bottomrule
\end{tabular}}
\end{table}

For RS VG tasks, different metrics are adopted based on the number of target objects. Specifically, IoU@0.5 accuracy \cite{2016Generation} is used for single-object referring grounding to assess the localization accuracy of the target object. For multi-object referring grounding, AP@50 \cite{xia2018dota} is employed to comprehensively evaluate the model’s grounding performance. It should be noted that a higher IoU threshold imposes stricter requirements on grounding precision, and IoU@0.5 is widely recognized as the standard threshold in the field of RS VG.

Table \ref{tab:vg_region} reveals a significant performance gap between advanced and early MLLMs in region-level VG. Leading models, such as RSCoVLM \cite{li2025rscovlm} and GeoZero \cite{wang2025geozero}, score above 50 points across datasets, including RSVG \cite{sun2022rsvg}, DIOR-RSVG \cite{zhan2023dior-rsvg}, and VRSBench \cite{li2024vrsbench}, demonstrating strong geolocation capabilities. In contrast, earlier models, such as GeoChat \cite{kuckreja2024geochat} and LHRS-Bot \cite{muhtar2024lhrs}, generally score below 25 points, with performance on the high-resolution AVVG dataset \cite{zhou2024geoground} approaching zero. This gap highlights rapid technological progress in the field. The AVVG dataset serves as a challenging benchmark, where all models perform significantly worse; even the top model, RSCoVLM, achieves only 29.40 points.

Table \ref{tab:pixel_vg} further illustrates a trend of capability differentiation in pixel-level RS MLLMs, marked by a competition between basic segmentation accuracy and advanced reasoning. UniGeoSeg \cite{ni2025unigeoseg} excels in both tasks, particularly on the challenging EarthReason dataset \cite{li2025segearth}, demonstrating strong generalization and reasoning abilities for complex instructions that involve geographical knowledge. It can be considered an all-round performer. Conversely, GeoPixel \cite{shabbir2025geopixel} leads in traditional segmentation on the RRSIS-D \cite{liu2024rotated} dataset, particularly in the precise localization metric P@0.5; however, its performance declines on the EarthReason task. This suggests that models optimized for traditional segmentation may not be designed for the high-level semantic understanding required by reasoning-intensive tasks. Additionally, SegEarth shows minimal change in basic segmentation metrics from version R1 to R2 \cite{xin2025segearthr2}, but a clear improvement in reasoning, indicating a focused effort on enhancing reasoning capabilities, which aligns with the field's move toward a deeper geographical understanding.

In summary, the two tables above clearly outline a watershed in the current development of RS
MLLMs: models are no longer solely pursuing accurate vision but are also diverging into a new track of deep reasoning.

\subsection{Limitations of Current Evaluation}

The rapid proliferation of RS MLLMs has exposed fundamental flaws in existing evaluation paradigms, where inflated performance metrics often bear little relation to a model's true generalization ability. A core and pervasive issue is information leakage, which undermines the validity of benchmark comparisons and misguides research directions. We first systematically categorize three distinct yet interconnected information leakage pathways prevalent in RS-VLM benchmarks:

\begin{itemize}
\item \textbf{Pre-training leakage} has become particularly prominent as large-scale pre-trained VLMs \cite{liu2024remoteclip,wang2025sarclip} have become a standard paradigm. Current general-domain pre-training corpora such as CLIP \cite{hafner2021clip} incorporate billions of RS images and text descriptions \cite{schuhmann2022laion}, leading downstream task evaluation to essentially become a seen-answer test that severely overestimates the model’s zero-shot and few-shot performance. However, the risk of pre-training leakage is often overlooked in the era of large models.

\item \textbf{Metadata leakage} occurs because the visual-language datasets used for training and testing are constructed based on the same RS meta-datasets. As shown in Fig. \ref{fig:usage_count}, as many as 17 RS visual-language datasets have been built on DOTA \cite{xia2018dota}, resulting in severe metadata leakage risks among these datasets. The presence of identical images in the test and training sets can still inflate evaluation accuracy, even if the text annotations are not fully consistent.

\item \textbf{Reasoning grounding leakage} arises when reasoning referring texts are directly annotated on the basis of original referring expressions. With the development of visual reasoning, an increasing number of reasoning grounding datasets \cite{li2025segearth, xin2025segearthr2,jiang2025grasp,2026DVGBench} have emerged. Some of these datasets directly generate reasoning texts from existing VG datasets without changing the referred objects, allowing MLLMs fine-tuned on non-reasoning VG datasets to blindly predict the answers to reasoning grounding tasks.

\end{itemize}

Beyond the three types of information leakage discussed previously, discrepancies in evaluation implementation details can also undermine the impartiality of benchmarking. In what follows, we elaborate on three evaluation criteria that are frequently neglected:

\begin{itemize}
\item \textbf{Prompting Consistency Criteria}: For benchmark-focused papers, evaluate representative MLLMs’ zero-shot performance on the proposed benchmark \cite{wang2025xlrs}. Ensure task prompts are consistent with the original literature, as prompt variations significantly affect task performance -- e.g., Qwen2.5-VL’s support for absolute and 1000 \cite{wang2024qwen2} -- scaled normalized coordinates causes large precision gaps in VG.

\item \textbf{Answer Extraction Criteria}: For VQA tasks with stringent output format constraints, final answers can be extracted via regular expression matching or constrained through multi-turn dialogue to enforce standardized output formats \cite{zhou2025avimath}. Uniform answer extraction strategies must be applied across all evaluated models to ensure consistent and comparable results.

\item \textbf{SFT Data Consistency Criteria}: For fine-tuning studies \cite{wang2025geozero, zhang2025geor1}, ensure all baseline MLLMs are fine-tuned on the same training dataset for fair comparison. Test sets often align with training data, so fine-tuned models gain advantages that may obscure architectural or strategy contributions, a detail sometimes overlooked to highlight method superiority.

\end{itemize}

\subsection{Future Direction of Benchmarks}

At this stage, RS vision-language benchmarks have gradually evolved from single-temporal and single-modal to multi-temporal and multi-modal settings \cite{irvin2024teochat}, from image chips to large-scale RS scenes \cite{wang2025xlrs}, and from shallow semantic matching to deep cognitive reasoning \cite{2026DVGBench}.

However, two promising trends can be foreseen in future benchmark development. First, benchmarks are shifting from general capability evaluation to domain specialization. They are moving beyond generic scene understanding toward professional scenarios such as disaster response \cite{wang2025disasterm3}, agricultural monitoring \cite{li2025agromind}, and photogrammetry \cite{zhou2025avimath}, enabling fine-grained assessment of domain semantics, and expert knowledge alignment.
Second, evaluation paradigms are transitioning from model-centric testing to agent-oriented assessment \cite{2025Land}, driven by the explosive rise and widespread adoption of AI agent technologies. Unlike standalone large models that are mostly constrained to static reasoning within fixed knowledge boundaries, AI agents can execute far more complex, multi-step real-world tasks through flexible tool invocation and modular collaborative workflows \cite{shabbir2025thinkgeo}.

\section{Challenges and Future Directions}
\label{sec:future}

\subsection{Reasoning Tasks}

The research focus of RS vision-language models is shifting from fundamental perception tasks to more challenging, complex reasoning tasks. This transition stems from a core challenge: conventional tasks such as object detection \cite{yang2024thinking} and semantic segmentation \cite{he2017mask} can often be effectively accomplished without leveraging the deep comprehension and reasoning capabilities of ML'LMs, making it difficult to fully exploit the advantages of MLLMs. To break through this bottleneck, researchers have begun constructing a series of new benchmark datasets that require deep intent understanding and logical reasoning, aiming to evaluate and promote model performance at higher cognitive levels. These emerging reasoning tasks primarily present challenges across several dimensions. Implicit VG \cite{li2025segearth} requires models to understand instructions that imply specific attributes or functional relationships rather than directly referring to object appearance. Temporal reasoning \cite{irvin2024teochat} demands that models analyze multi-temporal RS images to understand dynamic processes such as urban expansion and disaster evolution. Relational reasoning \cite{luo2024skysensegpt} challenges models to comprehend intricate interactive relationships among multiple ground objects within an image. 3D spatial reasoning \cite{zhou2025airspatialbot} elevates the challenge from the two-dimensional image plane to the three-dimensional spatial level, requiring an understanding of object height, volume, and 3D spatial relationships. Multimodal mathematical reasoning \cite{zhou2025avimath} extends the challenge to integrating mathematical logic and visual information, testing the model's ability to transform visual information into mathematical models and perform calculations. In summary, the frontier challenges in RS vision-language data for reasoning tasks are centered on advancing models beyond basic seeing and recognizing towards deeper cognitive capabilities. These challenges are guiding RS vision-language research towards more profound and general spatial intelligence.

\subsection{Ultra-High-Resolution RS Imagery}

The processing of very large-scale RS imagery poses a systemic challenge for current RS MLLMs, spanning the dimensions of data, algorithms, and computational resources. From a data foundation perspective, the cost of constructing high-quality benchmarking and training datasets remains prohibitively high. Such imagery requires fine-grained annotation reliant on specialized RS domain knowledge. While purely manual annotation can ensure quality, it suffers from efficiency limitations. For instance, the development of benchmarks like XLRS-Bench \cite{wang2025xlrs} necessitates annotating dozens of tasks for a single image; this extremely time-consuming nature directly constrains the accumulation of large-scale, high-quality, contextually rich data. Through manual annotation of 4K UAV imagery, the AVVG \cite{zhou2024geoground} created a high-resolution vehicle VG dataset, which provides fine-grained attributes including vehicle color, brand, and model.

At the model input and information retention level, a fundamental contradiction exists. Drastically compressing imagery with tens of thousands of pixels down to the standard size acceptable for models results in a severe loss of rich spatial contextual information and fine object details \cite{zhang2025imagerag}. Conversely, avoiding compression is often infeasible as the original image dimensions far exceed the processing limits of most models, preventing them from effectively capturing long-range dependencies. Since the core of RS interpretation is precise measurement of surface information, this contradiction ultimately results in subpar model performance in critical tasks such as object counting and fine-grained object identification.

Furthermore, analyzing internal processing efficiency highlights the pronounced challenge posed by the low semantic density characteristic of RS imagery. Vast areas of ineffective background, such as oceans and forests, prompt the visual encoder to generate a massive number of redundant visual tokens \cite{luo2025large}. These tokens contain minor meaningful semantics, consume scarce computational resources, and can interfere with the modeling of crucial targets.

In summary, addressing these challenges requires continued and in-depth exploration from researchers in areas such as data and image construction, model algorithm innovation \cite{zhou2025zoom}, and the development of evaluation benchmarks.

\subsection{Multi-source RS Imagery}

Currently, a critical limitation in the field is that the vast majority of existing automatic and semi-automatic vision-language data construction pipelines are still largely restricted to optical imagery, due to the lack of powerful and reliable intelligent agents capable of processing multi-spectral \cite{zavras2025gaia}, synthetic aperture radar (SAR) \cite{wei2025sarlang}, and other specialized remote sensing (RS) modalities. This makes it difficult for existing general-purpose MLLMs to directly and accurately interpret their visual content, thereby severely degrading the reliability of automatically and semi-automatically generated image-text paired data. Particularly for SAR images, their nature as syntheses of radar echoes necessitates deep domain expertise for object recognition, posing substantial difficulties for automated data annotation.

To address this challenge, the community has explored various innovative solutions. One effective strategy is modality transfer. For instance, the SARLANG-1M \cite{wei2025sarlang} dataset adopts this approach by leveraging paired SAR and optical images: a general vision-language model first generates descriptions for the more interpretable optical images, and these descriptions are then transferred to the corresponding SAR images as textual annotations. This method skillfully circumvents the difficulty of directly interpreting specialized imagery, though it usually still requires expert manual review to ensure final quality. Another approach involves constructing more sophisticated automated annotation frameworks to improve description quality. For example, SAR-TEXT \cite{he2025sartext} developed a pipeline that first uses a model to generate preliminary captions from optical images, simultaneously converting segmented images into textual descriptions via algorithms, and then feeds both pieces of information into a large language model. Leveraging its in-context learning capability, the framework produces high-quality, detailed descriptions. Although this method demands higher input requirements (paired SAR images, optical images, and segmentation maps), it can yield very fine-grained annotations. When paired optical images are unavailable, researchers have also explored methods that do not rely on external paired data. The SARCAP \cite{jiang2025sarcap} dataset provides a typical example: it directly utilizes existing object detection or semantic segmentation annotations of SAR images. It generates corresponding textual descriptions based on predefined rules and templates. Similarly, the OSVQA \cite{zhao2025osvqa} dataset first annotates basic attributes of images under expert guidance and then uses templates to generate diverse visual question-answer pairs. This approach fully exploits the potential of existing specialized datasets, enabling the construction of valuable vision-language data even with a single data modality. Furthermore, by automatically associating large-scale unlabeled RS imagery with rich semantic information from open geoinformation databases, large-scale and semantically diverse datasets can also be built, as successfully demonstrated by the SkyScript \cite{wang2024skyscript} dataset.

It is worth noting that the core ideas behind the above methodologies are not limited to addressing data construction for SAR imagery. They possess high generalizability and transfer value and can be effectively applied to other specialized RS data domains. For instance, in modalities such as nighttime light \cite{WU2023103495} and thermal infrared \cite{zhu2023thermal} RS imagery, which also face annotation difficulties, these approaches can provide important references for constructing high-quality vision-language datasets.

\subsection{UAV Imagery}

Embodied intelligence has emerged as a prominent frontier in AI research \cite{hao2025mimoembodied}. UAVs, often referred to as aerial robots, situate themselves at the intersection of embodied intelligence and RS technologies. Existing vision-language datasets for UAVs are expanding beyond foundational capabilities, such as navigation \cite{chu2024towards,gao2025openfly}, to progressively encompass diverse competency dimensions, including 3D spatial perception \cite{zhou2025airspatialbot, zhang2025spatialsky}, mathematical reasoning \cite{zhou2025avimath}, complex decision-making \cite{guo2025bedi}, multi-agent collaboration \cite{zha2025aircop}, and security assurance \cite{wang2025jtd}. As a critical enabler for the development of aerial embodied intelligence, these UAV vision-language datasets are constructing a comprehensive capability framework that spans perception, cognition, decision-making, collaboration, and safety, thereby propelling the field from basic environmental awareness towards deep human-UAV interaction and sophisticated decision-making.

Through the efficient integration of simulated and real-world data \cite{gao2025openfly,wang2025uavflow}, coupled with deep learning from multimodal and multi-agent technologies \cite{zhang2025open3dvqa}, UAVs hold significant potential to evolve from passive aerial cameras that merely execute tasks into proactive, embodied agents capable of autonomous environmental perception, intent understanding, interactive collaboration, and independent decision-making.

\subsection{Agentic AI in RS}

One of the most remarkable advancements in the current AI domain is the rapid development of AI agents. These agents not only represent a key direction of technological evolution, but are also regarded by industries as a critical force for fostering new productive forces. However, a significant bottleneck that hinders the progress of AI agents from RS lies in the acquisition and construction of high-quality task-specific datasets. This process faces three core challenges. First, the field involves extremely high expertise barriers. RS interpretation relies on profound knowledge of Earth sciences. Tasks such as land surface temperature retrieval or nighttime light analysis \cite{WU2023103495} involve complex physical models and specialized interpretation rules. Consequently, data annotation must be performed by domain experts, leading to high costs and prolonged cycles. Second, task logic is inherently complex. Practical RS workflows, like disaster assessment or environmental monitoring, often require multi-step quantitative reasoning and the coordinated invocation of various specialized tools, for example, spectral index calculation \cite{2025Land} or change detection models \cite{deng2025changechat}. This imposes stringent requirements on the design and validation of task sequences within datasets. Third, data and task formats are highly heterogeneous. The diversity of RS data sources and the dispersion of task objectives make standardized, large-scale annotation exceptionally difficult.

To address these challenges, current research primarily depends on manual construction paradigms. For instance, datasets like UnivEarth \cite{kao2025univearth} and GeoPlan-Bench \cite{li2025geoplan} employ a reverse engineering approach, deriving questions from existing scientific reports or predefined tool pathways to ensure answer accuracy and verifiability. The ThinkGeo \cite{shabbir2025thinkgeo} dataset comprises expert-designed and validated reasoning chains, encompassing thousands of steps to ensure logical rigor. Earth-Bench further encapsulates domain knowledge into hundreds of standardized tools, constructing problems around tool combinations. These methods collectively form the mainstream benchmarking frameworks, providing valuable high-quality standards for evaluating the fundamental capabilities of AI agents. However, their fundamental limitation is the deep reliance on expert-led design, annotation, and validation. This results in constrained dataset scales, typically limited to a few hundred or thousands of samples, poor scalability, difficulty in covering real-world long-tail scenarios, and a nature more suited for testing rather than training sets.

Therefore, breaking through the data scale bottleneck is crucial for advancing this field. Establishing sustainable, low-cost, and high-quality data augmentation mechanisms represents an urgent and promising research direction to unlock the practical application potential of RS AI agents.

\subsection{RS Application Scenarios}

Building high-quality, domain-specific datasets is pivotal for advancing RS technology from mere perception to deeper cognition and informed decision-making. Their core value lies in bridging the gap between general-purpose MLLMs and the specialized demands of the field. By infusing models with expert domain knowledge, these datasets enable an understanding of the unique perspectives, professional geo-objects, and technical terminology characteristic of RS imagery. This capability is fundamental for supporting complex analytical and decision-making tasks, such as disaster assessment \cite{maryam2021floodnet} and urban planning \cite{wang2024earthvqa}. Furthermore, such datasets form the cornerstone of natural human-machine interaction, allowing experts to efficiently query data using intuitive instructions like analyzing road blockages after an earthquake, thereby significantly boosting operational efficiency \cite{wang2025disasterm3}.

However, the construction process is fraught with challenges. The primary hurdle is achieving annotation expertise and consistency. Interpreting RS images requires profound domain knowledge, making precise annotation inherently costly. To ensure quality, a human-machine collaborative strategy is often essential, typically involving AI-powered initial screening followed by verification by domain experts, supplemented by rigorous annotator training, cross-validation, and expert sampling audits \cite{wang2026cityvlm}. Another critical challenge is balancing task design, scientificity, and practicality. Datasets must not only pursue high-precision metrics but also effectively address real-world industry pain points. This demands a deep understanding of business logic from builders, who must design diverse problems, correlate multi-source economic data, and even simulate real disaster scenarios to ensure the data underpins effective decision-making. Ultimately, striking a balance between data scale, quality, and cost is crucial. A feasible path currently involves leveraging AI-assisted generation with expert oversight to strike a balance between efficiency and quality.

\section{Conclusion}
\label{sec:conclusion}

This paper concludes by presenting a comprehensive, data-centric guide for tailoring Multimodal Large Language Models (MLLMs) in RS, centered on the novel GeoChef framework that systematizes the entire pipeline from dataset curation to model evaluation. Through a systematic survey of over 120 vision-language datasets, a detailed taxonomy, and a community-driven data warehouse, we provide practical methodologies for constructing, fine-tuning, and assessing MLLMs, empowering researchers to overcome resource constraints and achieve specialized model performance. We anticipate that this work can provide practical guidance for academia and industry, promoting the development of more diverse and higher-quality vision-language datasets for specialized RS fields. Ultimately, our long-term vision is to enable the organic integration of these scattered data resources, gradually forming a comprehensive, geographically widespread, and annotation-standardized super RS vision-language corpus. This will not only serve as the foundation for efficient and reproducible research and development but also lay a solid and indispensable data cornerstone for advancing toward a GeoAI with robust generalization and cognitive capabilities.



%




\ifCLASSOPTIONcaptionsoff
  \newpage
\fi



\bibliographystyle{IEEEtran}
\bibliography{IEEEabrv,Mendeley_Tarot}

\begin{thebibliography}{100}
\providecommand{\url}[1]{#1}
\csname url@samestyle\endcsname
\providecommand{\newblock}{\relax}
\providecommand{\bibinfo}[2]{#2}
\providecommand{\BIBentrySTDinterwordspacing}{\spaceskip=0pt\relax}
\providecommand{\BIBentryALTinterwordstretchfactor}{4}
\providecommand{\BIBentryALTinterwordspacing}{\spaceskip=\fontdimen2\font plus
\BIBentryALTinterwordstretchfactor\fontdimen3\font minus \fontdimen4\font\relax}
\providecommand{\BIBforeignlanguage}[2]{{%
\expandafter\ifx\csname l@#1\endcsname\relax
\typeout{** WARNING: IEEEtran.bst: No hyphenation pattern has been}%
\typeout{** loaded for the language `#1'. Using the pattern for}%
\typeout{** the default language instead.}%
\else
\language=\csname l@#1\endcsname
\fi
#2}}
\providecommand{\BIBdecl}{\relax}
\BIBdecl

\bibitem{radford2021learning}
A.~Radford, J.~W. Kim, C.~Hallacy, A.~Ramesh, G.~Goh, S.~Agarwal, G.~Sastry, A.~Askell, P.~Mishkin, J.~Clark \emph{et~al.}, ``Learning transferable visual models from natural language supervision,'' in \emph{International conference on machine learning}.\hskip 1em plus 0.5em minus 0.4em\relax PmLR, 2021, pp. 8748--8763.

\bibitem{hafner2021clip}
M.~Hafner, M.~Katsantoni, T.~K{\"o}ster, J.~Marks, J.~Mukherjee, D.~Staiger, J.~Ule, and M.~Zavolan, ``Clip and complementary methods,'' \emph{Nature Reviews Methods Primers}, vol.~1, no.~1, p.~20, 2021.

\bibitem{sun2024alpha}
Z.~Sun, Y.~Fang, T.~Wu, P.~Zhang, Y.~Zang, S.~Kong, Y.~Xiong, D.~Lin, and J.~Wang, ``Alpha-clip: A clip model focusing on wherever you want,'' in \emph{Proceedings of the IEEE/CVF conference on computer vision and pattern recognition}, 2024, pp. 13\,019--13\,029.

\bibitem{moor2023foundation}
M.~Moor, O.~Banerjee, Z.~S.~H. Abad, H.~M. Krumholz, J.~Leskovec, E.~J. Topol, and P.~Rajpurkar, ``Foundation models for generalist medical artificial intelligence,'' \emph{Nature}, vol. 616, no. 7956, pp. 259--265, 2023.

\bibitem{qin2024tool}
Y.~Qin, S.~Hu, Y.~Lin, W.~Chen, N.~Ding, G.~Cui, Z.~Zeng, X.~Zhou, Y.~Huang, C.~Xiao \emph{et~al.}, ``Tool learning with foundation models,'' \emph{ACM Computing Surveys}, vol.~57, no.~4, pp. 1--40, 2024.

\bibitem{awais2025foundation}
M.~Awais, M.~Naseer, S.~Khan, R.~M. Anwer, H.~Cholakkal, M.~Shah, M.-H. Yang, and F.~S. Khan, ``Foundation models defining a new era in vision: a survey and outlook,'' \emph{IEEE Transactions on Pattern Analysis and Machine Intelligence}, 2025.

\bibitem{li2024vision}
X.~Li, C.~Wen, Y.~Hu, Z.~Yuan, and X.~X. Zhu, ``Vision-language models in remote sensing: Current progress and future trends,'' \emph{IEEE Geoscience and Remote Sensing Magazine}, vol.~12, no.~2, pp. 32--66, 2024.

\bibitem{zhang2024vision}
J.~Zhang, J.~Huang, S.~Jin, and S.~Lu, ``Vision-language models for vision tasks: A survey,'' \emph{IEEE transactions on pattern analysis and machine intelligence}, vol.~46, no.~8, pp. 5625--5644, 2024.

\bibitem{zhang2025large}
Z.~Zhang, J.~Wang, F.~Wen, Y.~Guo, X.~Zhao, X.~Fang, S.~Ding, Z.~Jia, J.~Xiao, Y.~Shen \emph{et~al.}, ``Large multimodal models evaluation: a survey,'' \emph{Science China Information Sciences}, vol.~68, no.~12, p. 221301, 2025.

\bibitem{zeng2023x}
Y.~Zeng, X.~Zhang, H.~Li, J.~Wang, J.~Zhang, and W.~Zhou, ``X2-vlm: All-in-one pre-trained model for vision-language tasks,'' \emph{IEEE transactions on pattern analysis and machine intelligence}, vol.~46, no.~5, pp. 3156--3168, 2023.

\bibitem{zhao2025med}
Y.~Zhao, E.~Zhong, C.~Yuan, Y.~Li, M.~Zhao, C.~Li, J.~Hu, W.~Liu, and C.~Liu, ``Med-vlm: Enhancing medical image segmentation accuracy through vision-language model,'' in \emph{Proceedings of the IEEE/CVF International Conference on Computer Vision}, 2025, pp. 7283--7293.

\bibitem{hao2025mimoembodied}
X.~Hao, L.~Zhou, Z.~Huang, Z.~Hou, Y.~Tang, L.~Zhang, G.~Li, Z.~Lu, S.~Ren, X.~Meng, Y.~Zhang, J.~Wu, J.~Lu, C.~Dang, J.~Guan, J.~Wu, Z.~Hou, H.~Li, S.~Xia, M.~Zhou, Y.~Zheng, Z.~Yue, S.~Gu, H.~Tian, Y.~Shen, J.~Cui, W.~Zhang, S.~Xu, B.~Wang, H.~Sun, Z.~Zhu, Y.~Jiang, Z.~Guo, C.~Gong, C.~Zhang, W.~Ding, K.~Ma, G.~Chen, R.~Cai, D.~Xiang, H.~Qu, F.~Luo, H.~Ye, and L.~Chen, ``Mimo-embodied: X-embodied foundation model technical report,'' 2025.

\bibitem{weng2025vision}
X.~Weng, C.~Pang, and G.-S. Xia, ``Vision-language modeling meets remote sensing: Models, datasets, and perspectives,'' \emph{IEEE Geoscience and Remote Sensing Magazine}, 2025.

\bibitem{xue2025regression}
X.~Xue and X.~X. Zhu, ``Regression in earth observation: Are vision--language models up to the challenge?'' \emph{IEEE Geoscience and Remote Sensing Magazine}, 2025.

\bibitem{achiam2023gpt}
J.~Achiam, S.~Adler, S.~Agarwal, L.~Ahmad, I.~Akkaya, F.~L. Aleman, D.~Almeida, J.~Altenschmidt, S.~Altman, S.~Anadkat \emph{et~al.}, ``Gpt-4 technical report,'' \emph{arXiv preprint arXiv:2303.08774}, 2023.

\bibitem{chen2023minigpt}
J.~Chen, D.~Zhu, X.~Shen, X.~Li, Z.~Liu, P.~Zhang, R.~Krishnamoorthi, V.~Chandra, Y.~Xiong, and M.~Elhoseiny, ``Minigpt-v2: large language model as a unified interface for vision-language multi-task learning,'' \emph{arXiv preprint arXiv:2310.09478}, 2023.

\bibitem{bai2023qwen}
J.~Bai, S.~Bai, S.~Yang, S.~Wang, S.~Tan, P.~Wang, J.~Lin, C.~Zhou, and J.~Zhou, ``Qwen-vl: A frontier large vision-language model with versatile abilities,'' \emph{arXiv preprint arXiv:2308.12966}, 2023.

\bibitem{chen2023internvl}
Z.~Chen, J.~Wu, W.~Wang, W.~Su, G.~Chen, S.~Xing, M.~Zhong, Q.~Zhang, X.~Zhu, L.~Lu \emph{et~al.}, ``Internvl: Scaling up vision foundation models and aligning for generic visual-linguistic tasks,'' in \emph{Proceedings of the IEEE/CVF conference on computer vision and pattern recognition}, 2024, pp. 24\,185--24\,198.

\bibitem{lu2024visual}
M.~Y. Lu, B.~Chen, D.~F. Williamson, R.~J. Chen, I.~Liang, T.~Ding, G.~Jaume, I.~Odintsov, L.~P. Le, G.~Gerber \emph{et~al.}, ``A visual-language foundation model for computational pathology,'' \emph{Nature medicine}, vol.~30, no.~3, pp. 863--874, 2024.

\bibitem{chen2026overview}
H.~Chen, H.~Chen, Z.~Zhao, K.~Han, G.~Zhu, Y.~Zhao, Y.~Du, W.~Xu, and Q.~Shi, ``An overview of domain-specific foundation model: key technologies, applications and challenges,'' \emph{Science China Information Sciences}, vol.~69, no.~1, p. 111301, 2026.

\bibitem{schuhmann2022laion}
C.~Schuhmann, R.~Beaumont, R.~Vencu, C.~Gordon, R.~Wightman, M.~Cherti, T.~Coombes, A.~Katta, C.~Mullis, M.~Wortsman \emph{et~al.}, ``Laion-5b: An open large-scale dataset for training next generation image-text models,'' \emph{Advances in neural information processing systems}, vol.~35, pp. 25\,278--25\,294, 2022.

\bibitem{luo2023lexlip}
Z.~Luo, P.~Zhao, C.~Xu, X.~Geng, T.~Shen, C.~Tao, J.~Ma, Q.~Lin, and D.~Jiang, ``Lexlip: Lexicon-bottlenecked language-image pre-training for large-scale image-text sparse retrieval,'' in \emph{Proceedings of the IEEE/CVF international conference on computer vision}, 2023, pp. 11\,206--11\,217.

\bibitem{yu2024capsfusion}
Q.~Yu, Q.~Sun, X.~Zhang, Y.~Cui, F.~Zhang, Y.~Cao, X.~Wang, and J.~Liu, ``Capsfusion: Rethinking image-text data at scale,'' in \emph{Proceedings of the IEEE/CVF Conference on Computer Vision and Pattern Recognition}, 2024, pp. 14\,022--14\,032.

\bibitem{liu2024remoteclip}
F.~Liu, D.~Chen, Z.~Guan, X.~Zhou, J.~Zhu, Q.~Ye, L.~Fu, and J.~Zhou, ``Remoteclip: A vision language foundation model for remote sensing,'' \emph{IEEE Transactions on Geoscience and Remote Sensing}, vol.~62, pp. 1--16, 2024.

\bibitem{liu2025text2earth}
C.~Liu, K.~Chen, R.~Zhao, Z.~Zou, and Z.~Shi, ``Text2earth: Unlocking text-driven remote sensing image generation with a global-scale dataset and a foundation model,'' \emph{IEEE Geoscience and Remote Sensing Magazine}, 2025.

\bibitem{wang2025dynamic}
H.~Wang, Y.~Nie, Y.~Ye, Y.~Wang, S.~Li, H.~Yu, J.~Lu, and C.~Huang, ``Dynamic-vlm: Simple dynamic visual token compression for videollm,'' in \emph{Proceedings of the IEEE/CVF International Conference on Computer Vision}, 2025, pp. 20\,812--20\,823.

\bibitem{zhang2025critic}
D.~Zhang, J.~Lei, J.~Li, X.~Wang, Y.~Liu, Z.~Yang, J.~Li, W.~Wang, S.~Yang, J.~Wu \emph{et~al.}, ``Critic-v: Vlm critics help catch vlm errors in multimodal reasoning,'' in \emph{Proceedings of the Computer Vision and Pattern Recognition Conference}, 2025, pp. 9050--9061.

\bibitem{prabhu2025trust}
V.~Prabhu, S.~Purushwalkam, A.~Yan, C.~Xiong, and R.~Xu, ``Trust but verify: Programmatic vlm evaluation in the wild,'' in \emph{Proceedings of the IEEE/CVF International Conference on Computer Vision}, 2025, pp. 3258--3267.

\bibitem{wu2024visionllm}
J.~Wu, M.~Zhong, S.~Xing, Z.~Lai, Z.~Liu, Z.~Chen, W.~Wang, X.~Zhu, L.~Lu, T.~Lu \emph{et~al.}, ``Visionllm v2: An end-to-end generalist multimodal large language model for hundreds of vision-language tasks,'' \emph{Advances in Neural Information Processing Systems}, vol.~37, pp. 69\,925--69\,975, 2024.

\bibitem{truong2025insect}
T.-D. Truong, H.-Q. Nguyen, X.-B. Nguyen, A.~Dowling, X.~Li, and K.~Luu, ``Insect-foundation: A foundation model and large multimodal dataset for vision-language insect understanding,'' \emph{International Journal of Computer Vision}, vol. 133, no.~10, pp. 7128--7153, 2025.

\bibitem{xu2024lvlm}
P.~Xu, W.~Shao, K.~Zhang, P.~Gao, S.~Liu, M.~Lei, F.~Meng, S.~Huang, Y.~Qiao, and P.~Luo, ``Lvlm-ehub: A comprehensive evaluation benchmark for large vision-language models,'' \emph{IEEE Transactions on Pattern Analysis and Machine Intelligence}, 2024.

\bibitem{lobry2020rsvqa}
S.~Lobry, D.~Marcos, J.~Murray, and D.~Tuia, ``Rsvqa: Visual question answering for remote sensing data,'' \emph{IEEE Transactions on Geoscience and Remote Sensing}, vol.~58, no.~12, pp. 8555--8566, 2020.

\bibitem{kuckreja2024geochat}
K.~Kuckreja, M.~S. Danish, M.~Naseer, A.~Das, S.~Khan, and F.~S. Khan, ``Geochat: Grounded large vision-language model for remote sensing,'' in \emph{Proceedings of the IEEE/CVF Conference on Computer Vision and Pattern Recognition}, 2024, pp. 27\,831--27\,840.

\bibitem{wei2025hq}
Z.~Wei, G.~Wang, X.~Ma, K.~Mei, H.~Chen, Y.~Jin, and F.~Rao, ``Hq-clip: Leveraging large vision-language models to create high-quality image-text datasets and clip models,'' in \emph{Proceedings of the IEEE/CVF International Conference on Computer Vision}, 2025, pp. 22\,447--22\,456.

\bibitem{chen2024vitamin}
J.~Chen, Q.~Yu, X.~Shen, A.~Yuille, and L.-C. Chen, ``Vitamin: Designing scalable vision models in the vision-language era,'' in \emph{Proceedings of the IEEE/CVF Conference on Computer Vision and Pattern Recognition}, 2024, pp. 12\,954--12\,966.

\bibitem{dai2023instructblip}
W.~Dai, J.~Li, D.~Li, A.~Tiong, J.~Zhao, W.~Wang, B.~Li, P.~N. Fung, and S.~Hoi, ``Instructblip: Towards general-purpose vision-language models with instruction tuning,'' \emph{Advances in neural information processing systems}, vol.~36, pp. 49\,250--49\,267, 2023.

\bibitem{pol2024hugging}
U.~R. Pol, P.~S. Vadar, and T.~T. Moharekar, ``Hugging face: revolutionizing ai and nlp,'' \emph{International Journal for Research in Applied Science and Engineering Technology}, vol.~12, no.~8, pp. 1121--1124, 2024.

\bibitem{xia2018dota}
G.-S. Xia, X.~Bai, J.~Ding, Z.~Zhu, S.~Belongie, J.~Luo, M.~Datcu, M.~Pelillo, and L.~Zhang, ``Dota: A large-scale dataset for object detection in aerial images,'' in \emph{Proceedings of the IEEE conference on computer vision and pattern recognition}, 2018, pp. 3974--3983.

\bibitem{li2020object}
K.~Li, G.~Wan, G.~Cheng, L.~Meng, and J.~Han, ``Object detection in optical remote sensing images: A survey and a new benchmark,'' \emph{ISPRS journal of photogrammetry and remote sensing}, vol. 159, pp. 296--307, 2020.

\bibitem{sun2022fair1m}
X.~Sun, P.~Wang, Z.~Yan, F.~Xu, R.~Wang, W.~Diao, J.~Chen, J.~Li, Y.~Feng, T.~Xu \emph{et~al.}, ``Fair1m: A benchmark dataset for fine-grained object recognition in high-resolution remote sensing imagery,'' \emph{ISPRS Journal of Photogrammetry and Remote Sensing}, vol. 184, pp. 116--130, 2022.

\bibitem{yang2022detecting}
X.~Yang, G.~Zhang, X.~Yang, Y.~Zhou, W.~Wang, J.~Tang, T.~He, and J.~Yan, ``Detecting rotated objects as gaussian distributions and its 3-d generalization,'' \emph{IEEE Transactions on Pattern Analysis and Machine Intelligence}, vol.~45, no.~4, pp. 4335--4354, 2022.

\bibitem{zhou2022mmrotate}
Y.~Zhou, X.~Yang, G.~Zhang, J.~Wang, Y.~Liu, L.~Hou, X.~Jiang, X.~Liu, J.~Yan, C.~Lyu \emph{et~al.}, ``Mmrotate: A rotated object detection benchmark using pytorch,'' in \emph{Proceedings of the 30th ACM International Conference on Multimedia}, 2022, pp. 7331--7334.

\bibitem{yang2021r3det}
X.~Yang, J.~Yan, Z.~Feng, and T.~He, ``R3det: Refined single-stage detector with feature refinement for rotating object,'' in \emph{Proceedings of the AAAI conference on artificial intelligence}, vol.~35, no.~4, 2021, pp. 3163--3171.

\bibitem{su2017local}
T.~Su and S.~Zhang, ``Local and global evaluation for remote sensing image segmentation,'' \emph{ISPRS Journal of Photogrammetry and Remote Sensing}, vol. 130, pp. 256--276, 2017.

\bibitem{volpi2018deep}
M.~Volpi and D.~Tuia, ``Deep multi-task learning for a geographically-regularized semantic segmentation of aerial images,'' \emph{ISPRS journal of photogrammetry and remote sensing}, vol. 144, pp. 48--60, 2018.

\bibitem{wang2023samrs}
D.~Wang, J.~Zhang, B.~Du, M.~Xu, L.~Liu, D.~Tao, and L.~Zhang, ``Samrs: Scaling-up remote sensing segmentation dataset with segment anything model,'' \emph{Advances in Neural Information Processing Systems}, vol.~36, pp. 8815--8827, 2023.

\bibitem{he2023multimodal}
Q.~He, X.~Sun, W.~Diao, Z.~Yan, F.~Yao, and K.~Fu, ``Multimodal remote sensing image segmentation with intuition-inspired hypergraph modeling,'' \emph{IEEE Transactions on Image Processing}, vol.~32, pp. 1474--1487, 2023.

\bibitem{li2024review}
J.~Li, Y.~Cai, Q.~Li, M.~Kou, and T.~Zhang, ``A review of remote sensing image segmentation by deep learning methods,'' \emph{International Journal of Digital Earth}, vol.~17, no.~1, p. 2328827, 2024.

\bibitem{liu2023visual}
H.~Liu, C.~Li, Q.~Wu, and Y.~J. Lee, ``Visual instruction tuning,'' \emph{Advances in neural information processing systems}, vol.~36, pp. 34\,892--34\,916, 2023.

\bibitem{zhang2023instruction}
S.~Zhang, L.~Dong, X.~Li, S.~Zhang, X.~Sun, S.~Wang, J.~Li, R.~Hu, T.~Zhang, G.~Wang \emph{et~al.}, ``Instruction tuning for large language models: A survey,'' \emph{ACM Computing Surveys}, 2023.

\bibitem{shi2024instruction}
Z.~Shi, A.~X. Yang, B.~Wu, L.~Aitchison, E.~Yilmaz, and A.~Lipani, ``Instruction tuning with loss over instructions,'' \emph{Advances in Neural Information Processing Systems}, vol.~37, pp. 69\,176--69\,205, 2024.

\bibitem{liu2024improved}
H.~Liu, C.~Li, Y.~Li, and Y.~J. Lee, ``Improved baselines with visual instruction tuning,'' in \emph{Proceedings of the IEEE/CVF conference on computer vision and pattern recognition}, 2024, pp. 26\,296--26\,306.

\bibitem{deng2025changechat}
P.~Deng, W.~Zhou, and H.~Wu, ``Changechat: An interactive model for remote sensing change analysis via multimodal instruction tuning,'' in \emph{ICASSP 2025-2025 IEEE International Conference on Acoustics, Speech and Signal Processing (ICASSP)}.\hskip 1em plus 0.5em minus 0.4em\relax IEEE, 2025, pp. 1--5.

\bibitem{gupta2023visual}
T.~Gupta and A.~Kembhavi, ``Visual programming: Compositional visual reasoning without training,'' in \emph{Proceedings of the IEEE/CVF conference on computer vision and pattern recognition}, 2023, pp. 14\,953--14\,962.

\bibitem{malkinski2023review}
M.~Ma{\l}ki{\'n}ski and J.~Ma{\'n}dziuk, ``A review of emerging research directions in abstract visual reasoning,'' \emph{Information Fusion}, vol.~91, pp. 713--736, 2023.

\bibitem{chen2023large}
L.~Chen, B.~Li, S.~Shen, J.~Yang, C.~Li, K.~Keutzer, T.~Darrell, and Z.~Liu, ``Large language models are visual reasoning coordinators,'' \emph{Advances in Neural Information Processing Systems}, vol.~36, pp. 70\,115--70\,140, 2023.

\bibitem{chen2024visual}
Z.~Chen, Q.~Zhou, Y.~Shen, Y.~Hong, Z.~Sun, D.~Gutfreund, and C.~Gan, ``Visual chain-of-thought prompting for knowledge-based visual reasoning,'' in \emph{Proceedings of the AAAI Conference on Artificial Intelligence}, vol.~38, no.~2, 2024, pp. 1254--1262.

\bibitem{huang2024vtimellm}
B.~Huang, X.~Wang, H.~Chen, Z.~Song, and W.~Zhu, ``Vtimellm: Empower llm to grasp video moments,'' in \emph{Proceedings of the IEEE/CVF Conference on Computer Vision and Pattern Recognition}, 2024, pp. 14\,271--14\,280.

\bibitem{nie2024slowfocus}
M.~Nie, D.~Ding, C.~Wang, Y.~Guo, J.~Han, H.~Xu, and L.~Zhang, ``Slowfocus: Enhancing fine-grained temporal understanding in video llm,'' \emph{Advances in Neural Information Processing Systems}, vol.~37, pp. 81\,808--81\,835, 2024.

\bibitem{shu2025audio}
F.~Shu, L.~Zhang, H.~Jiang, and C.~Xie, ``Audio-visual llm for video understanding,'' in \emph{Proceedings of the IEEE/CVF International Conference on Computer Vision}, 2025, pp. 4246--4255.

\bibitem{yuan2025videorefer}
Y.~Yuan, H.~Zhang, W.~Li, Z.~Cheng, B.~Zhang, L.~Li, X.~Li, D.~Zhao, W.~Zhang, Y.~Zhuang \emph{et~al.}, ``Videorefer suite: Advancing spatial-temporal object understanding with video llm,'' in \emph{Proceedings of the Computer Vision and Pattern Recognition Conference}, 2025, pp. 18\,970--18\,980.

\bibitem{wang2025videotree}
Z.~Wang, S.~Yu, E.~Stengel-Eskin, J.~Yoon, F.~Cheng, G.~Bertasius, and M.~Bansal, ``Videotree: Adaptive tree-based video representation for llm reasoning on long videos,'' in \emph{Proceedings of the Computer Vision and Pattern Recognition Conference}, 2025, pp. 3272--3283.

\bibitem{hou2025geocogent}
S.~Hou, H.~Jiao, J.~Liang, Z.~Shen, A.~Zhao, and H.~Wu, ``Geocogent: an llm-based agent for geospatial code generation,'' \emph{International Journal of Geographical Information Science}, pp. 1--34, 2025.

\bibitem{zhang2024geospatial}
Y.~Zhang, J.~Li, Z.~Wang, Z.~He, Q.~Guan, J.~Lin, and W.~Yu, ``Geospatial large language model trained with a simulated environment for generating tool-use chains autonomously,'' \emph{International Journal of Applied Earth Observation and Geoinformation}, p. 104312, 2024.

\bibitem{yu2025monkuu}
C.~Yu, Y.~Yao, X.~Zhang, G.~Zhu, Y.~Guo, X.~Shao, M.~Shibasaki, Z.~Hu, L.~Dai, Q.~Guan \emph{et~al.}, ``Monkuu: a llm-powered natural language interface for geospatial databases with dynamic schema mapping,'' \emph{International Journal of Geographical Information Science}, pp. 1--22, 2025.

\bibitem{singh2024geollm}
S.~Singh, M.~Fore, and D.~Stamoulis, ``Geollm-engine: A realistic environment for building geospatial copilots,'' in \emph{Proceedings of the IEEE/CVF Conference on Computer Vision and Pattern Recognition}, 2024, pp. 585--594.

\bibitem{zhu2025ger}
H.~Zhu, Z.~Li, and J.~Jin, ``Ger-llm: Efficient and effective geospatial entity resolution with large language model,'' in \emph{Proceedings of the 2025 Conference on Empirical Methods in Natural Language Processing}, 2025, pp. 23\,272--23\,288.

\bibitem{scheider2023geoai}
S.~Scheider and K.-F. Richter, ``Geoai,'' \emph{KI-K{\"u}nstliche Intelligenz}, vol.~37, no.~1, pp. 5--9, 2023.

\bibitem{janowicz2020geoai}
K.~Janowicz, S.~Gao, G.~McKenzie, Y.~Hu, and B.~Bhaduri, ``Geoai: spatially explicit artificial intelligence techniques for geographic knowledge discovery and beyond,'' pp. 625--636, 2020.

\bibitem{mai2022review}
G.~Mai, K.~Janowicz, Y.~Hu, S.~Gao, B.~Yan, R.~Zhu, L.~Cai, and N.~Lao, ``A review of location encoding for geoai: methods and applications,'' \emph{International Journal of Geographical Information Science}, vol.~36, no.~4, pp. 639--673, 2022.

\bibitem{zhou2024towards}
Y.~Zhou, L.~Feng, Y.~Ke, X.~Jiang, J.~Yan, X.~Yang, and W.~Zhang, ``Towards vision-language geo-foundation model: A survey,'' \emph{arXiv preprint arXiv:2406.09385}, 2024.

\bibitem{zhang2024geoscience}
H.~Zhang, J.-J. Xu, H.-W. Cui, L.~Li, Y.~Yang, C.-S. Tang, and N.~Boers, ``When geoscience meets foundation models: Toward a general geoscience artificial intelligence system,'' \emph{IEEE geoscience and remote sensing magazine}, 2024.

\bibitem{xiao2024foundation}
A.~Xiao, W.~Xuan, J.~Wang, J.~Huang, D.~Tao, S.~Lu, and N.~Yokoya, ``Foundation models for remote sensing and earth observation: A survey,'' \emph{IEEE Geoscience and Remote Sensing Magazine}, 2025.

\bibitem{lecun2015deep}
Y.~LeCun, Y.~Bengio, and G.~Hinton, ``Deep learning,'' \emph{nature}, vol. 521, no. 7553, pp. 436--444, 2015.

\bibitem{rusk2016deep}
N.~Rusk, ``Deep learning,'' \emph{Nature Methods}, vol.~13, no.~1, pp. 35--35, 2016.

\bibitem{zhang2016deep}
L.~Zhang, L.~Zhang, and B.~Du, ``Deep learning for remote sensing data: A technical tutorial on the state of the art,'' \emph{IEEE Geoscience and remote sensing magazine}, vol.~4, no.~2, pp. 22--40, 2016.

\bibitem{li2018deep}
Y.~Li, H.~Zhang, X.~Xue, Y.~Jiang, and Q.~Shen, ``Deep learning for remote sensing image classification: A survey,'' \emph{Wiley Interdisciplinary Reviews: Data Mining and Knowledge Discovery}, vol.~8, no.~6, p. e1264, 2018.

\bibitem{li2022deep}
J.~Li, D.~Hong, L.~Gao, J.~Yao, K.~Zheng, B.~Zhang, and J.~Chanussot, ``Deep learning in multimodal remote sensing data fusion: A comprehensive review,'' \emph{International Journal of Applied Earth Observation and Geoinformation}, vol. 112, p. 102926, 2022.

\bibitem{lai2023lisa}
X.~Lai, Z.~Tian, Y.~Chen, Y.~Li, Y.~Yuan, S.~Liu, and J.~Jia, ``Lisa: Reasoning segmentation via large language model,'' in \emph{Proceedings of the IEEE/CVF conference on computer vision and pattern recognition}, 2024, pp. 9579--9589.

\bibitem{zhang2023next}
A.~Zhang, L.~Zhao, C.-W. Xie, Y.~Zheng, W.~Ji, and T.-S. Chua, ``Next-chat: An lmm for chat, detection and segmentation,'' \emph{arXiv preprint arXiv:2311.04498}, 2023.

\bibitem{wang2024visionllm}
W.~Wang, Z.~Chen, X.~Chen, J.~Wu, X.~Zhu, G.~Zeng, P.~Luo, T.~Lu, J.~Zhou, Y.~Qiao \emph{et~al.}, ``Visionllm: Large language model is also an open-ended decoder for vision-centric tasks,'' \emph{Advances in Neural Information Processing Systems}, vol.~36, 2024.

\bibitem{uchida2024using}
S.~Uchida, ``Using early llms for corpus linguistics: Examining chatgpt's potential and limitations,'' \emph{Applied Corpus Linguistics}, vol.~4, no.~1, p. 100089, 2024.

\bibitem{yu2025opencsg}
Y.~Yu, Z.~Dai, Z.~Wang, W.~Wang, R.~Chen, and J.~Pei, ``Opencsg chinese corpus: A series of high-quality chinese datasets for llm training,'' \emph{arXiv preprint arXiv:2501.08197}, 2025.

\bibitem{mall2023remote}
U.~K. Mall, C.~P. Phoo, M.~Liu, C.~Vondrick, B.~Hariharan, and K.~Bala, ``Remote sensing vision-language foundation models without annotations via ground remote alignment,'' in \emph{International Conference on Learning Representations}, vol. 2024, 2024, pp. 49\,294--49\,314.

\bibitem{xu2024benchmarking}
Z.~Xu, Y.~Zhu, S.~Deng, A.~Mittal, Y.~Chen, M.~Wang, P.~Favaro, J.~Tighe, and D.~Modolo, ``Benchmarking zero-shot recognition with vision-language models: Challenges on granularity and specificity,'' in \emph{Proceedings of the IEEE/CVF Conference on Computer Vision and Pattern Recognition}, 2024, pp. 1827--1836.

\bibitem{wang2024qwen2}
P.~Wang, S.~Bai, S.~Tan, S.~Wang, Z.~Fan, J.~Bai, K.~Chen, X.~Liu, J.~Wang, W.~Ge \emph{et~al.}, ``Qwen2-vl: Enhancing vision-language model's perception of the world at any resolution,'' \emph{arXiv preprint arXiv:2409.12191}, 2024.

\bibitem{xiong2022earthnets}
Z.~Xiong, F.~Zhang, Y.~Wang, Y.~Shi, and X.~X. Zhu, ``Earthnets: Empowering ai in earth observation,'' \emph{arXiv preprint arXiv:2210.04936}, 2022.

\bibitem{wang2025disasterm3}
J.~Wang, W.~Xuan, H.~Qi, Z.~Liu, K.~Liu, Y.~Wu, H.~Chen, J.~Song, J.~Xia, Z.~Zheng \emph{et~al.}, ``Disasterm3: A remote sensing vision-language dataset for disaster damage assessment and response,'' \emph{Advances in Neural Information Processing Systems}, vol.~38, 2026.

\bibitem{wu2025fit}
H.~Wu, Z.~Du, D.~Zhong, Y.~Wang, and C.~Tao, ``Fsvlm: A vision-language model for remote sensing farmland segmentation,'' \emph{IEEE Transactions on Geoscience and Remote Sensing}, vol.~63, pp. 1--13, 2025.

\bibitem{wang2026cityset}
J.~Wang, W.~Xuan, H.~Qi, Z.~Chen, H.~Chen, Z.~Zheng, J.~Xia, Y.~Zhong, and N.~Yokoya, ``Cityvlm: Towards sustainable urban development via multi-view coordinated vision–language model,'' \emph{ISPRS Journal of Photogrammetry and Remote Sensing}, vol. 232, pp. 62--74, 2026.

\bibitem{lin2024data}
X.~Lin, W.~Wang, Y.~Li, S.~Yang, F.~Feng, Y.~Wei, and T.-S. Chua, ``Data-efficient fine-tuning for llm-based recommendation,'' in \emph{Proceedings of the 47th international ACM SIGIR conference on research and development in information retrieval}, 2024, pp. 365--374.

\bibitem{zheng2025learning}
H.~Zheng, L.~Shen, A.~Tang, Y.~Luo, H.~Hu, B.~Du, Y.~Wen, and D.~Tao, ``Learning from models beyond fine-tuning,'' \emph{Nature Machine Intelligence}, vol.~7, no.~1, pp. 6--17, 2025.

\bibitem{anisuzzaman2024fine}
D.~Anisuzzaman, J.~G. Malins, P.~A. Friedman, and Z.~I. Attia, ``Fine-tuning llms for specialized use cases,'' \emph{Mayo Clinic Proceedings: Digital Health}, 2024.

\bibitem{tian2024hydralora}
C.~Tian, Z.~Shi, Z.~Guo, L.~Li, and C.-Z. Xu, ``Hydralora: An asymmetric lora architecture for efficient fine-tuning,'' \emph{Advances in Neural Information Processing Systems}, vol.~37, pp. 9565--9584, 2024.

\bibitem{bian2025lora}
J.~Bian, L.~Wang, L.~Zhang, and J.~Xu, ``Lora-fair: Federated lora fine-tuning with aggregation and initialization refinement,'' in \emph{Proceedings of the IEEE/CVF International Conference on Computer Vision}, 2025, pp. 3737--3746.

\bibitem{NEURIPS2024_fe2fc7dc}
T.~Lee, H.~Tu, C.~H. Wong, W.~Zheng, Y.~Zhou, Y.~Mai, J.~S. Roberts, M.~Yasunaga, H.~Yao, C.~Xie, and P.~Liang, ``Vhelm: A holistic evaluation of vision language models,'' in \emph{Advances in Neural Information Processing Systems}, A.~Globerson, L.~Mackey, D.~Belgrave, A.~Fan, U.~Paquet, J.~Tomczak, and C.~Zhang, Eds., vol.~37.\hskip 1em plus 0.5em minus 0.4em\relax Curran Associates, Inc., 2024, pp. 140\,632--140\,666.

\bibitem{shool2025systematic}
S.~Shool, S.~Adimi, R.~Saboori~Amleshi, E.~Bitaraf, R.~Golpira, and M.~Tara, ``A systematic review of large language model (llm) evaluations in clinical medicine,'' \emph{BMC Medical Informatics and Decision Making}, vol.~25, no.~1, p. 117, 2025.

\bibitem{hu2023mdas}
J.~Hu, R.~Liu, D.~Hong, A.~Camero, J.~Yao, M.~Schneider, F.~Kurz, K.~Segl, and X.~X. Zhu, ``Mdas: A new multimodal benchmark dataset for remote sensing,'' \emph{Earth System Science Data}, vol.~15, no.~1, pp. 113--131, 2023.

\bibitem{hong2023cross}
D.~Hong, B.~Zhang, H.~Li, Y.~Li, J.~Yao, C.~Li, M.~Werner, J.~Chanussot, A.~Zipf, and X.~X. Zhu, ``Cross-city matters: A multimodal remote sensing benchmark dataset for cross-city semantic segmentation using high-resolution domain adaptation networks,'' \emph{Remote Sensing of Environment}, vol. 299, p. 113856, 2023.

\bibitem{hossin2015review}
M.~Hossin and M.~N. Sulaiman, ``A review on evaluation metrics for data classification evaluations,'' \emph{International journal of data mining \& knowledge management process}, vol.~5, no.~2, p.~1, 2015.

\bibitem{dey2020robust}
E.~K. Dey and M.~Awrangjeb, ``A robust performance evaluation metric for extracted building boundaries from remote sensing data,'' \emph{IEEE Journal of Selected Topics in Applied Earth Observations and Remote Sensing}, vol.~13, pp. 4030--4043, 2020.

\bibitem{naidu2023review}
G.~Naidu, T.~Zuva, and E.~M. Sibanda, ``A review of evaluation metrics in machine learning algorithms,'' in \emph{Computer science on-line conference}.\hskip 1em plus 0.5em minus 0.4em\relax Springer, 2023, pp. 15--25.

\bibitem{soni2024earthdial}
S.~Soni, A.~Dudhane, H.~Debary, M.~Fiaz, M.~A. Munir, M.~S. Danish, P.~Fraccaro, C.~D. Watson, L.~J. Klein, F.~S. Khan \emph{et~al.}, ``Earthdial: Turning multi-sensory earth observations to interactive dialogues,'' in \emph{Proceedings of the Computer Vision and Pattern Recognition Conference}, 2025, pp. 14\,303--14\,313.

\bibitem{ye2025street2sat}
J.~Ye, J.~He, X.~Zhang, Y.~Lin, H.~Lin, C.~He, and W.~Li, ``Satellite image synthesis from street view with fine-grained spatial textual guidance: A novel framework,'' \emph{IEEE Geoscience and Remote Sensing Magazine}, vol.~13, no.~3, pp. 395--414, 2025.

\bibitem{ye2024cross}
J.~Ye, H.~Lin, L.~Ou, D.~Chen, Z.~Wang, Q.~Zhu, C.~He, and W.~Li, ``Where am i? cross-view geo-localization with natural language descriptions,'' in \emph{Proceedings of the IEEE/CVF International Conference on Computer Vision}, 2025, pp. 5890--5900.

\bibitem{yao2025falcon}
K.~Yao, N.~Xu, R.~Yang, Y.~Xu, Z.~Gao, T.~Kitrungrotsakul, Y.~Ren, P.~Zhang, J.~Wang, N.~Wei \emph{et~al.}, ``Falcon: A remote sensing vision-language foundation model,'' \emph{arXiv preprint arXiv:2503.11070}, 2025.

\bibitem{yang2025fusarklip}
Y.~Yang, X.~Zhang, Q.~Fang, J.~Liu, Z.~Ye, R.~Li, L.~Liu, and H.~Wang, ``Fusar-klip: Towards multimodal foundation models for remote sensing,'' 2025.

\bibitem{zavras2025gaia}
A.~Zavras, D.~Michail, X.~X. Zhu, B.~Demir, and I.~Papoutsis, ``Gaia: A global, multimodal, multiscale vision--language dataset for remote sensing image analysis,'' \emph{IEEE Geoscience and Remote Sensing Magazine}, 2026.

\bibitem{xiong2025geolangbind}
Z.~Xiong, Y.~Wang, W.~Yu, A.~J. Stewart, J.~Zhao, N.~Lehmann, T.~Dujardin, Z.~Yuan, P.~Ghamisi, and X.~X. Zhu, ``Geolangbind: Unifying earth observation with agglomerative vision-language foundation models,'' \emph{arXiv preprint arXiv:2503.06312}, 2025.

\bibitem{lu2025gleam}
X.~Lu, Z.~Zheng, Y.~Wan, Y.~Yao, A.~Wang, R.~Zhang, P.~Xia, Q.~Wu, Q.~Li, W.~Lin, X.~Zhao, P.~Ma, X.~Yang, and H.~Li, ``Gleam: Learning to match and explain in cross-view geo-localization,'' 2025.

\bibitem{czerkawski2023laion}
M.~Czerkawski and A.~Francis, ``From laion-5b to laion-eo: Filtering billions of images using anchor datasets for satellite image extraction,'' \emph{arXiv preprint arXiv:2309.15535}, 2023.

\bibitem{muhtar2024lhrs}
D.~Muhtar, Z.~Li, F.~Gu, X.~Zhang, and P.~Xiao, ``Lhrs-bot: Empowering remote sensing with vgi-enhanced large multimodal language model,'' in \emph{European Conference on Computer Vision}.\hskip 1em plus 0.5em minus 0.4em\relax Springer, 2024, pp. 440--457.

\bibitem{li2025lhrsbotnova}
Z.~Li, D.~Muhtar, F.~Gu, Y.~He, X.~Zhang, P.~Xiao, G.~He, and X.~Zhu, ``Lhrs-bot-nova: Improved multimodal large language model for remote sensing vision-language interpretation,'' \emph{ISPRS Journal of Photogrammetry and Remote Sensing}, vol. 227, pp. 539--550, 2025.

\bibitem{muhtar2025scorers}
D.~Muhtar, E.~Zhang, Z.~Li, F.~Gu, Y.~He, P.~Xiao, and X.~Zhang, ``Quality-driven curation of remote sensing vision-language data via learned scoring models,'' \emph{Advances in Neural Information Processing Systems}, vol.~38, pp. 171\,088--171\,133, 2026.

\bibitem{si2025spex}
D.~Si, D.~Wang, E.~Gao, X.~Qin, L.~Zhao, J.~Zhang, M.~Xu, J.~Zhan, J.~Wang, L.~Liu \emph{et~al.}, ``Spex: a vision-language model for land cover extraction on spectral remote sensing images,'' \emph{IEEE Transactions on Geoscience and Remote Sensing}, 2026.

\bibitem{marimo2025beyond}
C.~T. Marimo, B.~Blumenstiel, M.~Nitsche, J.~Jakubik, and T.~Brunschwiler, ``Beyond the visible: Multispectral vision-language learning for earth observation,'' in \emph{Joint European Conference on Machine Learning and Knowledge Discovery in Databases}.\hskip 1em plus 0.5em minus 0.4em\relax Springer, 2025, pp. 359--375.

\bibitem{ma2025landsat30au}
S.~Ma, Z.~Li, and J.~A. Taylor, ``Landsat30-au: A vision-language dataset for australian landsat imagery,'' in \emph{Proceedings of the AAAI Conference on Artificial Intelligence}, vol.~40, no.~10, 2026, pp. 7809--7817.

\bibitem{chen2025lrsclip}
W.~Chen, J.~Chen, Y.~Deng, J.~Chen, Y.~Feng, Z.~Xi, D.~Liu, K.~Li, and Y.~Meng, ``Lrsclip: A vision-language foundation model for aligning remote sensing image with longer text,'' \emph{arXiv preprint arXiv:2503.19311}, 2025.

\bibitem{zhang2024earthgpt}
W.~Zhang, M.~Cai, T.~Zhang, Y.~Zhuang, and X.~Mao, ``Earthgpt: A universal multimodal large language model for multisensor image comprehension in remote sensing domain,'' \emph{IEEE Transactions on Geoscience and Remote Sensing}, vol.~62, pp. 1--20, 2024.

\bibitem{luo2024mmm}
J.~Luo, Y.~Wang, Z.~Gu, Y.~Qiu, S.~Yao, F.~Wang, C.~Xu, W.~Zhang, D.~Wang, and Z.~Cui, ``Mmm-rs: a multi-modal, multi-gsd, multi-scene remote sensing dataset and benchmark for text-to-image generation,'' in \emph{Proceedings of the 38th International Conference on Neural Information Processing Systems}, 2024, pp. 12\,151--12\,163.

\bibitem{pan2024moco}
L.~Pan, C.~Song, X.~Gan, K.~Xu, and Y.~Xie, ``Military image captioning for low-altitude uav or ugv perspectives,'' \emph{Drones}, vol.~8, no.~9, 2024.

\bibitem{wang2025sarclip}
P.~Wang, Z.~Lu, Y.~Li, B.~Ding, and D.~Zhang, ``Sarclip: The first vision–language foundation model for sar image,'' \emph{IEEE Transactions on Geoscience and Remote Sensing}, vol.~63, pp. 1--11, 2025.

\bibitem{zhang2025earthgpt}
W.~Zhang, M.~Cai, Y.~Ning, T.~Zhang, Y.~Zhuang, H.~Chen, J.~Li, and X.~Mao, ``Earthgpt-x: Enabling mllms to flexibly and comprehensively understand multi-source remote sensing imagery,'' \emph{arXiv preprint arXiv:2504.12795}, 2025.

\bibitem{cheng2022nwpu-cap}
Q.~Cheng, H.~Huang, Y.~Xu, Y.~Zhou, H.~Li, and Z.~Wang, ``Nwpu-captions dataset and mlca-net for remote sensing image captioning,'' \emph{IEEE Transactions on Geoscience and Remote Sensing}, vol.~60, pp. 1--19, 2022.

\bibitem{li2025reobench}
X.~Li, Y.~Tao, S.~Zhang, S.~Liu, Z.~Xiong, C.~Luo, L.~Liu, M.~Pechenizkiy, X.~Zhu, and T.~Huang, ``Reobench: Benchmarking robustness of earth observation foundation models,'' \emph{Advances in Neural Information Processing Systems}, vol.~38, 2026.

\bibitem{zhang2024rs5m}
Z.~Zhang, T.~Zhao, Y.~Guo, and J.~Yin, ``Rs5m and georsclip: A large-scale vision- language dataset and a large vision-language model for remote sensing,'' \emph{IEEE Transactions on Geoscience and Remote Sensing}, vol.~62, pp. 1--23, 2024.

\bibitem{lu2018rsicd}
X.~Lu, B.~Wang, X.~Zheng, and X.~Li, ``Exploring models and data for remote sensing image caption generation,'' \emph{IEEE Transactions on Geoscience and Remote Sensing}, vol.~56, no.~4, pp. 2183--2195, 2018.

\bibitem{yuan2022rsitmd}
Z.~Yuan, W.~Zhang, K.~Fu, X.~Li, C.~Deng, H.~Wang, and X.~Sun, ``Exploring a fine-grained multiscale method for cross-modal remote sensing image retrieval,'' \emph{IEEE Transactions on Geoscience and Remote Sensing}, vol.~60, pp. 1--19, 2022.

\bibitem{hu2025ringmoagent}
H.~Hu, P.~Wang, Y.~Feng, K.~Wei, W.~Yin, W.~Diao, M.~Wang, H.~Bi, K.~Kang, T.~Ling \emph{et~al.}, ``Ringmo-agent: A unified remote sensing foundation model for multi-platform and multi-modal reasoning,'' \emph{arXiv preprint arXiv:2507.20776}, 2025.

\bibitem{jiang2025sarcap}
C.~Jiang, C.~Wang, F.~Wu, P.~Ma, L.~Zou, T.~Li, J.~Ning, and Y.~Tang, ``Sarclip: a multimodal foundation framework for sar imagery via contrastive language-image pre-training,'' \emph{ISPRS Journal of Photogrammetry and Remote Sensing}, vol. 231, pp. 17--34, 2026.

\bibitem{ma2025sarchat}
Z.~Ma, X.~Xiao, S.~Dong, P.~Wang, H.~Wang, and Q.~Pan, ``Sarchat-bench-2m: A multi-task vision-language benchmark for sar image interpretation,'' \emph{arXiv preprint arXiv:2502.08168}, 2025.

\bibitem{wei2025sarlang}
Y.~Wei, A.~Xiao, Y.~Ren, Y.~Zhu, H.~Chen, J.~Xia, and N.~Yokoya, ``Sarlang-1m: A benchmark for vision-language modeling in sar image understanding,'' \emph{IEEE Transactions on Geoscience and Remote Sensing}, 2026.

\bibitem{he2025sartext}
Y.~He, X.~Cheng, J.~Zhu, C.~Qiu, J.~Wang, X.~Zhang, Q.~Huang, and K.~Yang, ``Sar-text: A large-scale sar image-text dataset built with sar-narrator and a progressive learning strategy for downstream tasks,'' 2025.

\bibitem{ma2025sarvlm}
Q.~Ma, Z.~Wang, W.~Liu, X.~Lu, B.~Deng, P.~Duan, X.~Kang, and S.~Li, ``Sarvlm: A vision language foundation model for semantic understanding and target recognition in sar imagery,'' 2025.

\bibitem{wang2024skyscript}
Z.~Wang, R.~Prabha, T.~Huang, J.~Wu, and R.~Rajagopal, ``Skyscript: A large and semantically diverse vision-language dataset for remote sensing,'' in \emph{Proceedings of the AAAI Conference on Artificial Intelligence}, vol.~38, no.~6, 2024, pp. 5805--5813.

\bibitem{qu2016cap}
B.~Qu, X.~Li, D.~Tao, and X.~Lu, ``Deep semantic understanding of high resolution remote sensing image,'' in \emph{2016 International Conference on Computer, Information and Telecommunication Systems (CITS)}, 2016, pp. 1--5.

\bibitem{hoxha2022uavcap}
G.~Hoxha and F.~Melgani, ``A novel svm-based decoder for remote sensing image captioning,'' \emph{IEEE Transactions on Geoscience and Remote Sensing}, vol.~60, pp. 1--14, 2022.

\bibitem{li2025ddfav}
H.~Li, X.~Zhang, and H.~Qu, ``Ddfav: Remote sensing large vision language models dataset and evaluation benchmark,'' \emph{Remote Sensing}, vol.~17, no.~4, p. 719, 2025.

\bibitem{luo2024skysensegpt}
J.~Luo, Z.~Pang, Y.~Zhang, T.~Wang, L.~Wang, B.~Dang, J.~Lao, J.~Wang, J.~Chen, Y.~Tan \emph{et~al.}, ``Skysensegpt: A fine-grained instruction tuning dataset and model for remote sensing vision-language understanding,'' \emph{arXiv preprint arXiv:2406.10100}, 2024.

\bibitem{danish2024geobench}
M.~Danish, M.~A. Munir, S.~R.~A. Shah, K.~Kuckreja, F.~S. Khan, P.~Fraccaro, A.~Lacoste, and S.~Khan, ``Geobench-vlm: Benchmarking vision-language models for geospatial tasks,'' in \emph{Proceedings of the IEEE/CVF International Conference on Computer Vision}, 2025, pp. 7132--7142.

\bibitem{ou2025geopix}
R.~Ou, Y.~Hu, F.~Zhang, J.~Chen, and Y.~Liu, ``Geopix: A multimodal large language model for pixel-level image understanding in remote sensing,'' \emph{IEEE Geoscience and Remote Sensing Magazine}, 2025.

\bibitem{zhao2025luojiahog}
Y.~Zhao, M.~Zhang, B.~Yang, Z.~Zhang, J.~Kang, and J.~Gong, ``Luojiahog: A hierarchy oriented geo-aware image caption dataset for remote sensing image--text retrieval,'' \emph{ISPRS Journal of Photogrammetry and Remote Sensing}, vol. 222, pp. 130--151, 2025.

\bibitem{xu2024rs}
L.~Xu, L.~Zhao, W.~Guo, Q.~Li, K.~Long, K.~Zou, Y.~Wang, and H.~Li, ``Rs-gpt4v: A unified multimodal instruction-following dataset for remote sensing image understanding,'' \emph{arXiv preprint arXiv:2406.12479}, 2024.

\bibitem{chen2025rscc}
Z.~Chen, C.~Wang, N.~Zhang, and F.~Zhang, ``Rscc: A large-scale remote sensing change caption dataset for disaster events,'' \emph{Advances in Neural Information Processing Systems}, vol.~38, 2026.

\bibitem{hu2023rsgpt}
Y.~Hu, J.~Yuan, C.~Wen, X.~Lu, Y.~Liu, and X.~Li, ``Rsgpt: A remote sensing vision language model and benchmark,'' \emph{ISPRS Journal of Photogrammetry and Remote Sensing}, vol. 224, pp. 272--286, 2025.

\bibitem{ge2024rsteller}
J.~Ge, X.~Zhang, Y.~Zheng, K.~Guo, and J.~Liang, ``Rsteller: Scaling up visual language modeling in remote sensing with rich linguistic semantics from openly available data and large language models,'' \emph{ISPRS Journal of Photogrammetry and Remote Sensing}, vol. 226, pp. 146--163, 2025.

\bibitem{zhang2025earthmaker}
W.~Zhang, M.~Cai, T.~Zhang, Y.~Zhuang, J.~Li, and X.~Mao, ``Earthmarker: A visual prompting multimodal large language model for remote sensing,'' \emph{IEEE Transactions on Geoscience and Remote Sensing}, vol.~63, pp. 1--19, 2025.

\bibitem{pang2024vhm}
C.~Pang, X.~Weng, J.~Wu, J.~Li, Y.~Liu, J.~Sun, W.~Li, S.~Wang, L.~Feng, G.-S. Xia \emph{et~al.}, ``Vhm: Versatile and honest vision language model for remote sensing image analysis,'' in \emph{Proceedings of the AAAI Conference on Artificial Intelligence}, vol.~39, no.~6, 2025, pp. 6381--6388.

\bibitem{li2024vrsbench}
X.~Li, J.~Ding, and M.~Elhoseiny, ``Vrsbench: A versatile vision-language benchmark dataset for remote sensing image understanding,'' \emph{Advances in Neural Information Processing Systems}, vol.~37, pp. 3229--3242, 2024.

\bibitem{wang2025xlrs}
F.~Wang, H.~Wang, Z.~Guo, D.~Wang, Y.~Wang, M.~Chen, Q.~Ma, L.~Lan, W.~Yang, J.~Zhang \emph{et~al.}, ``Xlrs-bench: Could your multimodal llms understand extremely large ultra-high-resolution remote sensing imagery?'' in \emph{Proceedings of the Computer Vision and Pattern Recognition Conference}, 2025, pp. 14\,325--14\,336.

\bibitem{bashmal2023capera}
L.~Bashmal, Y.~Bazi, M.~M. Al~Rahhal, M.~Zuair, and F.~Melgani, ``Capera: Captioning events in aerial videos,'' \emph{Remote Sensing}, vol.~15, no.~8, p. 2139, 2023.

\bibitem{elgendy2024geollava}
H.~Elgendy, A.~Sharshar, A.~Aboeitta, Y.~Ashraf, and M.~Guizani, ``Geollava: Efficient fine-tuned vision-language models for temporal change detection in remote sensing,'' \emph{arXiv preprint arXiv:2410.19552}, 2024.

\bibitem{liu2022levir-cc}
C.~Liu, R.~Zhao, H.~Chen, Z.~Zou, and Z.~Shi, ``Remote sensing image change captioning with dual-branch transformers: A new method and a large scale dataset,'' \emph{IEEE Transactions on Geoscience and Remote Sensing}, vol.~60, pp. 1--20, 2022.

\bibitem{shabbir2025geopixel}
A.~Shabbir, M.~Zumri, M.~Bennamoun, F.~S. Khan, and S.~Khan, ``Geopixel: Pixel grounding large multimodal model in remote sensing,'' in \emph{International Conference on Machine Learning}.\hskip 1em plus 0.5em minus 0.4em\relax PMLR, 2025, pp. 54\,095--54\,111.

\bibitem{zhou2025avimath}
Y.~Zhou, L.~Feng, M.~Lan, X.~Yang, Q.~Li, Y.~Ke, X.~Jiang, and W.~Zhang, ``Multimodal mathematical reasoning embedded in aerial vehicle imagery: Benchmarking, analysis, and exploration,'' \emph{ISPRS Journal of Photogrammetry and Remote Sensing}, vol. 230, pp. 289--303, 2025.

\bibitem{li2025agromind}
Q.~Li, Y.~Zhang, Z.~Mai, Y.~Chen, H.~Huang, J.~Zhang, Z.~Zhang, Y.~Wen, W.~Li, H.~Fu \emph{et~al.}, ``Can large multimodal models understand agricultural scenes? benchmarking with agromind,'' \emph{Advances in Neural Information Processing Systems}, vol.~38, 2026.

\bibitem{mo2025a2seek}
M.~Mo, X.~Tong, M.~Tan, J.~Leng, J.~Zheng, Y.~Liu, H.~Chen, J.~Gan, W.~Li, and X.~Gao, ``A2seek: Towards reasoning-centric benchmark for aerial anomaly understanding,'' \emph{Advances in Neural Information Processing Systems}, vol.~38, 2026.

\bibitem{guo2025bedi}
M.~Guo, M.~Wu, J.~He, S.~Li, H.~Li, and C.~Tao, ``Bedi: A comprehensive benchmark for evaluating embodied agents on uavs,'' \emph{ISPRS Journal of Photogrammetry and Remote Sensing}, vol. 232, pp. 910--936, 2026.

\bibitem{an2024coreval}
X.~An, J.~Sun, Z.~Gui, and W.~He, ``Coreval: A comprehensive and objective benchmark for evaluating the remote sensing capabilities of large vision-language models,'' \emph{arXiv preprint arXiv:2411.18145}, 2024.

\bibitem{zhang2023crsvqa}
M.~Zhang, F.~Chen, and B.~Li, ``Multistep question-driven visual question answering for remote sensing,'' \emph{IEEE Transactions on Geoscience and Remote Sensing}, vol.~61, pp. 1--12, 2023.

\bibitem{feng2025earthagent}
P.~Feng, Z.~Lv, J.~Ye, X.~Wang, X.~Huo, J.~Yu, W.~Xu, W.~Zhang, L.~Bai, C.~He, and W.~Li, ``Earth-agent: Unlocking the full landscape of earth observation with agents,'' 2025.

\bibitem{wang2024earthvqa}
J.~Wang, Z.~Zheng, Z.~Chen, A.~Ma, and Y.~Zhong, ``Earthvqa: Towards queryable earth via relational reasoning-based remote sensing visual question answering,'' in \emph{Thirtieth AAAI Conference on Artificial Intelligence}, 2024.

\bibitem{shao2025socrates}
R.~Shao, Z.~Li, Z.~Zhang, L.~Xu, X.~He, H.~Yuan, B.~He, Y.~Dai, Y.~Yan, Y.~Chen, W.~Guo, and H.~Li, ``Asking like socrates: Socrates helps vlms understand remote sensing images,'' 2025.

\bibitem{maryam2021floodnet}
M.~Rahnemoonfar, T.~Chowdhury, A.~Sarkar, D.~Varshney, M.~Yari, and R.~R. Murphy, ``Floodnet: A high resolution aerial imagery dataset for post flood scene understanding,'' \emph{IEEE Access}, vol.~9, pp. 89\,644--89\,654, 2021.

\bibitem{li2025geoplan}
K.~Li, J.~Wang, Z.~Wang, H.~Qiao, W.~Zhang, D.~Meng, and X.~Cao, ``Designing domain-specific agents via hierarchical task abstraction mechanism,'' 2025.

\bibitem{tsujimoto2025geo3dvqa}
M.~Tsujimoto, J.~Wang, W.~Xuan, and N.~Yokoya, ``Geo3dvqa: Evaluating vision-language models for 3d geospatial reasoning from aerial imagery,'' in \emph{Proceedings of the IEEE/CVF Winter Conference on Applications of Computer Vision}, 2026, pp. 4994--5003.

\bibitem{li2024hrvqa}
K.~Li, G.~Vosselman, and M.~Y. Yang, ``Hrvqa: A visual question answering benchmark for high-resolution aerial images,'' \emph{ISPRS Journal of Photogrammetry and Remote Sensing}, vol. 214, pp. 65--81, 2024.

\bibitem{chen2025canglingknowflow}
Z.~Chen, H.~Wang, J.~Yao, P.~Ghamisi, J.~Zhou, P.~M. Atkinson, and B.~Zhang, ``Cangling-knowflow: A unified knowledge-and-flow-fused agent for comprehensive remote sensing applications,'' 2025.

\bibitem{xin2025segearthr2}
Z.~Xin, K.~Li, L.~Chen, W.~Li, Y.~Xiao, H.~Qiao, W.~Zhang, D.~Meng, and X.~Cao, ``Segearth-r2: Towards comprehensive language-guided segmentation for remote sensing images,'' 2025.

\bibitem{liu2025zoomearth}
R.~Liu, B.~Fu, J.~Song, K.~Li, W.~Li, L.~Xue, H.~Qiao, W.~Zhang, D.~Meng, and X.~Cao, ``Zoomearth: Active perception for ultra-high-resolution geospatial vision-language tasks,'' 2025.

\bibitem{luo2025large}
J.~Luo, Y.~Zhang, X.~Yang, K.~Wu, Q.~Zhu, L.~Liang, J.~Chen, and Y.~Li, ``When large vision-language model meets large remote sensing imagery: Coarse-to-fine text-guided token pruning,'' in \emph{Proceedings of the IEEE/CVF International Conference on Computer Vision}, 2025, pp. 9206--9217.

\bibitem{zhang2024mme}
Y.~Zhang, H.~Zhang, H.~Tian, C.~Fu, S.~Zhang, J.~Wu, F.~Li, K.~Wang, Q.~Wen, Z.~Zhang \emph{et~al.}, ``Mme-realworld: Could your multimodal llm challenge high-resolution real-world scenarios that are difficult for humans?'' in \emph{International Conference on Learning Representations}, vol. 2025, 2025, pp. 89\,655--89\,701.

\bibitem{dai2025mmuavbench}
S.~Dai, Z.~Ma, Z.~Luo, X.~Yang, Y.~Huang, W.~Zhang, C.~Chen, Z.~Guo, W.~Xu, Y.~Sun, and M.~Sun, ``Mm-uavbench: How well do multimodal large language models see, think, and plan in low-altitude uav scenarios?'' 2025.

\bibitem{zhao2025osvqa}
Z.~Zhao, C.~Zhou, Y.~Zhang, C.~Li, X.~Ma, and J.~Tang, ``Text-guided coarse-to-fine fusion network for robust remote sensing visual question answering,'' \emph{ISPRS Journal of Photogrammetry and Remote Sensing}, vol. 230, pp. 1--17, 2025.

\bibitem{zheng2021mutual}
X.~Zheng, B.~Wang, X.~Du, and X.~Lu, ``Mutual attention inception network for remote sensing visual question answering,'' \emph{IEEE Transactions on Geoscience and Remote Sensing}, vol.~60, pp. 1--14, 2021.

\bibitem{lobry2021rsvqaxben}
S.~Lobry, B.~Demir, and D.~Tuia, ``Rsvqa meets bigearthnet: A new, large-scale, visual question answering dataset for remote sensing,'' in \emph{2021 IEEE International Geoscience and Remote Sensing Symposium (IGARSS)}.\hskip 1em plus 0.5em minus 0.4em\relax IEEE, 2021, pp. 1218--1221.

\bibitem{shabbir2025thinkgeo}
A.~Shabbir, M.~A. Munir, A.~Dudhane, M.~U. Sheikh, M.~H. Khan, P.~Fraccaro, J.~B. Moreno, F.~S. Khan, and S.~Khan, ``Thinkgeo: Evaluating tool-augmented agents for remote sensing tasks,'' 2025.

\bibitem{kao2025univearth}
C.~H. Kao, W.~Zhao, S.~Revankar, S.~Speas, S.~Bhagat, R.~Datta, C.~P. Phoo, U.~Mall, C.~Vondrick, K.~Bala, and B.~Hariharan, ``Towards llm agents for earth observation,'' 2025.

\bibitem{zhao2025urbanvideobench}
B.~Zhao, J.~Fang, Z.~Dai, Z.~Wang, J.~Zha, W.~Zhang, C.~Gao, Y.~Wang, J.~Cui, X.~Chen, and Y.~Li, ``Urbanvideo-bench: Benchmarking vision-language models on embodied intelligence with video data in urban spaces,'' in \emph{Proceedings of the 63rd Annual Meeting of the Association for Computational Linguistics (Volume 1: Long Papers)}.\hskip 1em plus 0.5em minus 0.4em\relax Association for Computational Linguistics, 2025, p. 32400–32423.

\bibitem{zhang2024good}
C.~Zhang and S.~Wang, ``Good at captioning bad at counting: Benchmarking gpt-4v on earth observation data,'' in \emph{Proceedings of the IEEE/CVF Conference on Computer Vision and Pattern Recognition}, 2024, pp. 7839--7849.

\bibitem{wang2025vicotagent}
C.~Wang, Z.~Luo, R.~Liu, C.~Ran, S.~Fan, X.~Chen, and C.~He, ``Vicot-agent: A vision-interleaved chain-of-thought framework for interpretable multimodal reasoning and scalable remote sensing analysis,'' 2025.

\bibitem{al2022open}
M.~M. Al~Rahhal, Y.~Bazi, S.~O. Alsaleh, M.~Al-Razgan, M.~L. Mekhalfi, M.~Al~Zuair, and N.~Alajlan, ``Open-ended remote sensing visual question answering with transformers,'' \emph{International Journal of Remote Sensing}, vol.~43, no.~18, pp. 6809--6823, 2022.

\bibitem{irvin2024teochat}
J.~Irvin, E.~Liu, J.~Chen, I.~Dormoy, J.~Kim, S.~Khanna, Z.~Zheng, and S.~Ermon, ``Teochat: A large vision-language assistant for temporal earth observation data,'' in \emph{International Conference on Learning Representations}, vol. 2025, 2025, pp. 68\,883--68\,911.

\bibitem{zhou2024urbench}
B.~Zhou, H.~Yang, D.~Chen, J.~Ye, T.~Bai, J.~Yu, S.~Zhang, D.~Lin, C.~He, and W.~Li, ``Urbench: A comprehensive benchmark for evaluating large multimodal models in multi-view urban scenarios,'' in \emph{Proceedings of the AAAI Conference on Artificial Intelligence}, vol.~39, no.~10, 2025, pp. 10\,707--10\,715.

\bibitem{li2024qvg}
K.~Li, F.~Dong, D.~Wang, S.~Li, Q.~Wang, X.~Gao, and T.-S. Chua, ``Show me what and where has changed? question answering and grounding for remote sensing change detection,'' \emph{arXiv preprint arXiv:2410.23828}, 2024.

\bibitem{yuan2022cdvqa}
Z.~Yuan, L.~Mou, Z.~Xiong, and X.~X. Zhu, ``Change detection meets visual question answering,'' \emph{IEEE Transactions on Geoscience and Remote Sensing}, vol.~60, pp. 1--13, 2022.

\bibitem{zhou2024geoground}
Y.~Zhou, M.~Lan, X.~Li, Y.~Ke, X.~Jiang, L.~Feng, and W.~Zhang, ``Geoground: A unified large vision-language model. for remote sensing visual grounding,'' \emph{arXiv preprint arXiv:2411.11904}, 2024.

\bibitem{zhan2023dior-rsvg}
Y.~Zhan, Z.~Xiong, and Y.~Yuan, ``Rsvg: Exploring data and models for visual grounding on remote sensing data,'' \emph{IEEE Transactions on Geoscience and Remote Sensing}, vol.~61, pp. 1--13, 2023.

\bibitem{chu2024towards}
M.~Chu, Z.~Zheng, W.~Ji, T.~Wang, and T.-S. Chua, ``Towards natural language-guided drones: Geotext-1652 benchmark with spatial relation matching,'' in \emph{European Conference on Computer Vision}.\hskip 1em plus 0.5em minus 0.4em\relax Springer, 2024, pp. 213--231.

\bibitem{li2024opt-rsvg}
K.~Li, D.~Wang, H.~Xu, H.~Zhong, and C.~Wang, ``Language-guided progressive attention for visual grounding in remote sensing images,'' \emph{IEEE Transactions on Geoscience and Remote Sensing}, vol.~62, pp. 1--13, 2024.

\bibitem{sun2022rsvg}
Y.~Sun, S.~Feng, X.~Li, Y.~Ye, J.~Kang, and X.~Huang, ``Visual grounding in remote sensing images,'' in \emph{Proceedings of the 30th ACM International Conference on Multimedia}, 2022, pp. 404--412.

\bibitem{lan2024lqvg}
M.~Lan, F.~Rong, H.~Jiao, Z.~Gao, and L.~Zhang, ``Language query-based transformer with multiscale cross-modal alignment for visual grounding on remote sensing images,'' \emph{IEEE Transactions on Geoscience and Remote Sensing}, vol.~62, pp. 1--13, 2024.

\bibitem{liu2025aerialvg}
J.~Liu, Q.~Chen, Z.~Wang, Y.~Tang, Y.~Zhang, C.~Yan, D.~Wang, X.~Li, and B.~Zhao, ``Aerialvg: A challenging benchmark for aerial visual grounding by exploring positional relations,'' in \emph{Proceedings of the IEEE/CVF International Conference on Computer Vision}, 2025, pp. 5177--5187.

\bibitem{yao2025remotesam}
L.~Yao, F.~Liu, D.~Chen, C.~Zhang, Y.~Wang, Z.~Chen, W.~Xu, S.~Di, and Y.~Zheng, ``Remotesam: Towards segment anything for earth observation,'' in \emph{Proceedings of the 33rd ACM International Conference on Multimedia}, 2025, pp. 3027--3036.

\bibitem{sun2025refdrone}
Z.~Sun, Y.~Liu, Z.~Su, H.~Zhu, Y.~Gu, Y.~Zou, Z.~Liu, G.-S. Xia, B.~Du, and Y.~Xu, ``Refdrone: A challenging benchmark for referring expression comprehension in drone scenes,'' 2025.

\bibitem{ni2025unigeoseg}
S.~Ni, D.~Wang, H.~Chen, H.~Guo, N.~Zhang, and J.~Zhang, ``Unigeoseg: Towards unified open-world segmentation for geospatial scenes,'' 2025.

\bibitem{jiang2025grasp}
C.~Jiang, Y.~Zhou, J.~Yan, J.~Li, J.~Li, Y.~Zhou, H.~He, and J.~Li, ``Grasp: Geospatial pixel reasoning via structured policy learning,'' 2025.

\bibitem{quenum2025lisat}
J.~Quenum, W.-H. Hsieh, T.-H.~P. Wu, R.~Gupta, T.~Darrell, and D.~Chan, ``Lisat: Language-instructed segmentation assistant for satellite imagery,'' \emph{Advances in Neural Information Processing Systems}, vol.~38, 2026.

\bibitem{dong2024risbench}
Z.~Dong, Y.~Sun, Y.~Gu, and T.~Liu, ``Cross-modal bidirectional interaction model for referring remote sensing image segmentation,'' \emph{arXiv preprint arXiv:2410.08613}, 2024.

\bibitem{ye2025rislad}
K.~Ye, Y.~Luan, Z.~Chen, G.~Meng, P.~Dai, and L.~Cao, ``Ris-lad: A benchmark and model for referring low-altitude drone image segmentation,'' 2025.

\bibitem{liu2024rotated}
S.~Liu, Y.~Ma, X.~Zhang, H.~Wang, J.~Ji, X.~Sun, and R.~Ji, ``Rotated multi-scale interaction network for referring remote sensing image segmentation,'' in \emph{Proceedings of the IEEE/CVF Conference on Computer Vision and Pattern Recognition}, 2024, pp. 26\,658--26\,668.

\bibitem{li2025segearth}
K.~Li, Z.~Xin, L.~Pang, C.~Pang, Y.~Deng, J.~Yao, G.~Xia, D.~Meng, Z.~Wang, and X.~Cao, ``Segearth-r1: Geospatial pixel reasoning via large language model,'' \emph{arXiv preprint arXiv:2504.09644}, 2025.

\bibitem{yang2025nwpurefer}
Z.~Yang, H.~Yao, L.~Tian, X.~Zhao, Q.~Li, and Q.~Wang, ``A large-scale referring remote sensing image segmentation dataset and benchmark,'' 2025.

\bibitem{yuan2024rrsis}
Z.~Yuan, L.~Mou, Y.~Hua, and X.~X. Zhu, ``Rrsis: Referring remote sensing image segmentation,'' \emph{IEEE Transactions on Geoscience and Remote Sensing}, vol.~62, pp. 1--12, 2024.

\bibitem{genc2022uavcap}
G.~Hoxha and F.~Melgani, ``A novel svm-based decoder for remote sensing image captioning,'' \emph{IEEE Transactions on Geoscience and Remote Sensing}, vol.~60, pp. 1--14, 2022.

\bibitem{chen2025dgtrsd}
W.~Chen, Y.~Deng, W.~Jin, J.~Chen, J.~Chen, Y.~Feng, Z.~Xi, D.~Liu, K.~Li, and Y.~Meng, ``Dgtrsd and dgtrsclip: A dual-granularity remote sensing image--text dataset and vision--language foundation model for alignment,'' \emph{IEEE Journal of Selected Topics in Applied Earth Observations and Remote Sensing}, vol.~18, pp. 29\,113--29\,130, 2025.

\bibitem{marimo2025ssl4eos}
C.~T. Marimo, B.~Blumenstiel, M.~Nitsche, J.~Jakubik, and T.~Brunschwiler, ``Beyond the visible: Multispectral vision-language learning for earth observation,'' pp. 359--375, 2025.

\bibitem{li2025rscovlm}
Q.~Li, S.~Ma, J.~Luo, Y.~Yu, Y.~Zhou, F.~Wang, X.~Lu, X.~Wang, X.~He, Y.~Chen \emph{et~al.}, ``Co-training vision-language models for remote sensing multi-task learning,'' \emph{Remote Sensing}, vol.~18, no.~2, p. 222, 2026.

\bibitem{xiao2021next}
J.~Xiao, X.~Shang, A.~Yao, and T.-S. Chua, ``Next-qa: Next phase of question-answering to explaining temporal actions,'' in \emph{Proceedings of the IEEE/CVF conference on computer vision and pattern recognition}, 2021, pp. 9777--9786.

\bibitem{xue2024reo}
X.~Xue, G.~Wei, H.~Chen, H.~Zhang, F.~Lin, C.~Shen, and X.~X. Zhu, ``Reo-vlm: Transforming vlm to meet regression challenges in earth observation,'' \emph{arXiv preprint arXiv:2412.16583}, 2024.

\bibitem{li2025simple}
Q.~Li, X.~He, X.~Shu, Y.~Yu, D.~Chen, Y.~Chen, and X.~Yang, ``A simple aerial detection baseline of multimodal language models,'' in \emph{IGARSS 2025-2025 IEEE International Geoscience and Remote Sensing Symposium}.\hskip 1em plus 0.5em minus 0.4em\relax IEEE, 2025, pp. 6833--6837.

\bibitem{zhan2025skyeyegpt}
Y.~Zhan, Z.~Xiong, and Y.~Yuan, ``Skyeyegpt: Unifying remote sensing vision-language tasks via instruction tuning with large language model,'' \emph{ISPRS Journal of Photogrammetry and Remote Sensing}, vol. 221, pp. 64--77, 2025.

\bibitem{zhang2025earthgptx}
W.~Zhang, M.~Cai, Y.~Ning, T.~Zhang, Y.~Zhuang, S.~Lu, H.~Chen, J.~Li, and X.~Mao, ``Earthgpt-x: A spatial mllm for multilevel multisource remote sensing imagery understanding with visual prompting,'' \emph{IEEE Transactions on Geoscience and Remote Sensing}, vol.~63, pp. 1--21, 2025.

\bibitem{an2025choice}
X.~An, J.~Sun, Z.~Gui, and W.~He, ``Choice: benchmarking the remote sensing capabilities of large vision-language models,'' \emph{Advances in Neural Information Processing Systems}, vol.~38, 2026.

\bibitem{zhou2025airspatialbot}
Y.~Zhou, R.~Ding, X.~Yang, X.~Jiang, and X.~Liu, ``Airspatialbot: A spatially aware aerial agent for fine-grained vehicle attribute recognition and retrieval,'' \emph{IEEE Transactions on Geoscience and Remote Sensing}, vol.~63, pp. 1--12, 2025.

\bibitem{lin2023comparison}
J.~C. Lin, D.~N. Younessi, S.~S. Kurapati, O.~Y. Tang, and I.~U. Scott, ``Comparison of gpt-3.5, gpt-4, and human user performance on a practice ophthalmology written examination,'' \emph{Eye}, vol.~37, no.~17, pp. 3694--3695, 2023.

\bibitem{benzon2025llm}
W.~L. Benzon, ``From llm mechanisms to ring-composition: A conversation with claude 3.5 working paper,'' \emph{Available at SSRN}, 2025.

\bibitem{team2023gemini}
G.~Team, R.~Anil, S.~Borgeaud, J.-B. Alayrac, J.~Yu, R.~Soricut, J.~Schalkwyk, A.~M. Dai, A.~Hauth, K.~Millican \emph{et~al.}, ``Gemini: a family of highly capable multimodal models,'' \emph{arXiv preprint arXiv:2312.11805}, 2023.

\bibitem{liu2023llava}
H.~Liu, C.~Li, Q.~Wu, and Y.~J. Lee, ``Visual instruction tuning,'' \emph{Advances in neural information processing systems}, vol.~36, pp. 34\,892--34\,916, 2023.

\bibitem{wang2026cityvlm}
J.~Wang, W.~Xuan, H.~Qi, Z.~Chen, H.~Chen, Z.~Zheng, J.~Xia, Y.~Zhong, and N.~Yokoya, ``Cityvlm: Towards sustainable urban development via multi-view coordinated vision–language model,'' \emph{ISPRS Journal of Photogrammetry and Remote Sensing}, vol. 232, pp. 62--74, 2026.

\bibitem{dior}
K.~Li, G.~Wan, G.~Cheng, L.~Meng, and J.~Han, ``Object detection in optical remote sensing images: A survey and a new benchmark,'' \emph{ISPRS J. Photogramm. Remote Sens.}, vol. 159, pp. 296--307, 2020.

\bibitem{zhang2019hierarchical}
Y.~Zhang, Y.~Yuan, Y.~Feng, and X.~Lu, ``Hierarchical and robust convolutional neural network for very high-resolution remote sensing object detection,'' \emph{IEEE Transactions on Geoscience and Remote Sensing}, vol.~57, no.~8, pp. 5535--5548, 2019.

\bibitem{song2020semantic}
A.~Song and Y.~Kim, ``Semantic segmentation of remote-sensing imagery using heterogeneous big data: International society for photogrammetry and remote sensing potsdam and cityscape datasets,'' \emph{ISPRS International Journal of Geo-Information}, vol.~9, no.~10, p. 601, 2020.

\bibitem{christie2018functional}
G.~Christie, N.~Fendley, J.~Wilson, and R.~Mukherjee, ``Functional map of the world,'' in \emph{Proceedings of the IEEE Conference on Computer Vision and Pattern Recognition}, 2018, pp. 6172--6180.

\bibitem{2021_4e732ced}
\BIBentryALTinterwordspacing
J.~Wang, Z.~Zheng, A.~Ma, X.~Lu, and Y.~Zhong, ``Loveda: A remote sensing land-cover dataset for domain adaptive semantic segmentation,'' in \emph{Proceedings of the Neural Information Processing Systems Track on Datasets and Benchmarks}, J.~Vanschoren and S.~Yeung, Eds., vol.~1.\hskip 1em plus 0.5em minus 0.4em\relax Curran Associates, Inc., 2021. [Online]. Available: \url{https://datasets-benchmarks-proceedings.neurips.cc/paper_files/paper/2021/file/4e732ced3463d06de0ca9a15b6153677-Paper-round2.pdf}
\BIBentrySTDinterwordspacing

\bibitem{zoph2020rethinking}
B.~Zoph, G.~Ghiasi, T.-Y. Lin, Y.~Cui, H.~Liu, E.~D. Cubuk, and Q.~Le, ``Rethinking pre-training and self-training,'' \emph{Advances in neural information processing systems}, vol.~33, pp. 3833--3845, 2020.

\bibitem{chen2023vlp}
F.-L. Chen, D.-Z. Zhang, M.-L. Han, X.-Y. Chen, J.~Shi, S.~Xu, and B.~Xu, ``Vlp: A survey on vision-language pre-training,'' \emph{Machine Intelligence Research}, vol.~20, no.~1, pp. 38--56, 2023.

\bibitem{ma2025exploring}
W.~Ma, R.~Zhang, X.~Ning, H.~Zhang, and Y.~Chen, ``Exploring the potential of vlms in remote sensing through prompt optimization,'' \emph{The International Archives of the Photogrammetry, Remote Sensing and Spatial Information Sciences}, vol.~48, pp. 219--226, 2025.

\bibitem{lin2024vila}
J.~Lin, H.~Yin, W.~Ping, P.~Molchanov, M.~Shoeybi, and S.~Han, ``Vila: On pre-training for visual language models,'' in \emph{Proceedings of the IEEE/CVF conference on computer vision and pattern recognition}, 2024, pp. 26\,689--26\,699.

\bibitem{yang2024cross}
J.~Yang, Z.~Li, S.~Xie, W.~Zhu, W.~Yu, and S.~Li, ``Cross-modal adapter: Parameter-efficient transfer learning approach for vision-language models,'' in \emph{2024 IEEE International Conference on Multimedia and Expo (ICME)}.\hskip 1em plus 0.5em minus 0.4em\relax IEEE, 2024, pp. 1--6.

\bibitem{li2023lift}
J.~Li and H.~Sun, ``Lift: Transfer learning in vision-language models for downstream adaptation and generalization,'' in \emph{Proceedings of the 31st ACM International Conference on Multimedia}, 2023, pp. 4678--4687.

\bibitem{lv2024full}
K.~Lv, Y.~Yang, T.~Liu, Q.~Guo, and X.~Qiu, ``Full parameter fine-tuning for large language models with limited resources,'' in \emph{Proceedings of the 62nd Annual Meeting of the Association for Computational Linguistics (Volume 1: Long Papers)}, 2024, pp. 8187--8198.

\bibitem{liu2024hift}
Y.~Liu, Y.~Zhang, Q.~Li, T.~Liu, S.~Feng, D.~Wang, Y.~Zhang, and H.~Sch{\"u}tze, ``Hift: A hierarchical full parameter fine-tuning strategy,'' \emph{arXiv preprint arXiv:2401.15207}, 2024.

\bibitem{ding2023parameter}
N.~Ding, Y.~Qin, G.~Yang, F.~Wei, Z.~Yang, Y.~Su, S.~Hu, Y.~Chen, C.-M. Chan, W.~Chen \emph{et~al.}, ``Parameter-efficient fine-tuning of large-scale pre-trained language models,'' \emph{Nature machine intelligence}, vol.~5, no.~3, pp. 220--235, 2023.

\bibitem{wang2025parameter}
L.~Wang, S.~Chen, L.~Jiang, S.~Pan, R.~Cai, S.~Yang, and F.~Yang, ``Parameter-efficient fine-tuning in large language models: a survey of methodologies,'' \emph{Artificial Intelligence Review}, vol.~58, no.~8, p. 227, 2025.

\bibitem{dong2024abilities}
G.~Dong, H.~Yuan, K.~Lu, C.~Li, M.~Xue, D.~Liu, W.~Wang, Z.~Yuan, C.~Zhou, and J.~Zhou, ``How abilities in large language models are affected by supervised fine-tuning data composition,'' in \emph{Proceedings of the 62nd Annual Meeting of the Association for Computational Linguistics (Volume 1: Long Papers)}, 2024, pp. 177--198.

\bibitem{wang2023self}
R.~Wang, Y.~Hao, L.~Hu, J.~Chen, M.~Chen, and D.~Wu, ``Self-supervised learning with data-efficient supervised fine-tuning for crowd counting,'' \emph{IEEE Transactions on Multimedia}, vol.~25, pp. 1538--1546, 2023.

\bibitem{fan2023dpok}
Y.~Fan, O.~Watkins, Y.~Du, H.~Liu, M.~Ryu, C.~Boutilier, P.~Abbeel, M.~Ghavamzadeh, K.~Lee, and K.~Lee, ``Dpok: Reinforcement learning for fine-tuning text-to-image diffusion models,'' \emph{Advances in Neural Information Processing Systems}, vol.~36, pp. 79\,858--79\,885, 2023.

\bibitem{zhai2024fine}
S.~Zhai, H.~Bai, Z.~Lin, J.~Pan, P.~Tong, Y.~Zhou, A.~Suhr, S.~Xie, Y.~LeCun, Y.~Ma \emph{et~al.}, ``Fine-tuning large vision-language models as decision-making agents via reinforcement learning,'' \emph{Advances in neural information processing systems}, vol.~37, pp. 110\,935--110\,971, 2024.

\bibitem{ziegler2025craft}
I.~Ziegler, A.~K{\"o}ksal, D.~Elliott, and H.~Sch{\"u}tze, ``Craft your dataset: Task-specific synthetic dataset generation through corpus retrieval and augmentation,'' \emph{Transactions of the Association for Computational Linguistics}, vol.~13, pp. 1693--1721, 2025.

\bibitem{hu2022lora}
E.~J. Hu, Y.~Shen, P.~Wallis, Z.~Allen-Zhu, Y.~Li, S.~Wang, L.~Wang, W.~Chen \emph{et~al.}, ``Lora: Low-rank adaptation of large language models.'' \emph{ICLR}, vol.~1, no.~2, p.~3, 2022.

\bibitem{yakoub2024rsllava}
Y.~Bazi, L.~Bashmal, M.~M. Al~Rahhal, R.~Ricci, and F.~Melgani, ``Rs-llava: A large vision-language model for joint captioning and question answering in remote sensing imagery,'' \emph{Remote Sensing}, vol.~16, no.~9, 2024.

\bibitem{pan2025medvlm}
J.~Pan, C.~Liu, J.~Wu, F.~Liu, J.~Zhu, H.~B. Li, C.~Chen, C.~Ouyang, and D.~Rueckert, ``Medvlm-r1: Incentivizing medical reasoning capability of vision-language models (vlms) via reinforcement learning,'' in \emph{International Conference on Medical Image Computing and Computer-Assisted Intervention}.\hskip 1em plus 0.5em minus 0.4em\relax Springer, 2025, pp. 337--347.

\bibitem{rodriguez2025integrating}
J.~Rodriguez-Juan, D.~Ortiz-Perez, J.~Garcia-Rodriguez, D.~Tom{\'a}s, and G.~J. Nalepa, ``Integrating advanced vision-language models for context recognition in risks assessment,'' \emph{Neurocomputing}, vol. 618, p. 129131, 2025.

\bibitem{wu2025fsvlm}
H.~Wu, Z.~Du, D.~Zhong, Y.~Wang, and C.~Tao, ``Fsvlm: A vision-language model for remote sensing farmland segmentation,'' \emph{IEEE Transactions on Geoscience and Remote Sensing}, 2025.

\bibitem{boisseau2024imitation}
{\'E}.~Boisseau, ``Imitation and large language models,'' \emph{Minds and Machines}, vol.~34, no.~4, p.~42, 2024.

\bibitem{alvarado2025turing}
R.~C. Alvarado, ``Turing’s children: Large language models and imitative intelligence,'' \emph{Social Science Computer Review}, p. 08944393251364293, 2025.

\bibitem{agnihotri2025efficient}
R.~Agnihotri and A.~Pandey, ``Efficient online rft with plug-and-play llm judges: Unlocking state-of-the-art performance,'' \emph{arXiv preprint arXiv:2506.05748}, 2025.

\bibitem{zhang2025geor1}
Z.~Zhang, Z.~Guan, T.~Zhao, H.~Shen, Y.~Cai, Z.~Su, Y.~Shang, Z.~Liu, J.~Yin, and X.~Li, ``Geo-r1: Improving few-shot geospatial referring expression understanding with reinforcement fine-tuning,'' \emph{ISPRS Journal of Photogrammetry and Remote Sensing}, vol. 237, pp. 113--129, 2026.

\bibitem{wang2025geozero}
D.~Wang, S.~Liu, W.~Jiang, F.~Wang, Y.~Liu, X.~Qin, Z.~Luo, C.~Zhou, H.~Guo, J.~Zhang, B.~Du, D.~Tao, and L.~Zhang, ``Geozero: Incentivizing reasoning from scratch on geospatial scenes,'' 2025.

\bibitem{stechly2024chain}
K.~Stechly, K.~Valmeekam, and S.~Kambhampati, ``Chain of thoughtlessness? an analysis of cot in planning,'' \emph{Advances in Neural Information Processing Systems}, vol.~37, pp. 29\,106--29\,141, 2024.

\bibitem{xia2025beyond}
Y.~Xia, R.~Wang, X.~Liu, M.~Li, T.~Yu, X.~Chen, J.~McAuley, and S.~Li, ``Beyond chain-of-thought: A survey of chain-of-x paradigms for llms,'' in \emph{Proceedings of the 31st International Conference on Computational Linguistics}, 2025, pp. 10\,795--10\,809.

\bibitem{lyu2023faithful}
Q.~Lyu, S.~Havaldar, A.~Stein, L.~Zhang, D.~Rao, E.~Wong, M.~Apidianaki, and C.~Callison-Burch, ``Faithful chain-of-thought reasoning,'' in \emph{The 13th International Joint Conference on Natural Language Processing and the 3rd Conference of the Asia-Pacific Chapter of the Association for Computational Linguistics (IJCNLP-AACL 2023)}, 2023.

\bibitem{2026DVGBench}
Y.~Zhou, J.~Chen, Z.~Zhang, P.~Huang, R.~Ding, Z.~Zou, P.~F. Gao, Y.~Wei, K.~Li, and X.~Yang, ``Dvgbench: Implicit-to-explicit visual grounding benchmark in uav imagery with large vision–language models,'' \emph{ISPRS Journal of Photogrammetry and Remote Sensing}, vol. 232, no.~c, pp. 831--847, 2026.

\bibitem{silva2024rscapret}
J.~D. Silva, J.~Magalhães, D.~Tuia, and B.~Martins, ``Large language models for captioning and retrieving remote sensing images,'' 2024.

\bibitem{2002BLEU}
K.~Papineni, S.~Roukos, T.~Ward, and W.~J. Zhu, ``Bleu: a method for automatic evaluation of machine translation,'' in \emph{Proc Meeting of the Association for Computational Linguistics}, 2002.

\bibitem{2005METEOR}
B.~Satanjeev, ``Meteor: An automatic metric for mt evaluation with improved correlation with human judgments,'' \emph{ACL-2005}, pp. 228--231, 2005.

\bibitem{2004ROUGE}
C.~Y. Lin, ``Rouge: A package for automatic evaluation of summaries,'' 2004.

\bibitem{2014CIDEr}
R.~Vedantam, C.~L. Zitnick, and D.~Parikh, ``Cider: Consensus-based image description evaluation,'' 2014.

\bibitem{2015VQA}
S.~Antol, A.~Agrawal, J.~Lu, M.~Mitchell, and D.~Parikh, ``Vqa: Visual question answering,'' \emph{International Journal of Computer Vision}, vol. 123, no.~1, pp. 4--31, 2015.

\bibitem{2015Exploring}
M.~Ren, R.~Kiros, and R.~Zemel, ``Exploring models and data for image question answering,'' \emph{Litoral Revista De La Poesía Y El Pensamiento}, pp. 2953--2961, 2015.

\bibitem{liu2024rsunivlm}
X.~Liu and Z.~Lian, ``Rsunivlm: A unified vision-language model for remote sensing via granularity-oriented moe,'' \emph{Pattern Recognition}, p. 113717, 2026.

\bibitem{zhang2025imagerag}
Z.~Zhang, H.~Shen, T.~Zhao, Z.~Guan, B.~Chen, Y.~Wang, X.~Jia, Y.~Cai, Y.~Shang, and J.~Yin, ``Enhancing ultrahigh resolution remote sensing imagery analysis with imagerag: A new framework,'' \emph{IEEE Geoscience and Remote Sensing Magazine}, vol.~13, no.~3, pp. 369--394, 2025.

\bibitem{zhou2025zoom}
Y.~Zhou, C.~Jiang, C.~Yuan, and J.~Li, ``Look where it matters: Training-free ultra-hr remote sensing vqa via adaptive zoom search,'' 2025.

\bibitem{yao2025remotereasoner}
L.~Yao, F.~Liu, H.~Lu, C.~Zhang, R.~Min, S.~Xu, S.~Di, and P.~Peng, ``Remotereasoner: Towards unifying geospatial reasoning workflow,'' in \emph{Proceedings of the AAAI Conference on Artificial Intelligence}, vol.~40, no.~14, 2026, pp. 11\,883--11\,891.

\bibitem{2016Generation}
J.~Mao, J.~Huang, A.~Toshev, O.~Camburu, and K.~Murphy, ``Generation and comprehension of unambiguous object descriptions,'' in \emph{2016 IEEE Conference on Computer Vision and Pattern Recognition (CVPR)}, 2016.

\bibitem{2025Land}
M.~Zhou, H.~Guo, X.~Ouyang, D.~Gunasekera, and Z.~Sun, ``Land surface temperature retrieval from sdgsat-1: assessment of different retrieval algorithms with different atmospheric reanalysis data,'' \emph{International Journal of Digital Earth}, vol.~18, no.~1, 2025.

\bibitem{yang2024thinking}
J.~Yang, S.~Yang, A.~W. Gupta, R.~Han, L.~Fei-Fei, and S.~Xie, ``Thinking in space: How multimodal large language models see, remember, and recall spaces,'' in \emph{Proceedings of the Computer Vision and Pattern Recognition Conference}, 2025, pp. 10\,632--10\,643.

\bibitem{he2017mask}
K.~He, G.~Gkioxari, P.~Doll{\'a}r, and R.~Girshick, ``Mask r-cnn,'' in \emph{Proceedings of the IEEE international conference on computer vision}, 2017, pp. 2961--2969.

\bibitem{WU2023103495}
B.~Wu, H.~Huang, Y.~Wang, S.~Shi, J.~Wu, and B.~Yu, ``Global spatial patterns between nighttime light intensity and urban building morphology,'' \emph{International Journal of Applied Earth Observation and Geoinformation}, vol. 124, p. 103495, 2023.

\bibitem{zhu2023thermal}
J.~Zhu, Y.~Xu, L.~Hoegner, and U.~Stilla, ``Generation of thermal point clouds from uncalibrated thermal infrared image sequences and mobile laser scans,'' \emph{IEEE Transactions on Instrumentation and Measurement}, vol.~72, pp. 1--16, 2023.

\bibitem{gao2025openfly}
Y.~Gao, C.~Li, Z.~You, J.~Liu, Z.~Li, P.~Chen, Q.~Chen, Z.~Tang, L.~Wang, P.~Yang \emph{et~al.}, ``Openfly: A versatile toolchain and large-scale benchmark for aerial vision-language navigation,'' \emph{arXiv e-prints}, pp. arXiv--2502, 2025.

\bibitem{zhang2025spatialsky}
L.~Zhang, Y.~Zhang, H.~Li, H.~Fu, Y.~Tang, H.~Ye, L.~Chen, X.~Liang, X.~Hao, and W.~Ding, ``Is your vlm sky-ready? a comprehensive spatial intelligence benchmark for uav navigation,'' \emph{arXiv preprint arXiv:2511.13269}, 2025.

\bibitem{zha2025aircop}
J.~Zha, Y.~Fan, T.~Zhang, G.~Chen, Y.~Chen, C.~Gao, and X.~Chen, ``Aircopbench: A benchmark for multi-drone collaborative embodied perception and reasoning,'' in \emph{Proceedings of the AAAI Conference on Artificial Intelligence}, vol.~40, no.~2, 2026, pp. 1507--1515.

\bibitem{wang2025jtd}
Y.~Wang, J.~Zhao, Z.~Fan, X.~Zhang, X.~Wu, Y.~Zhang, L.~Jin, X.~Li, G.~Wang, M.~Jia \emph{et~al.}, ``Jtd-uav: Mllm-enhanced joint tracking and description framework for anti-uav systems,'' in \emph{Proceedings of the Computer Vision and Pattern Recognition Conference}, 2025, pp. 1633--1644.

\bibitem{wang2025uavflow}
X.~Wang, D.~Yang, Y.~Liao, W.~Zheng, B.~Dai, H.~Li, S.~Liu \emph{et~al.}, ``Uav-flow colosseo: A real-world benchmark for flying-on-a-word uav imitation learning,'' \emph{Advances in Neural Information Processing Systems}, vol.~38, 2026.

\bibitem{zhang2025open3dvqa}
W.~Zhang, Z.~Zhou, X.~Zeng, L.~Xuchen, J.~Fang, C.~Gao, J.~Cui, Y.~Li, X.~Chen, and X.-P. Zhang, ``Open3d-vqa: A benchmark for embodied spatial concept reasoning with multimodal large language model in open space,'' ser. MM '25.\hskip 1em plus 0.5em minus 0.4em\relax New York, NY, USA: Association for Computing Machinery, 2025, p. 12784–12791.

\end{thebibliography}
\end{document}